\documentclass[10pt,twocolumn,letterpaper]{article}
\pdfoutput=1
\usepackage{cvpr25}              
\usepackage[numbers,sort&compress]{natbib}
\usepackage[utf8]{inputenc} 
\usepackage[T1]{fontenc}    
\usepackage{url}            
\usepackage{booktabs}       
\usepackage{amsfonts}       
\usepackage{nicefrac}       
\usepackage{microtype}      
\usepackage{cite}
\usepackage{amsmath,amssymb,amsfonts}
\usepackage{graphicx}
\usepackage{textcomp}
\usepackage{nicefrac}
\usepackage{pifont}
\usepackage{soul}
\usepackage[font=footnotesize,labelfont=bf]{caption}
\usepackage{enumitem}
\usepackage[table,dvipsnames]{xcolor}
\usepackage{anyfontsize}
\usepackage[normalem]{ulem}
\usepackage{wrapfig}
\usepackage[fixed]{fontawesome5}
\usepackage{multirow}
\usepackage[export]{adjustbox}
\usepackage{stackengine}    
\usepackage{float}    
\usepackage[edges]{forest}
\usepackage{array}
\usepackage{stfloats}
\usepackage[outline]{contour}
\usepackage{xspace}
\usepackage{dcolumn}
\usepackage{diagbox} 
\usepackage{placeins} 
\usepackage{enumitem}
\usepackage{balance}
\usepackage{paralist}

\usepackage{styles}

\usepackage{indentfirst} 
\usepackage{makecell}
\usepackage{subcaption}
\PassOptionsToPackage{hyphens}{url}
\usepackage{pgfplots}
\usepgfplotslibrary{fillbetween,statistics}
\pgfplotsset{compat=1.17}

\usepackage[para, hang,flushmargin]{footmisc}
\usepackage[accsupp]{axessibility}  

\newcommand{\REDACTED}[1]{#1}

\usepackage[implicit=false,colorlinks,bookmarks=false, pagebackref,breaklinks,colorlinks]{hyperref}

\usepackage[capitalize]{cleveref}
\crefname{section}{Sec.}{Secs.}
\Crefname{section}{Section}{Sections}
\Crefname{table}{Table}{Tables}
\crefname{table}{Tab.}{Tabs.}

\title{Open-Canopy: Towards Very High Resolution Forest Monitoring}

\def\paperID{827} 
\def\confName{CVPR}
\def\confYear{2025}

\author{
Fajwel Fogel\textsuperscript{4}
\and
Yohann Perron\textsuperscript{3,5}
\and
 Nikola Besic\textsuperscript{2}
\and 
Laurent Saint-André\textsuperscript{7}
\and 
Agnès Pellissier-Tanon\textsuperscript{1} 
\and 
Martin Schwartz\textsuperscript{1} 
\and 
Thomas Boudras\textsuperscript{1} 
\and
Ibrahim Fayad\textsuperscript{1,6} 
\and
\qquad Alexandre d'Aspremont\textsuperscript{4,6}
\and
Loic Landrieu\textsuperscript{3}
\and
Philippe Ciais\textsuperscript{1}\qquad
\and
\textsuperscript{1} \small  LSCE/IPSL, CEA-CNRS-UVSQ
\and
 \textsuperscript{2} \small LIF, IGN, ENSG
\and
 \textsuperscript{3} \small  LIGM, Ecole des Ponts, IP Paris CNRS, UGE
\and
\textsuperscript{4} \small CNRS \& École Normale Supérieure
\and
\textsuperscript{5} \small  EFEO
\and
\textsuperscript{6} \small Kayrros
\and
 \textsuperscript{7} \small INRAE, BEF
}

\begin{document}

\maketitle

\begin{abstract}
Estimating canopy height and its changes at meter resolution from satellite imagery is a significant challenge in computer vision with critical environmental applications. 
However, the lack of open-access datasets at this resolution hinders the reproducibility and evaluation of models. We introduce Open-Canopy, the first open-access, country-scale benchmark for very high-resolution (1.5 m) canopy height estimation, covering over 87,000 km² across France with 1.5 m resolution satellite imagery and aerial LiDAR data.  Additionally, we present Open-Canopy-$\Delta$, a benchmark for canopy height change detection between images from different years at tree level---a challenging task for current computer vision models.
 We evaluate state-of-the-art architectures on these benchmarks, highlighting significant challenges and opportunities for improvement. Our datasets and code are publicly available at [URL].
\end{abstract}

\section{Introduction}
Estimating canopy height at high spatial and temporal resolution is crucial for effective and responsive forest management \citep{tomppo2010national,pnfb}, conservation efforts, and policy-making in the face of climate change \citep{fassnacht2024forest,keenan2015climate,mcroberts2007remote}. Meter-level spatial resolutions allow for identifying small vegetation structures, such as individual trees and understory vegetation \citep{kalinicheva2022multi}, and for detecting local disturbances such as selective logging \citep{asner2004canopy,jackson2020remote}. Frequent temporal updates enable the monitoring of rapid changes in forest ecosystems \citep{pretzsch2023forest, das2015improving} caused by activities such as harvesting \citep{yu2004automatic}, illegal logging \citep{kuemmerle2009forest, thompson2021preventing}, or wildfires, storms, and pest outbreaks \citep{huertas2022mapping}. This high granularity is essential for precise biomass estimation \citep{stephenson2014rate}, biodiversity assessments \citep{getzin2012assessing}, and understanding the ecological processes that occur on local scales \citep{lecq2017importance}. 

\begin{figure}[t]
    \centering
\centering

\begin{tabular}{@{\!}c@{\,}c@{\,}c@{\,}}
\begin{subfigure}{0.325\linewidth}
  \begin{tikzpicture}
    \node[anchor=south west,inner sep=0] (image) at (0,0) {\includegraphics[width=\linewidth]{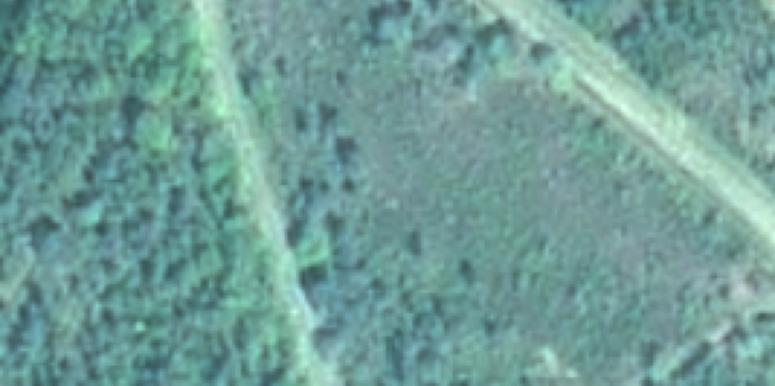}
};
    \begin{scope}[x={(image.south east)},y={(image.north west)}]
      \node[fill=none, draw=none, text=black] (n1) at (0.9,0.89) {\includegraphics[width={0.1}\textwidth]{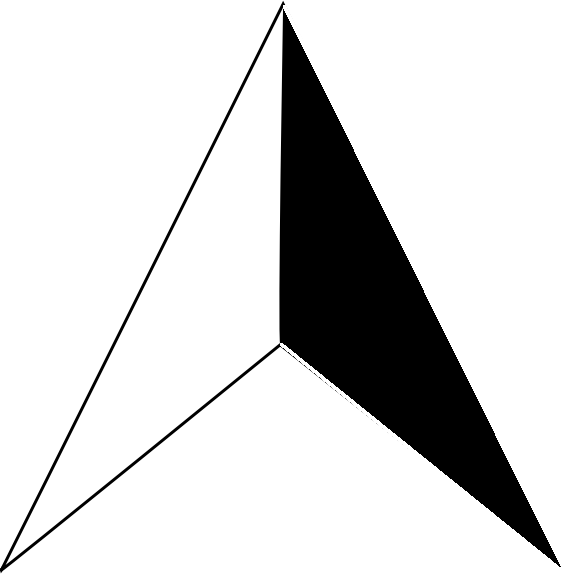}} ;
      \node[fill=none, draw=none, text=black] (n1) at (0.9,0.75) {\contour{white}{\scriptsize N}} ;
       \draw[-, fill=white, text=black, draw=white, ultra thick] (0.72,0.10) -- (0.93,0.10);
      \draw[<->, fill=white, text=black, draw=black,  thick] (0.70,0.10) -- (0.95,0.10);
      \draw[-, draw=black, thick] (0.825,0.10) -- (0.825,0.05);
      \node[fill=none, text=black, draw=none] at (0.825,0.18)  {\tiny  \contour{white}{\bf 50m}};
    \end{scope}
  \end{tikzpicture}
\caption{VHR images\vphantom{$^\dagger$}, 1.5m\\~}
\label{fig:teaser:a}
\end{subfigure}
     & 
\begin{subfigure}{0.325\linewidth}
\includegraphics[width=\linewidth]{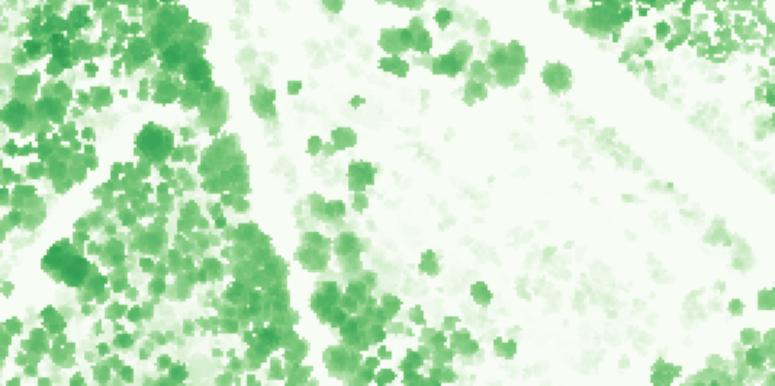}
\caption{ALS height map,\vphantom{$^\dagger$}\\\centering 1.5m}
\label{fig:teaser:b}
\end{subfigure}
     & 
\begin{subfigure}{0.325\linewidth}
\includegraphics[width=\linewidth]{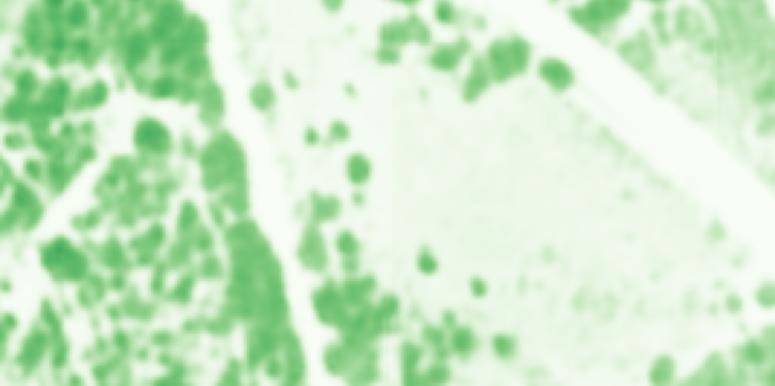}
\caption{Ours\vphantom{$^\dagger$}, 1.5m\\\centering MAE: 3.3m}
\label{fig:teaser:c}
\end{subfigure}
\\
\begin{subfigure}{0.325\linewidth}
\includegraphics[width=\linewidth]{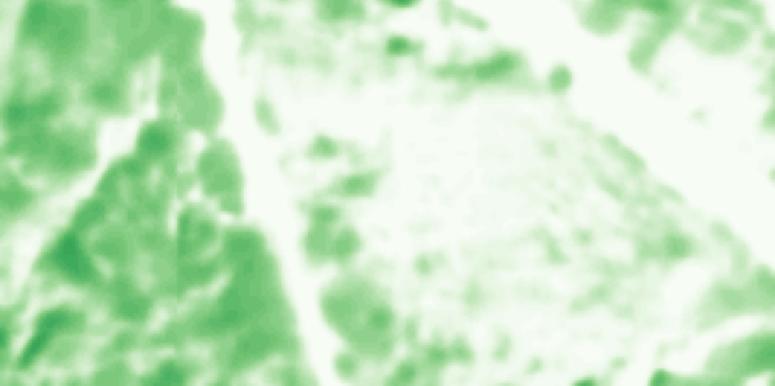}
\caption{Tolan$^\dagger$ \citep{tolan2024very}, 1.2m\\\centering MAE: 4.3m}
\label{fig:teaser:d}
\end{subfigure}
&
\begin{subfigure}{0.325\linewidth}
\includegraphics[width=\linewidth]{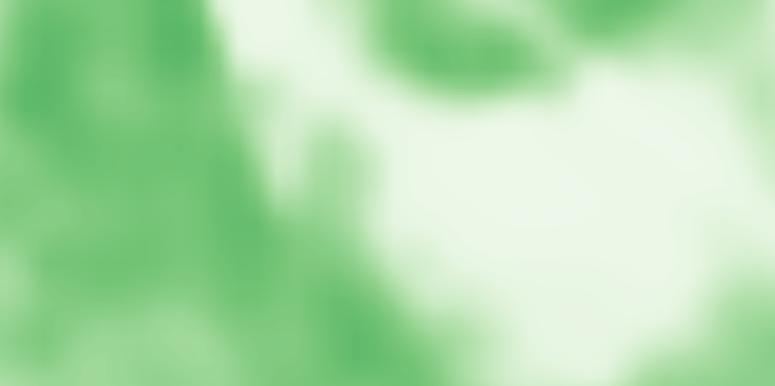}
\caption{Schwartz$^\dagger$ \citep{schwartz2023forms}, 10m \\ \centering MAE: 6.4m}
\label{fig:teaser:e}
\end{subfigure}
&
\begin{subfigure}{0.325\linewidth}
\includegraphics[width=\linewidth]{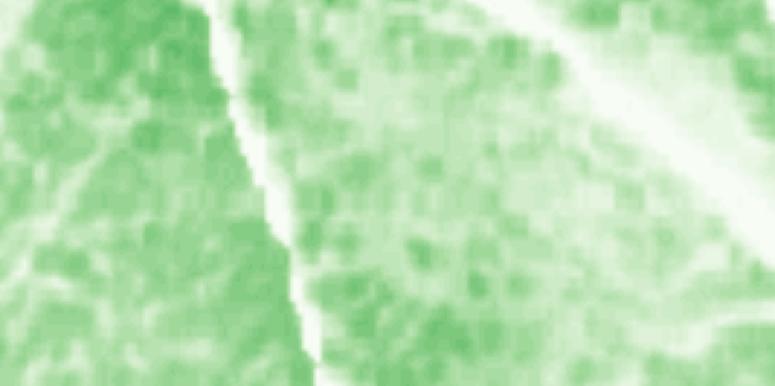}
\caption{Liu$^\dagger$ \citep{liu2023overlooked}, 3m\\\centering MAE: 6.9m}
\label{fig:teaser:f}
\end{subfigure}
\\
\begin{subfigure}{0.325\linewidth}
\includegraphics[width=\linewidth]{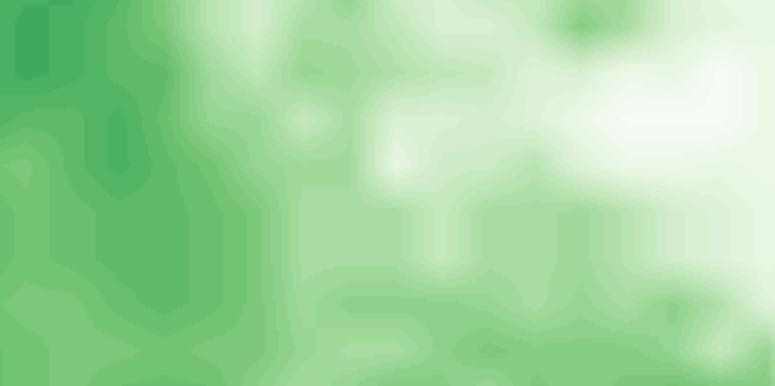}
\caption{Potapov$^\dagger$ \citep{potapov2021mapping}, 30m\\\centering MAE: 8.8m}
\label{fig:teaser:g}
\end{subfigure}
&
\begin{subfigure}{0.325\linewidth}
\includegraphics[width=\linewidth]{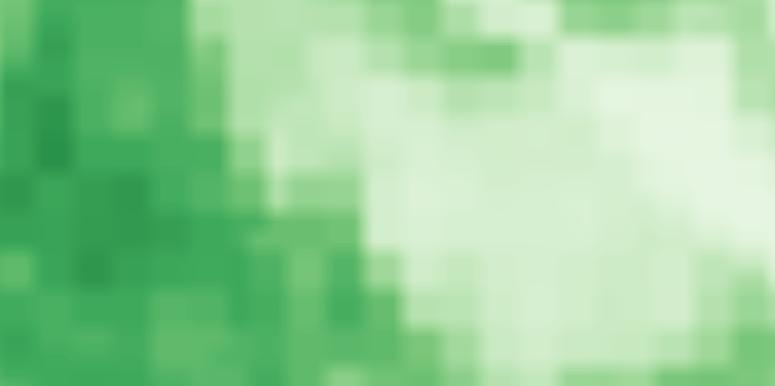}
\caption{Pauls{$^\dagger$} \citep{pauls2024estimating}, 10m\\\centering MAE: 9.0m}
\label{fig:teaser:h}
\end{subfigure}
&
\begin{subfigure}{0.325\linewidth}
\includegraphics[width=\linewidth]{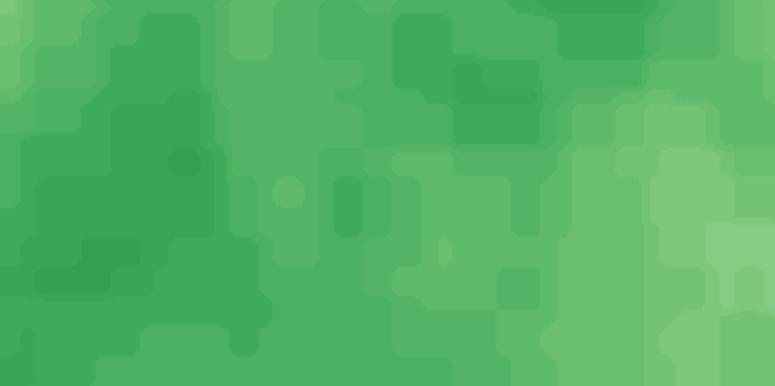}
\caption{Lang\vphantom{$^\dagger$} \citep{lang2023high}, 10m\\\centering MAE: 15.9m}
\label{fig:teaser:i}
\end{subfigure}
\\
\multicolumn{3}{c}{
\multirow{1}{*}[2mm]{
\def\colormapheight{3.65cm}\rotatebox{90}{
\pgfplotsset{
    /pgfplots/colormap={treeheights}{   
        rgb255(0cm)=(15,54,27)      
        rgb255(1cm)=(36,137,69), 
        rgb255(2cm)=(113,198,117), 
        rgb255(3cm)=(199,234,194), 
        rgb255(4cm)=(255,255,255), 
    }
}

\begin{tikzpicture}
    \begin{axis}[
        hide axis,
        scale only axis,
        height=5pt,
        width=50pt,
        colormap name=treeheights,
        colorbar,
           colorbar,
            colorbar style={
            width=.2cm,
            height=\colormapheight,
            ytick={0,1,2,3,4},
            yticklabels={\rotatebox{-90}{40m},\rotatebox{-90}{30m},\rotatebox{-90}{20m},\rotatebox{-90}{10m},\rotatebox{-90}{0m}},
            yticklabel style={font=\tiny},
            major tick length=1.5pt, 
            line width=.05mm,
            grid style={draw=none} 
        },
        point meta min=0,
        point meta max=4
    ]
    \addplot [draw=none] coordinates {(0,0)};
    \end{axis}
 
\end{tikzpicture}}}
}
\end{tabular}
    \caption{{\bf Canopy Height Estimation.} We represent a VHR image from the Open-Canopy test set \subref{fig:teaser:a}, alongside its corresponding ALS-derived canopy height \subref{fig:teaser:b}. We include the height map estimated by a PVPTv2 \citep{wang2021pvtv2} model trained on the Open-Canopy train set \subref{fig:teaser:c}, compared against other canopy height products \subref{fig:teaser:d}-\subref{fig:teaser:i}. For each image, we provide the spatial resolution of the evaluated map and its Mean Absolute Error. $\dagger$ the data or model used to generate these maps is not open-access.}
    \label{fig:teaser}
\end{figure}

Aerial Laser Scanning (ALS) provides precise measurements of canopy height \citep{gaveau2003quantifying}, but its high cost and logistic requirements make frequent data acquisition impractical. In contrast, satellite images are abundant. A cost-effective alternative to yearly ALS acquisition is to train models that estimate canopy height from a single Very High Resolution (VHR) satellite image using existing ALS data for supervision.  
Although recent studies have used VHR satellite images to predict canopy height \citep{tolan2024very, liu2023overlooked, wagner2306sub} and published global height maps, their data and models are often not openly accessible. Many rely on commercial or closed data sources, require substantial pre-processing, and do not always disclose the location of their training sets (see \cref{tab:datasets}).

To enable fair and reproducible evaluation of canopy height prediction models, we introduce Open-Canopy, the first entirely open-access benchmark for VHR canopy height estimation. Spanning over 87,000 km² across France, Open-Canopy combines SPOT 6-7 satellite imagery at 1.5m  resolution with aerial LiDAR data at densities superior to $10$ points/m$^2$, providing a rich dataset for training and evaluating machine learning models. Compared to datasets with similar accessibility \citep{lang2023high}, Open-Canopy improves the resolution of images by a factor of $6$ and the height supervision by a factor of $16$. Our downloadable dataset with ready-to-use training splits and data loaders aims at making forestry research more accessible to computer vision researchers.

\begin{table*}[t]  
\caption{{\bf Canopy Height Prediction Datasets.} We report information about the satellite images used as input, the canopy height used to train and evaluate models, and the number of ground truth (GT) data points (GEDI samples or ALS pixels). We group models by the resolution:  Medium-to-High ($\geq$ HR) and Very High Resolution (VHR).  
}
    \centering
\begin{tabular}{l@{}l}
\begin{tabular}{l@{\;\;}l @{}c@{\;}c@{\;}c @{}p{3mm}@{} lr @{}p{3mm}@{} lr @{}p{3mm}@{}  lrr}
\toprule
&   \multirow{3}{*}{Dataset} 
& \multicolumn{3}{c}{access}
&&  \multicolumn{2}{c}{extent} 
&& \multicolumn{2}{c}{images} 
&& \multicolumn{3}{c}{height ground truth}\\
\cline{3-5}\cline{7-8}\cline{10-11}\cline{13-15}
&& 
\multirow{2}{*}{code} & \multirow{2}{*}{img} & \multirow{2}{*}{GT} 
&& \multirow{2}{*}{scope}  & surface
&& \multirow{2}{*}{sensor} & res. 
&& \multirow{2}{*}{sensor} & res. & samples 
\\
&&
&&&
&&\scriptsize{$\times10^3$ km$^2$}
&&&\scriptsize{in m }& 
&&  \scriptsize{in m} & \scriptsize{$\times10^6$} \\\midrule
\multirow{4}{*}{\rotatebox{90}{$\geq$ HR}}&
 Schwartz \citep{schwartz2023forms}
 & \;\false 
 &\scriptsize{\textcolor{BrickRed}{\faQuestionCircle}}
 & \scriptsize{\textcolor{BrickRed}{\faCog \faQuestionCircle}}
 && France & 588 
 && S1/S2  & 10  
 && GEDI  & 25 & 90  \\
 &Lang \citep{lang2023high}  
 & \;\true 
 & \textcolor{green!70!black}{\faDownload}
 & \textcolor{green!70!black}{\faDownload}
 && Global &  14k
 && S2  & 10 
 && GEDI & 25 & 600 \\
 &Potapov \citep{potapov2021mapping}
 & \;\false  
 & \scriptsize{ \textcolor{BrickRed}{\faQuestionCircle}}  
 & \scriptsize{\textcolor{BrickRed}{\faCog \faQuestionCircle}}
 && Global & 150k 
 && Landsat  & 30
 && GEDI  & 25 & 372\\
 &Pauls \citep{pauls2024estimating}
 & \;\true 
 & \scriptsize{ \textcolor{BrickRed}{\faQuestionCircle}}  
 & \scriptsize{\textcolor{BrickRed}{\faCog \faQuestionCircle}}
 && Global & 2621 
 && S1/S2 & 10
 && GEDI  & 25 & \scriptsize{ \textcolor{BrickRed}{\faQuestionCircle}} 
 \\\greyrule
\multirow{4}{*}{\rotatebox{90}{VHR}}
 &
Tolan \citep{tolan2024very} 
& \;\false 
& \scriptsize{\textcolor{BrickRed}{\faDollarSign}}
& \textcolor{green!70!black}{\faDownload}
&& US & 5.8 
&& MAXAR & 1.2 
&& ALS+GEDI\!\!\!\!\!\!\!\!\!\!\!  & 1 & 5800  \\
& Wagner \citep{wagner2306sub}
& \;\false 
& \scriptsize{\textcolor{BrickRed}{\faLock\faQuestionCircle}} 
& \scriptsize{\textcolor{BrickRed}{\faQuestionCircle}} 
&&  US & 3.8
&& NAIP & 0.6
&& ALS  & 1 & 3784\\ 
& Liu \citep{liu2023overlooked}
& \;\false 
& \scriptsize{\textcolor{BrickRed}{\faLock\faQuestionCircle}}
& \scriptsize{\textcolor{BrickRed}{\faQuestionCircle}}
&& Europe & 700 
&& Planet  & 3 
&& ALS  & 3 & 77,777 \\
& \bf Open-Canopy
& \;\true 
& \textcolor{green!70!black}{\faDownload}
& \textcolor{green!70!black}{\faDownload}
&& France & 87 
&& SPOT 6-7  & 1.5 
&& ALS  & 1.5 & 38,876\\
\bottomrule
\end{tabular}
&
\begin{tabular}{c@{\;}l}
    \multirow{1}{*}{\textcolor{green!70!black}{\faDownload}}  &   $\substack{\text{direct \qquad\;\;\;\;}\\\text{download\qquad}}$\\~\\ 
    \textcolor{BrickRed}{\faDollarSign}  & \scriptsize \raisebox{1mm}{commercial} \\~\\
    \multirow{1}{*}{\scriptsize\textcolor{BrickRed}{\faCog}}  &  $\substack{\text{complex pre-}\\\text{processing\;\;\;\;}}$ \\~\\
    \textcolor{BrickRed}{\scriptsize\faLock}  & $\substack{\text{special }\\\text{access\;\;}}$   \\~\\
    \multirow{1}{*}{\scriptsize\textcolor{BrickRed}{\faQuestionCircle}}  & $\substack{\text{no train\;\;}\\\text{test split }}$   \\~\\
    \scriptsize S1 & \scriptsize Sentinel-1 \\
    \scriptsize S2 & \scriptsize Sentinel-2
\end{tabular}
\end{tabular}
    \label{tab:datasets}
\end{table*}

In contrast to costly ALS measurements, satellite imagery can be acquired every year or less, and enables dynamic estimation of canopy height. This task is a major concern for both forest managers and authorities, as emphasized by recent European regulations \citep{eudr} restricting the import of products related to deforestation.
We introduce Open-Canopy-$\Delta$, a subset of OpenCanopy with two consecutive ALS acquisitions. This data set allows us to formulate and benchmark the challenging problem of detecting dense forest height change, \ie segmenting areas with a significant reduction in canopy height between two VHR satellite images.

Although existing studies \citep{lang2023high, potapov2021mapping, schwartz2023forms, pauls2024estimating,wagner2306sub,liu2023overlooked} predominantly evaluate UNet-type architectures \citep{ronneberger2015u}, predicting canopy height from a single image presents a particularly challenging computer vision task that differs from typical natural image analysis settings. The satellite viewpoint deviates from conventional depth estimation scenarios, the inclusion of the crucial near-infrared band complicates the adaptation of RGB-based foundation models, and capturing the intricate relationships between tree radiometry, phylogeny, and allometry is difficult. In this paper, we evaluate a wide range of modern architectures and models, identify their current limitations, and present this task as a newly accessible, high-impact challenge for the computer vision community.
Our contributions are as follows:
\begin{compactitem}
    \item {\bf Open-Canopy:} An open-access dataset for canopy height estimation from VHR images with ALS annotations.
    \item {\bf Open-Canopy-$\Delta$:} An open-access dataset for segmenting canopy height changes between two images.
    \item {\bf Benchmark:} An evaluation of recent computer vision architectures, foundation models, and forest products for these two tasks.
\end{compactitem}





\section{Related Work}
This section details existing datasets and methods for the problem of canopy height estimation, grouped by annotation type: GEDI or ALS. See \cref{tab:datasets} for an exhaustive comparison of Open-Canopy with these datasets.


\para{GEDI-Based Datasets.}
The Global Ecosystem Dynamics Investigation (GEDI) mission consists of a LiDAR mounted on the ISS and provides global canopy height measurements with a footprint diameter of $25$m \citep{dubayah2020global}. GEDI captures a set of spatially discrete full waveform echoes along paths approximately $4$km wide. 
Models trained with GEDI data use it as a sparse and coarse supervisory signal to predict canopy height from medium to high resolution imagery such as Landsat images at $30$m resolution ({Potapov \etal} \citep{potapov2021mapping}) or Sentinel-2 at $10$m resolution ({Schwartz \etal} \citep{schwartz2023forms}, {Pauls \etal} \citep{pauls2024estimating} and  {Lang \etal} \citep{lang2023high}). However, GEDI's full waveform LiDAR can exhibit registration errors of up to $10$m  \citep{schleich2023improving}.

\para{ALS-Based Datasets.}
Aerial Laser Scanning (ALS) uses low-flying aircraft equipped with LiDAR to create dense 3D point clouds of the Earth's surface. These systems typically capture data at resolutions ranging from 10 to 60 points per square meter. This data is then rasterized along a high resolution grid, and used to estimate canopy height by subtracting the lowest quantiles height (ground surface) from the height of the highest quantiles (top of canopy). This allows the computation of ``true'' height maps which are then used to train models to predict canopy height from VHR images,
 at scales such as $1.2$m for {Tolan \etal} \citep{tolan2024very} and {Wagner \etal} \citep{wagner2306sub}, and $3$m for {Liu \etal} \citep{liu2023overlooked}. Open-Canopy uses ALS data from the LiDAR-HD \citep{lidarhd} program rasterized at $1.5$m.

\para{Canopy Height Estimation Models.}
Most canopy height prediction models employ fully supervised UNets \citep{ronneberger2015u} for their ease of use. The recent work by {Tolan \etal \citep{tolan2024very}} uses a Vision Transformer (ViT) \citep{dosovitskiy2020image} pretrained in a self-supervised fashion \citep{oquab2023dinov2} on 18 million images without ALS height data. 
In this paper, we benchmark a variety of modern deep learning architectures for dense prediction of VHR canopy height from SPOT images, including Unet \citep{ronneberger2015u}, Vision Transformers (ViT) \citep{dosovitskiy2020image}, and their hierarchical variants \citep{liu2021swin,wang2021pvtv2,chu2021twins}. We also explore how their pretraining impacts their ability to adapt from vision-related problems to the completely different task of canopy height estimation.

\para{Canopy Height Change Estimation.}
As forests experience rapid losses \citep{hansen2013high,macdicken2015globalb}, better understanding and monitoring of forest dynamics is critical \citep{macdicken2015global}.
Although existing studies have explored the long-term evolution of forests \citep{qin20233pg,ye2019projecting,peng2022research}, they focus on environmental or phenological variables and low-resolution ($500$m) images \citep{reyer2020profound}.
Models aiming to detect forest changes from images generally operate  at medium or high resolution images($10$-$30$m) \citep{erfanifard2022assessment,decuyper2022continuous,turubanova2023tree}. Some work have explored the use of ALS \citep{yu2006change} or drones \citep{zhang2019forests}, but typically operate over small areas and do not provide open-access data or code.
To the best of our knowledge, Open-Canopy-$\Delta$ is the first open-access VHR benchmark for canopy height change detection with LiDAR-derived ground truth.

\para{Data Access Policies.}
The seven studies in \cref{tab:datasets} all provide open-access predicted canopy height maps and often their trained models. However, only the work of {Lang} \etal provides its code and direct download links for their processed datasets. In contrast, the datasets used by {Tolan} \etal, Wagner \etal, and {Liu} \etal involve commercial satellite imagery or data that requires special access and cannot be redistributed. Although GEDI, Sentinel, and Landsat data are open-access, their preprocessing necessitates substantial expertise \citep{tang2023evaluating,tian2021calibrating}. 
Except for the studies of {Tolan \etal} and {Lang \etal}, these works also do not specify their training and testing splits, complicating their evaluation on external datasets due to potential overlap. Like the study of {Lang} \etal, our data, code, splits, and models are freely available. This transparency is crucial for advancing canopy height estimation as a mainstream application of vision models.

\section{The Open-Canopy Benchmark}
\label{sec:opencanopy}
We introduce Open-Canopy, an open-access country-scale benchmark for estimating canopy height at very high resolution. We first present our dataset  
(\cref{sec:datasetchar}), then the models evaluated (\cref{sec:models}), and finally the results (\cref{sec:models}) and limitations (\cref{sec:limitations}) of the benchmark.

\begin{figure}[t]
    \centering
    \input{figures/dataset}
    \vspace{-2mm}
    \caption{{\bf Open-Canopy.} Our training, validation, and test sets span the French territory and use a $1$km buffer \subref{fig:dataset:a}. We provide VHR images at a 1.5 m resolution \subref{fig:dataset:b} and associated LiDAR-derived canopy height maps \subref{fig:dataset:c}. }
    \label{fig:dataset}
\end{figure}

\subsection{Dataset Characteristics}
\label{sec:datasetchar}
We explain here the main characteristics of the dataset of the Open-Canopy benchmark. We report a detailed description of the dataset construction in the supplementary material.

\para{Why Just France?} 
France offers a unique opportunity for developing an open-access, very high-resolution canopy height benchmark due to recent national initiatives that have made critical data sources publicly available under the open EtaLab2.0 license \citep{etalab}: (i) DINAMIS \citep{dinamis} provides SPOT 6-7 very high-resolution (VHR) satellite imagery covering the entire French territory at a native panchromatic resolution of 1.5 m; and (ii) the LiDAR-HD project \citep{lidarhd} offers extensive airborne 3D point clouds with densities exceeding 10 points per square meter. While other countries provide open-access ALS data \citep{neon,swisstopo} and associated VHR images \citep{naip,swissortho}, they are local solutions. SPOT 6-7 images provide consistent global coverage and make an excellent test bed for estimating the relevance of meter-scale satellite imagery.

Moreover, the French metropolitan territory exhibits a wide range of climates---12 of the 18 Köppen-Geiger climate types found in continental Europe \citep{peel2007updated}, including temperate, Mediterranean, and Alpine environments. The French forest inventory lists 190 distinct tree species \citep{france_trees}. While models trained on the Open-Canopy dataset may not generalize globally, their performance within Europe is likely to be robust given this environmental diversity.

\para{Preprocessing.} We have compiled over 100,000 km$^2$ of data from different providers. Despite advancements in geo libraries and government APIs, downloading, processing, and curating data required manual intervention, the development of custom functions, and significant computation. Deriving the canopy height from the ALS 3D point clouds alone took over 100 hours of continuous computation on a dedicated cluster with 70 CPUs. To facilitate future extensions of OpenCanopy, we provide scripts on our online repository to streamline these processes.

\para{Extent and Splits.} We selected 87,383 tiles across France, each measuring $1\times1$ km$^2$. We divided the dataset into training ($66\,339$km$^2$ ), validation ($7\,369$km$^2$), and test sets ($13\,675$km$^2$). We added a 1 km buffer ($8\,046$km$^2$) between the test split and other splits to avoid data contamination, and ensured a representative distribution of each split among all bioclimatic regions of France \citep{greco}.

\para{VHR Satellite Images.} As illustrated in \cref{fig:dataset}\subref{fig:dataset:b}, we use orthorectified SPOT 6-7 images \citep{smith1995digital,yang2013ortho} from DINAMIS \citep{dinamis} with four spectral bands: red, green, blue, and near-infrared at a resolution of $6$m, and a panchromatic band at a resolution of $1.5$m. We apply pansharpening with the weighted Brovey algorithm \citep{gillespie1987color} to upsample all four spectral bands to a resolution of $1.5$m. We select images from the same year as the corresponding ALS acquisition campaign, in 2021, 2022 and 2023.

\para{ALS-Based Canopy Height.} As depicted in \cref{fig:dataset}\subref{fig:dataset:c}, we use ALS data from the LiDAR-HD project \citep{lidarhd} between 2021 and 2023, which provides a minimum density of $10$ points per m$^2$. The canopy height maps are calculated at the same resolution as the VHR images by taking the maximum height difference between each point and its nearest \emph{ground} point within each pixel. We interpolate ground height when necessary to be robust to very dense canopies.
As described in the supplementary material, we validated the obtained canopy height maps by comparing them at both plot and tree-level with extensive in-situ measurements. 

\para{Vegetation Mask.} 
As illustrated in \cref{fig:mask}, we construct a comprehensive vegetation mask by taking the union of the ALS-derived mask indicating vegetation over $1.5$m in height, with official forest plots outlines, both provided by IGN \citep{lidarhd, link_to_forest_masks}. The resulting vegetation mask, covering 49\% of the dataset, contains trees and shrubs both within forest plots and in other areas such as hedges and urban environments. This offers a more comprehensive evaluation area than the traditionally used forest boundaries \citep{schwartz2023forms}.

\begin{figure}[t]
    \centering
    \input{figures/mask}
    \vspace{-2mm}
    \caption{{\bf Vegetation Mask.} We combine an ALS-derived vegetation mask \subref{fig:mask:a} with official forest outlines \subref{fig:mask:b} to build a pixel-precise mask \subref{fig:mask:c} covering a wide range of vegetation types, as seen in the VHR image \subref{fig:mask:d}. }
    \label{fig:mask}
\end{figure}

\subsection{Evaluated Models}
\label{sec:models}

We evaluate different state-of-the-art computer vision approaches for canopy height estimation from a single VHR satellite image with 4 spectral bands. We list below the selected models and how we adapted them to our task.

\para{Selected Models.}
Given the ubiquity of convolutional models for canopy height estimation, we evaluate the \textbf{UNet}~\citep{ronneberger2015u} and \textbf{DeepLabv3} \citep{chen2017rethinking}  architectures. We select Vision Transformers (\textbf{ViT}) and their convolutional-hybrid variant (\textbf{HViT}) \citep{dosovitskiy2020image}, as they recently became standard in computer vision. We also explore hierarchical ViT architectures such as \textbf{SWIN} \citep{liu2021swin}, \textbf{PCPVT} \citep{chu2021twins}, and \textbf{PVTv2} \citep{wang2021pvtv2}.

To assess the impact of pretraining, we include models pretrained on \textbf{ImageNet} \citep{ridnik2021imagenet,ridnik2021imagenet}, but also large external datasets such as \textbf{DinoV2} \citep{oquab2023dinov2} and \textbf{CLIP-OPENAI} \citep{radford2021learning}. We also consider the \textbf{ScaleMAE} \citep{reed2023scale} model, pretrained on satellite imagery of various resolutions, and Tolan \etal 's model \citep{tolan2024very} for canopy height estimation from RGB images.
 
\para{Adaptating Models.}
We adapt the architecture of the considered models, originally designed for the semantic segmentation of RGB images, to our setting.
To handle the near-infrared channel, we change their input size from $3$ to $4$. We retain the pretrained weights related to RGB, and initialize the near-infrared channel weights with small random values drawn from a normal distribution $\mathcal{N}(0,0.01)$.
We use a transposed convolution as a decoder to predict continuous canopy heights and use the $L_1$ norm as a loss function.

\para{Dataloader and Evaluation.} 
During training, we sample random tiles of size 224$\times$224 pixels with data augmentation: random scaling from $0.5$ to $2$, and rotations $0$, $90$, $180$, or $270^\circ$). For inference, we sample tiles on a regular grid of $112$ pixels, and only keep the center half of each prediction of size 224$\times$224.
We train our model to predict the canopy height for all pixels, which may not correspond to vegetation. However, we only compute the evaluation metrics for pixels within the vegetation mask described in \cref{sec:datasetchar}.

\para{Parameters and Resources.}
We use a batch size of $64$ and the ADAM optimizer \citep{kingma2014adam} with a learning rate of $10^{-3}$, a linear warm-up of $1$ epoch, and a ReduceLROnPlateau scheduler \citep{reduce} with a patience of $1$ and a decay of $0.5$. 
We perform early stopping with a patience of $3$. These hyperparameters were selected by considering their impact on the UNet and ViT models. Reproducing all our experiments requires 1400 GPU-h with A100 GPUs. We estimate our hyperparameters search and initial experiments to 2000 GPU-h.
\begin{figure*}[t] 
    \centering
        \centering
    \begin{tabular}{ccccl}
     \begin{subfigure}{0.21\textwidth}
        \centering
        \begin{tikzpicture}
    \node[anchor=south west,inner sep=0] (image) at (0,0) {     \includegraphics[width=\linewidth]{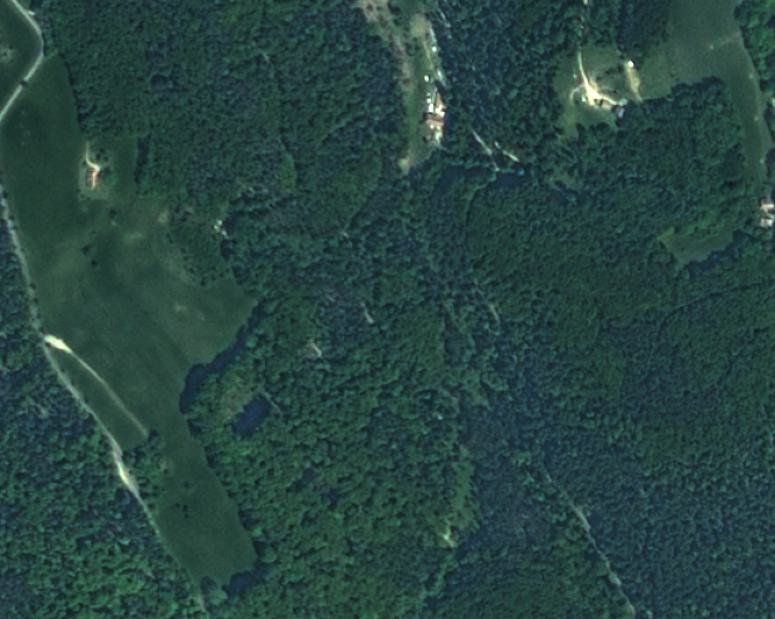}};
     \begin{scope}[x={(image.south east)},y={(image.north west)}]
     \node[fill=none, draw=none, text=white] (n1) at (0.9,0.89) {\includegraphics[width={0.1}\textwidth]{images/north.png}} ;
      \node[fill=none, draw=none, text=white] (n1) at (0.9,0.75) {{\scriptsize N}} ;
       \draw[-, fill=white, text=white, draw=white, ultra thick] (0.72,0.10) -- (0.93,0.10);
      \draw[<->, fill=white, text=white, draw=white,  thick] (0.70,0.10) -- (0.95,0.10);
      \draw[-, draw=white, thick] (0.825,0.10) -- (0.825,0.05);
      \node[fill=none, text=white, draw=none] at (0.825,0.18)  {\tiny  { 250m}};
     \end{scope}
\end{tikzpicture}

        \caption{VHR image}
        \label{fig:vhr}
    \end{subfigure}
    &
    \begin{subfigure}{0.21\textwidth}
        \centering
        \includegraphics[width=\linewidth]{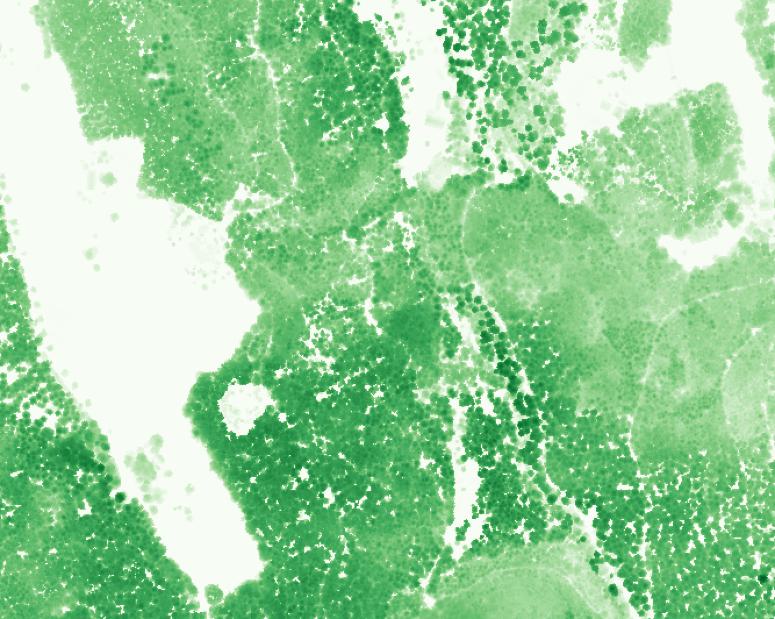}
        \caption{ALS-derived height}
        \label{fig:als}
    \end{subfigure}
    &
    \begin{subfigure}{0.21\textwidth}
        \centering
        \includegraphics[width=\linewidth]{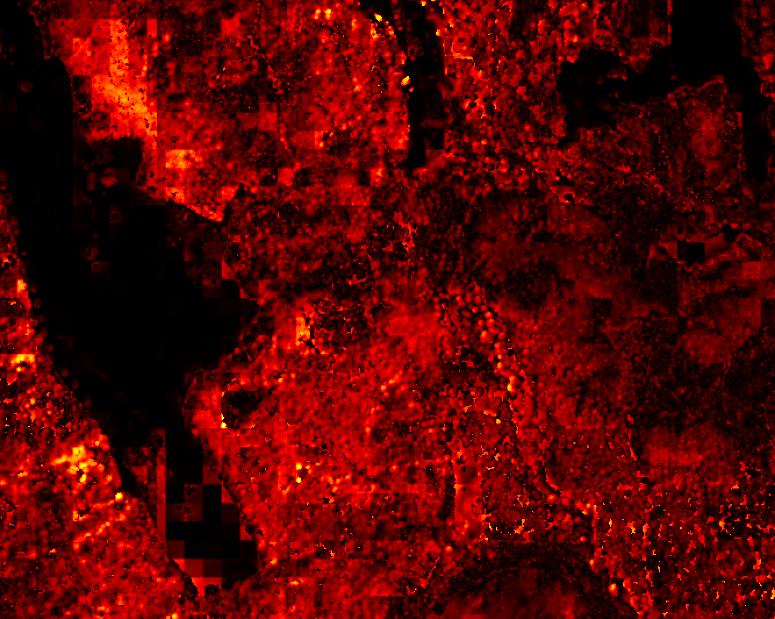}
        \caption{Vit-B Error, MAE= 8.0}
        \label{fig:unet:mae}
    \end{subfigure}
    &
    \begin{subfigure}{0.21\textwidth}
        \centering
        \includegraphics[width=\linewidth]{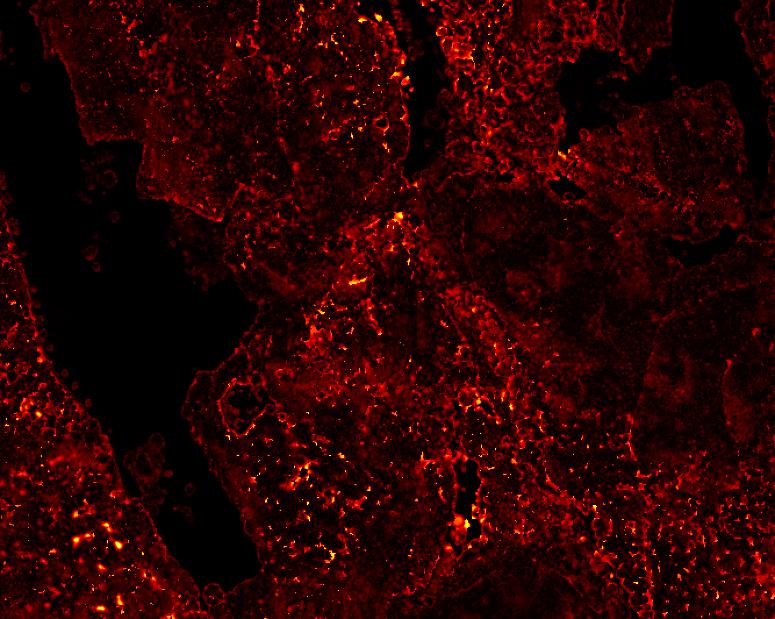}
        \caption{{PVTv2 Error, MAE= 3.9}}
        \label{fig:pvt:mae}
    \end{subfigure}
    \multirow{1}{*}[3.1cm]{
    \hspace{-0.5cm}
    \def\colormapheight{2.51cm}
\pgfplotsset{
    /pgfplots/colormap={heat}{
        rgb255(0cm)=(0,0,0),       
        rgb255(1cm)=(128,0,0),     
        rgb255(2cm)=(255,0,0),     
        rgb255(3cm)=(255,255,0),   
        rgb255(4cm)=(255,255,255)  
    }
}

\begin{tikzpicture}
    \begin{axis}[
        hide axis,
        scale only axis,
        height=5pt,
        width=50pt,
        xlabel={in m},
        colormap name=heat,
        colorbar,
           colorbar,
            colorbar style={
            width=.2cm,
            height=\colormapheight,
            ytick={0,1,2,3,4},
            yticklabels={0m,10m,20m,30m,40m},
            yticklabel style={font=\tiny},
            major tick length=1.5pt, 
            line width=.05mm,
            grid style={draw=none} 
        },
        point meta min=0,
        point meta max=4
    ]
    \addplot [draw=none] coordinates {(0,0)};
    \end{axis}
 
\end{tikzpicture}
    }
    \\
    \end{tabular}
    \vspace{-2mm}
    \caption{{\bf Difference Maps:} Per-pixel absolute (top row) and relative (bottom row) errors for ViT-B and PVTv2. PVT2 predictions are more precise with errors mostly under 10m while Unet mispredict a lot of tree by more than 20m}
    \label{fig:qualitative_small}
\end{figure*}

\subsection{Results and Analysis}
\label{sec:open:results}

We evaluate recent vision models in \cref{tab:quantitative}, as well as the accuracy of existing canopy height maps in \cref{tab:maps_quant}.

\begin{table*}[t]   
    \caption{{\bf Canopy Height Prediction Models.} We benchmark several backbone models for the task of predicting the canopy height of each pixel from a single satellite image. All models are pretrained on vision datasets and fine-tuned on our training set.}
    \label{tab:quantitative}
    \centering
\begin{tabular}{l l m{1cm}m{1cm}m{1cm}m{1cm}c}
\toprule
 \multirow{2}{*}{Model} & \multirow{2}{*}{pretraining} &  {MAE} & {nMAE} &{RMSE} & {Bias} &{Tree cov.}\\
&& in m &  {in \%} &{in m}&{in m} &{IoU in \%} 
  \\\midrule
UNet\footnotemark[4] \citep{ronneberger2015u} & ImageNet1k  \citep{russakovsky2015imagenet}
& 2.67	& 23.8 & \hphantom{1}4.18 & -0.30 & 90.4
\\
\rowcolor{gray!10} DeepLabv3\footnotemark[1] \citep{chen2017rethinking} & ImageNet1k \citep{russakovsky2015imagenet}
& 3.18	& 28.4 & \hphantom{1}4.83 & -0.26 & 88.0
\\ \greyrule
ViT-B\footnotemark[3] \citep{dosovitskiy2020image}& ImageNet21k \citep{chen2017rethinking} 
& 4.26 & 37.8 &	\hphantom{1}6.06 & -0.84 & 86.0
\\
\rowcolor{gray!10} HVIT\footnotemark[3] \citep{dosovitskiy2020image} & ImageNet21k \citep{chen2017rethinking}
& 2.65 & 24.0 &	\hphantom{1}4.18 & -0.13 & 90.2
\\ 
PCPVT\footnotemark[3] \citep{chu2021twins} & ImageNet1k \citep{chen2017rethinking}
& 2.57 & 23.1 & \hphantom{1}4.06 & -0.17 & 90.4
\\
\rowcolor{gray!10} SWIN\footnotemark[3] \citep{liu2021swin} & ImageNet21k \citep{chen2017rethinking}
& 2.54 & \textbf{22.8} &	\hphantom{1}\textbf{4.00} &-0.11	& 90.5
\\
PVTv2\footnotemark[3] \citep{wang2021pvtv2} & ImageNet1k \citep{chen2017rethinking}
& \textbf{2.52} & 22.9 & \hphantom{1}4.02 & \hphantom{-}\textbf{0.00}	& 90.5
\\ \greyrule
\rowcolor{gray!10} ScaleMAE \footnotemark[5] \citep{reed2023scale} & FotM \citep{christie2018functional}
& 3.45 & 31.2 & \hphantom{1}5.13 &	-0,48 & 88,2
\\
ViT-B\footnotemark[3] \citep{dosovitskiy2020image} & DINOv2\citep{oquab2023dinov2} & 4.84	& 43.2	& \hphantom{1}6.68 & -0.48 & 84.8 \\
\rowcolor{gray!10} ViT-B\footnotemark[2] \citep{dosovitskiy2020image}& CLIP\_OPENAI \citep{radford2021learning}
& 2.87 & 25.9 & \hphantom{1}4.43 & -0.07 & 89.7 
\\
ViT-L\footnotemark[6]  \citep{dosovitskiy2020image}& Tolan\citep{tolan2024very}
& 4.46 & 38.9 &	\hphantom{1}6.27 & -1.03 & 85.6
\\
\rowcolor{gray!10} SWIN \footnotemark[3] \citep{liu2021swin}& Satlas-pretrained \citep{10377451} & 2.56 &23.1 &\hphantom{1}4.09 &\hphantom{-}0.02 &\textbf{90.6}
\\
\bottomrule
\end{tabular}\\
     $^1$\scriptsize\url{pytorch.org/vision} \;
     $^2$\scriptsize\url{huggingface.co/laion} \;
     $^3$\scriptsize\url{timm.fast.ai/} \;
     $^4$\scriptsize \url{github.com/qubvel/segmentation_models.pytorch} \;
     $^5$\scriptsize\url{github.com/bair-climate-initiative/scale-mae} \;
     $^6$\scriptsize\url{github.com/facebookresearch/HighResCanopyHeight} \;

%
%
%
%
%
\end{table*}


\para{Metrics.}
We evaluate the performance of canopy height estimation models with five metrics: Root Mean Square Error (\textbf{RMSE}), Mean Absolute Error (\textbf{MAE}), normalized MAE (\textbf{nMAE})---which normalizes the absolute error by the target height, \textbf{Bias}---the error averaged across the test set, and Intersection over Union (IoU) for \textbf{Tree Cover} predictions. The tree cover IoU is calculated by comparing binary maps generated by thresholding both ground truth and predicted height maps at a 2m threshold. All metrics are computed only on pixels within the vegetation mask and with ground truth height below 60m. The nMAE is calculated only for pixels with ground truth heights above 2m.

\begin{table*}[t]
    \caption{{\bf Canopy Height Maps Evaluation.} We evaluate several available canopy height map products on our test set.}
    \label{tab:maps_quant}
    \centering
\begin{tabular}{l lc m{1cm}m{1cm}m{1cm}m{1cm}c}
\toprule
 \multirow{2}{*}{Map} & \multirow{2}{*}{Backbone} & {res.} &  {MAE} & {nMAE} &{RMSE} & {Bias} &{Tree cov.}\\
&& in m &  {in m}&   {in \%} &{in m}&{in m} &{IoU in \%} 
  \\\midrule
  Potapov \citep{potapov2021mapping} & UNet & 30 & 6.27  & 58.1 & \hphantom{1}8.68 & \hphantom{-}1.79& 78.0\\
  \rowcolor{gray!10} Schwartz \citep{schwartz2023forms, schwartz2023mapping}
  & UNet & 10 & 5.17  & 42.7 & \hphantom{1}7.20 & \hphantom{-}3.37 &76.8  \\
  Lang \citep{lang2023high}    & CNN & 10 & 9.22  & 89.5 & 17.14 & \hphantom{-}8.40 & 77.4 \\
  \rowcolor{gray!10} Pauls \citep{pauls2024estimating}
  & UNet & 10 & 6.70  & 58.3 & \hphantom{1}8.65 & \hphantom{-}5.22 &76.8
  \\\greyrule
Liu \citep{liu2023overlooked}     & UNet & 3.0 & 4.83 & 46.6 & \hphantom{1}6.90 & \hphantom{-}1.56& 84.1 \\
\rowcolor{gray!10} Tolan \citep{tolan2024very} & ViT-L   & 1.0 & 5.07 & 43.7 & \hphantom{1}7.15 & -2.95&78.8 \\\greyrule
  Open-Canopy &  UNet &  1.5 & 2.67	& 23.8 & \hphantom{1}4.18 & -0.30 & 90.4 \\
\rowcolor{gray!10} {Open-Canopy} &  {PVTv2} &  {1.5} & \textbf{2.52} & \textbf{22.9} & \hphantom{1}\textbf{4.02} & \textbf{0.00} & \textbf{90.5}\\
\bottomrule
\end{tabular}
\end{table*}

\para{Analysis.} We report the quantitative performance of all evaluated backbones in \cref{tab:quantitative}, and selected illustrations in \cref{fig:qualitative_small}. 
We make the following observations:
\begin{compactitem}
    \item {\bf Impact of Backbones.} Contrary to trends in natural image analysis, convolution-based approaches (UNet, HVIT) outperform ViTs, indicating that convolutions can more efficiently extract relevant local features than linear projections. However, hierarchical ViTs (SWIN, PCPVT, PVTv2) achieve the highest precision, underscoring the multi-scale structure of the task.
    \item {\bf Impact of Pretraining.} Interestingly, models pre-trained on ImageNet (UNet, PVTv2) perform better than foundation models trained on extensive databases of natural images (CLIP, DINO). These models do not generalize well to canopy height estimation, likely due to differences in viewpoint, task specificity, data type, and available spectral bands. Pre-training on satellite images does not either improve performance: a SWIN model pre-trained on the SATLAS dataset achieves similar performance as when pre-trained on ImageNet, while ScaleMAE,  and {Tolan \etal}'s ViT \citep{tolan2024very} do not adapt well to our task. We hypothesize that this is due to the spatial domain shift and the fact that they are trained without the near-infrared channel. See the ablation experiments in the supplementary material for additional analysis. 
    \item {\bf Overall Performance.} 
    The methods assessed in this benchmark exhibit commendable results, achieving tree cover detection with over 90\% IoU and an nMAE of around 20\% for the best-performing models.  However, we argue that there exists a significant margin of improvement, in particular for transferring foundation models to our setting.  
\end{compactitem}

\para{Impact of Initialization.} We report in \cref{tab:init} the performance of PVTv2 trained on ImageNet1k, our best performing model, when fine-tuned with different initialization strategies to accommodate the fourth near-infrared channel. Fully random initialization leads to poor performance, which shows that Open-Canopy is not large enough to train a ViT from scratch. LoRA \citep{hu2021lora} adaptation adapts better to the new channel, but fine-tuning all weights with a random first layer leads to significantly better results. Our proposed initialization scheme further improves the results by allowing the network to gradually accommodate the new channel. 

\para{Comparison with Existing Maps.} 
In \cref{tab:maps_quant}, we evaluate the precision of canopy height maps generated by UNet and PVTv2 networks trained on the Open-Canopy dataset against those from other research that we interpolate to a resolution of $1.5$m per pixel. With the caveats on the fairness of the comparison mentioned in \cref{sec:limitations}, our maps achieve significantly better precision. The low performance of models derived from low-resolution imagery is expected, as they are trained to estimate tree height at a different resolution. Among the ALS-based methods, Liu \etal's model performs best, likely due to its training data from Europe, which differs from Tolan et al.'s training in the continental US. Moreover, the Tolan \etal model relies solely on RGB data, while the inclusion of near-infrared is proven to be highly discriminative for vegetation analysis \citep{carlson1997relation}.
In \cref{fig:violins}, we report error plots across various vegetation height bins, highlighting that the PVTv2 model trained on Open-Canopy exhibits significantly lower bias and superior performance, especially in areas with tall trees.

\para{Out-of-domain Evaluation.} 
To assess the spatial generalization of models trained on Open-Canopy, we collected SPOT 6-7 satellite imagery (with DINAMIS \citep{dinamis}) and aerial VHR images (through NAIP \citep{naip}) for a 30km$^2$ area in Utah, United States. We used as ground truth the ALS-based canopy height map provided by NEON on site REDB \citep{neon}.
As detailed in \cref{tab:us_comparison}, a PVTv2 model trained on Open-Canopy and applied to the SPOT image achieves performance comparable to Tolan \etal's height map derived from MAXAR $0.6$m imagery \citep{tolan2024very}, despite their model being predominantly trained on data from the continental US. This demonstrates the robustness of models trained on Open-Canopy to evaluation outside of France. 

We  resampled NAIP aerial images to $1.5$m resolution and normalized them with histogram matching to the global spectral distribution of the entire Open-Canopy dataset. Evaluated on these images, the performance of the PVTv2 model decreases starkly, highlighting its dependency on SPOT data.

\begin{table}
    \centering
        \caption{{\bf Initialisation Strategy.} We evaluate different training strategy for a PVTv2 model trained on ImageNet1k.}
    \label{tab:init}
    \begin{tabular}{lll cccc}
\toprule
 \multirow{2}{*}{Initialization} & MAE 	&nMAE 	&RMSE 	&Bias
  \\
  & in m & in \% & in m & in m
  \\
  \midrule
\rowcolor{gray!10} {Fully random} & 11.17    &85.77   & 14.38 & -10.94 	\\
{LoRA (rank 4)}  & 4.54& 40.79& 6.42& -0.37  \\
\rowcolor{gray!10} {Rand. 1st layer}                   & 2.87& 24.3& 4.24& -0.04 	\\
 {Proposed}  &   \bf 2.52& \bf 22.9& \bf 4.02& \bf 0.00 
         \\\bottomrule     
\end{tabular}

\end{table}


\begin{table*}[t]
    \caption{{\bf Out-of-Domain Evaluation.} We evaluate different models on a 30km$^2$ area in Utah, US. We compare the height map of Tolan \etal\citep{tolan2024very} to a PVTv2 model trained on Open Canopy with SPOT6-7 data or NAIP images.}
    \label{tab:us_comparison}
    \centering
\begin{tabular}{l llc m{1cm}m{1cm}m{1cm}c}
\toprule
 \multirow{2}{*}{} & \multirow{2}{*}{Input data} & \multirow{1}{*}{Training} &  {MAE} & {nMAE} &{RMSE} & {Bias} &{Tree cov.}\\
&&area &  {in m}&   {in \%} &{in m}&{in m} &{IoU in \%} 
  \\\midrule
  Tolan Map \citep{tolan2024very}
  & MAXAR & US & \textbf{2.02}  & 47.4 & 3.58 & \hphantom{-}\textbf{0.57} & \textbf{70.5}
  \\\greyrule
  \rowcolor{gray!10} PVTv2
  & NAIP & OpenCanopy &4.38 & 71.0 & 6.42 & \hphantom{-}2.78& 49.6 	\\
  PVTv2
  & SPOT 6-7 & OpenCanopy & 2.08 & \textbf{33.9} & \textbf{3.20} & \hphantom{-}0.90 & 61.8 \\
\bottomrule
\end{tabular}
\end{table*}

\begin{figure*}
    \centering
\definecolor{ACOLOR}{RGB}{30, 150, 252}
\definecolor{BCOLOR}{RGB}{144, 190, 109}
\definecolor{CCOLOR}{RGB}{249, 65, 68}
\definecolor{DCOLOR}{RGB}{227, 85, 255}

\tikzset{
    STYLE/.style={
        color=#1,
        fill=#1, 
        fill opacity = 0.5,
        very thick,
        solid,
    }
}

\begin{tikzpicture}

\begin{axis}[
    ybar,
    bar width=50pt,
    width=.95\linewidth,
    height=6cm,
    axis y line*=right,
    axis x line=none,
    xmin= 0, xmax=35,
    ymin=-30, ymax=30,
    log ticks with fixed point,
    ytick={0,10,20,30},
    legend style={at={(1,1)},anchor=north east},
]

\addplot[mark=none, fill=black!10, draw=black!30, thick] coordinates {
    (2.5,23.2)
    (7.5,8.8)
    (12.5,14.4)
    (17.5,15.8)
    (22.5,16.4)
    (27.5,19.2)
    (32.5,2.2)
};

\addlegendentry{\footnotesize Proportion (in \%)}

\end{axis}

\begin{axis}[
    xtick={2.5,7.5,12.5,17.5,22.5,27.5,32.5}, 
    xticklabels={0-2m,2-5m, 5-10m, 10-15m, 15-20m,20-30m,30-60m}, 
    ylabel={Error (in m)},
    xlabel={Ground truth height},
    boxplot/draw direction=y,
    width=.95\linewidth,
    height=6cm,
    xmin= 0, xmax = 35,
    ymin = -30, ymax = 30,
    legend style={at={(0,0.0)},anchor=south west},
    legend cell align={left},
    y label style={at={(axis description cs:-0.035,.5)},rotate=0,anchor=south},
]

\addlegendimage{ACOLOR, ultra thick}
\addlegendentry{\footnotesize \bf Open-Canopy (1.5m)}

\addlegendimage{BCOLOR, ultra thick}
\addlegendentry{\footnotesize Tolan (1m)}

\addlegendimage{CCOLOR, ultra thick}
\addlegendentry{\footnotesize Liu (3m)}

\addlegendimage{DCOLOR, ultra thick}
\addlegendentry{\footnotesize Schwartz (10m)}

\addplot[mark=none, black!20] coordinates {
    (0,40) (40,40)
};
\addplot[mark=none, black!20] coordinates {
    (0,30) (40,30)
};
\addplot[mark=none, black!20] coordinates {
    (0,20) (40,20)
};
\addplot[mark=none, black!20] coordinates {
    (0,10) (40,10)
};
\addplot[mark=none, black, ultra thick] coordinates {
    (0,0) (40,0)
};
\addplot[mark=none, black!20] coordinates {
    (0,-10) (40,-10)
};
\addplot[mark=none, black!20] coordinates {
    (0,-20) (40,-20)
};
\addplot[mark=none, black!20] coordinates {
    (0,-30) (40,-30)
};

\addplot+[
    boxplot prepared={
        lower whisker=-1.26, lower quartile=-0.00,
        median=0.03, upper quartile=0.84,
        upper whisker=2.09
    },
    STYLE=ACOLOR,
    boxplot/draw position=1
] coordinates {};

\addplot+[
    boxplot prepared={
        lower whisker=-5.84, lower quartile=-1.24,
        median=-0.00, upper quartile=1.82,
        upper whisker=6.41
    },
    STYLE=ACOLOR,
    boxplot/draw position=6
] coordinates {};

\addplot+[
    boxplot prepared={
        lower whisker=-7.43, lower quartile=-1.57,
        median=0.22, upper quartile=2.34,
        upper whisker=8.20
    },
    STYLE=ACOLOR,
    boxplot/draw position=11
] coordinates {};

\addplot+[
    boxplot prepared={
        lower whisker=-8.07, lower quartile=-1.86,
        median=0.17, upper quartile=2.28,
        upper whisker=8.49
    },
    STYLE=ACOLOR,
    boxplot/draw position=16
] coordinates {};

\addplot+[
    boxplot prepared={
        lower whisker=-8.26, lower quartile=-2.27,
        median=-0.23, upper quartile=1.72,
        upper whisker=7.70
    },
    STYLE=ACOLOR,
    boxplot/draw position=21
] coordinates {};

\addplot+[
    boxplot prepared={
        lower whisker=-9.83, lower quartile=-3.68,
        median=-1.52, upper quartile=0.42,
        upper whisker=6.56
    },
    STYLE=ACOLOR,
    boxplot/draw position=26
] coordinates {};

\addplot+[
    boxplot prepared={
        lower whisker=-14.65, lower quartile=-7.12,
        median=-4.34, upper quartile=-2.10,
        upper whisker=5.43
    },
    STYLE=ACOLOR,
    boxplot/draw position=31
] coordinates {};

\addplot+[
    boxplot prepared={
        lower whisker=-0.50, lower quartile=-0.20,
        median=0.00, upper quartile=0.00,
        upper whisker=0.30
    },
    STYLE=BCOLOR,
    boxplot/draw position=2
] coordinates {};

\addplot+[
    boxplot prepared={
        lower whisker=-4.90, lower quartile=-3.28,
        median=-2.32, upper quartile=0.02,
        upper whisker=4.98
    },
    STYLE=BCOLOR,
    boxplot/draw position=7
] coordinates {};

\addplot+[
    boxplot prepared={
        lower whisker=-9.90, lower quartile=-5.80,
        median=-2.97, upper quartile=1.42,
        upper whisker=12.24
    },
    STYLE=BCOLOR,
    boxplot/draw position=12
] coordinates {};

\addplot+[
    boxplot prepared={
        lower whisker=-14.90, lower quartile=-7.29,
        median=-2.97, upper quartile=1.03,
        upper whisker=13.51
    },
    STYLE=BCOLOR,
    boxplot/draw position=17
] coordinates {};

\addplot+[
    boxplot prepared={
        lower whisker=-19.14, lower quartile=-7.99,
        median=-3.87, upper quartile=-0.55,
        upper whisker=10.60
    },
    STYLE=BCOLOR,
    boxplot/draw position=22
] coordinates {};

\addplot+[
    boxplot prepared={
        lower whisker=-21.20, lower quartile=-10.74,
        median=-6.90, upper quartile=-3.77,
        upper whisker=6.69
    },
    STYLE=BCOLOR,
    boxplot/draw position=27
] coordinates {};

\addplot+[
    boxplot prepared={
        lower whisker=-30.25, lower quartile=-18.43,
        median=-13.95, upper quartile=-10.56,
        upper whisker=1.26
    },
    STYLE=BCOLOR,
    boxplot/draw position=32
] coordinates {};

\addplot+[
    boxplot prepared={
        lower whisker=-1.90, lower quartile=0.00,
        median=0.00, upper quartile=6.84,
        upper whisker=17.10
    },
    STYLE=CCOLOR,
    boxplot/draw position=3
] coordinates {};

\addplot+[
    boxplot prepared={
        lower whisker=-4.90, lower quartile=-1.94,
        median=3.31, upper quartile=8.19,
        upper whisker=23.38
    },
    STYLE=CCOLOR,
    boxplot/draw position=8
] coordinates {};

\addplot+[
    boxplot prepared={
        lower whisker=-9.90, lower quartile=-0.61,
        median=3.33, upper quartile=7.72,
        upper whisker=20.22
    },
    STYLE=CCOLOR,
    boxplot/draw position=13
] coordinates {};

\addplot+[
    boxplot prepared={
        lower whisker=-12.91, lower quartile=-1.61,
        median=2.05, upper quartile=5.92,
        upper whisker=17.21
    },
    STYLE=CCOLOR,
    boxplot/draw position=18
] coordinates {};

\addplot+[
    boxplot prepared={
        lower whisker=-13.00, lower quartile=-2.95,
        median=0.42, upper quartile=3.76,
        upper whisker=13.81
    },
    STYLE=CCOLOR,
    boxplot/draw position=23
] coordinates {};

\addplot+[
    boxplot prepared={
        lower whisker=-14.42, lower quartile=-5.43,
        median=-2.38, upper quartile=0.57,
        upper whisker=9.56
    },
    STYLE=CCOLOR,
    boxplot/draw position=28
] coordinates {};

\addplot+[
    boxplot prepared={
        lower whisker=-19.63, lower quartile=-10.63,
        median=-7.43, upper quartile=-4.63,
        upper whisker=4.36
    },
    STYLE=CCOLOR,
    boxplot/draw position=33
] coordinates {};

\addplot+[
    boxplot prepared={
        lower whisker=0.03, lower quartile=3.89,
        median=6.50, upper quartile=11.55,
        upper whisker=23.05
    },
    STYLE=DCOLOR,
    boxplot/draw position=4
] coordinates {};

\addplot+[
    boxplot prepared={
        lower whisker=-2.89, lower quartile=1.69,
        median=4.57, upper quartile=9.29,
        upper whisker=20.68
    },
    STYLE=DCOLOR,
    boxplot/draw position=9
] coordinates {};

\addplot+[
    boxplot prepared={
        lower whisker=-7.93, lower quartile=0.37,
        median=3.48, upper quartile=7.62,
        upper whisker=18.50
    },
    STYLE=DCOLOR,
    boxplot/draw position=14
] coordinates {};

\addplot+[
    boxplot prepared={
        lower whisker=-9.54, lower quartile=-0.66,
        median=2.14, upper quartile=5.25,
        upper whisker=14.12
    },
    STYLE=DCOLOR,
    boxplot/draw position=19
] coordinates {};

\addplot+[
    boxplot prepared={
        lower whisker=-8.67, lower quartile=-1.52,
        median=0.80, upper quartile=3.25,
        upper whisker=10.41
    },
    STYLE=DCOLOR,
    boxplot/draw position=24
] coordinates {};

\addplot+[
    boxplot prepared={
        lower whisker=-10.11, lower quartile=-3.44,
        median=-1.20, upper quartile=1.01,
        upper whisker=7.67
    },
    STYLE=DCOLOR,
    boxplot/draw position=29
] coordinates {};

\addplot+[
    boxplot prepared={
        lower whisker=-15.22, lower quartile=-7.60,
        median=-4.83, upper quartile=-2.52,
        upper whisker=5.10
    },
    STYLE=DCOLOR,
    boxplot/draw position=34
] coordinates {};

\end{axis}

\end{tikzpicture}
    \vspace{-2mm}
    \caption{{\bf Distribution of Error.} We plot the distribution of errors according to the ground truth canopy height for a PVTv2 model trained on Open-Canopy and different canopy height map products.}
    \label{fig:violins}
\end{figure*}

\subsection{Limitations}
\label{sec:limitations}
While Open-Canopy represents a significant advancement in providing an open-access VHR benchmark for canopy height estimation, it has several limitations.
\begin{compactitem}
 \item \textbf{Constraints of Open-Access:}
    Only using freely distributable sources limits the extent as well as the spatial and temporal resolution of available data. Acquiring and distributing ALS and VHR satellite images at large scale is cost-prohibitive: Open-Canopy would cost millions of dollars to reproduce with commercial data.
    \item \textbf{Geographic Scope:} Although metropolitan France offers a unique combination of open-access data and diverse landscapes, it lacks critical forest types such as rainforests. This absence may affect the generalizability of models trained on Open-Canopy. We hope that this work will inspire similar open-access initiatives in other countries, leading to the creation of a truly global VHR canopy height dataset.
    \item \textbf{Limits of ALS: }     
     Our ground-truth canopy heights are derived from aerial LiDAR measurements, which can contain errors due to factors like multiple echoes. Although we estimate these errors with in-situ measurements (see the supplementary material), they cannot be entirely eliminated. Additionally, while the freely distributable SPOT 6-7 images are captured during spring and summer, the LiDAR measurements span all seasons. This temporal mismatch may encourage models to predict average heights rather than capturing seasonal variations.
    \item \textbf{Comparison with Other Products.}  Our evaluation of other canopy height maps is subject to limitations: (i) unknown training sets for some models might lead to data contamination, (ii) interpolation might distort results, (iii) forest losses and gains happening between the timing of images and ALS acquisitions can affect performance, (iv) maps derived at lower resolution are trained to predict the maximum canopy height in larger pixels, which bias them to higher values. To address these issues, we provide additional experiments in the supplementary material and openly release all related data to facilitate more accurate future evaluations.
\end{compactitem}

\if 10
\begin{figure*}[t]
    \centering
\centering
\begin{tabular}{c@{\,}c@{\,}c@{\,}c@{}c}
\begin{subfigure}{0.225\linewidth}
\includegraphics[width=\textwidth]{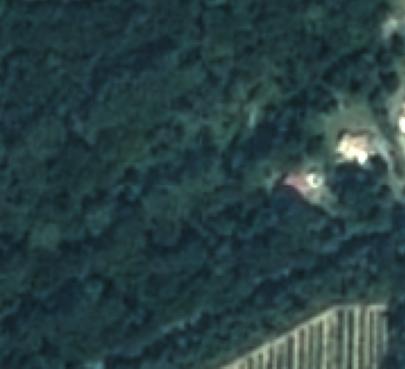}
\caption{VHR year XXX}
\label{fig:growth:a}
\end{subfigure}
&
\begin{subfigure}{0.225\linewidth}
\includegraphics[width=\textwidth]{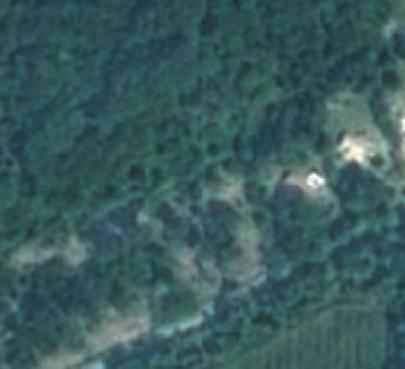}
\caption{VHR year XXX}
\label{fig:growth:b}
\end{subfigure}
&
\begin{subfigure}{0.225\linewidth}
\includegraphics[width=\textwidth]{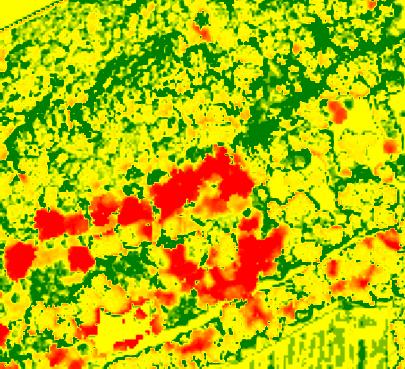}
\caption{growth map}
\label{fig:growth:c}
\end{subfigure}
&
\begin{subfigure}{0.225\linewidth}
\includegraphics[width=\textwidth]{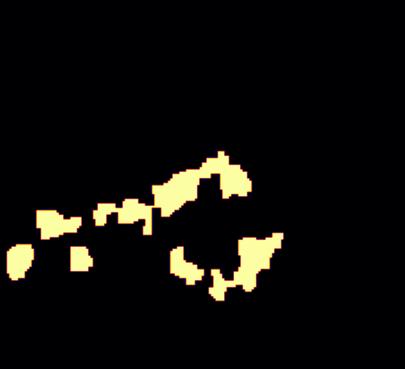}
\caption{disturbance map}
\label{fig:growth:d}
\end{subfigure}
&
\multirow[t]{1}{*}[+0.36cm]{
\hspace{-0.3cm}
\def\colormapheight{2.85cm}
\pgfplotsset{
    /pgfplots/colormap={treegrowth}{
    rgb255(0cm)=(255,0,0), 
    rgb255(1cm)=(255,125,111), 
    rgb255(2cm)=(255,255,0), 
    rgb255(3cm)=(130,192,0), 
    rgb255(4cm)=(0,130,0), 
    }
}

\begin{tikzpicture}
    \begin{axis}[
        hide axis,
        scale only axis,
        height=5pt,
        width=50pt,
        xlabel={in m},
        colormap name=treegrowth,
        colorbar,
            colorbar style={
            width=.2cm,
            height=\colormapheight,
            ytick={0,0.5,1,1.5,2,3,4},
            yticklabels={-20m,-15m,-10m,-5m,0m,+1m,+2m},
            yticklabel style={font=\tiny},
            major tick length=1.5pt,
            line width=.05mm,
            grid style={draw=none} 
        },
        point meta min=0,
        point meta max=4,
    ]
    \addplot [draw=none] coordinates {(0,0)};
    \end{axis}
 
\end{tikzpicture}
}
\end{tabular}
    \caption{{\bf Open-Canopy-$\Delta$.} We represent two VHR images taken in consecutive years \subref{fig:growth:a} and \subref{fig:growth:b} alongside ALS-based estimations of vegetation change and decline \subref{fig:growth:c}. Additionally, we include a binary disturbance map that highlights areas with significant vegetation loss.}
    \label{fig:change}
\end{figure*}
\fi

\section{Open-Canopy-$\Delta$}
\label{sec:opencanopydelta}
We present Open-Canopy-$\Delta$, a benchmark for canopy height change detection between consecutive VHR images. We describe the dataset in \cref{sec:delta:data} and our results in \cref{sec:delta:results}.

\subsection{Dataset Characteristics}
\label{sec:delta:data}

\para{Extent and Context.} Open-Canopy-$\Delta$ focuses on the Forêt de Chantilly, a declining forest due to climate change and of high concern for conservationists \citep{chantilly}. We consider two ALS acquisitions in February 2022 \citep{lidarhd} and September 2023 \citep{chantilly}, allowing us to build two consecutive canopy height maps and collect corresponding SPOT 6-7 satellite images. The studied area spans $16,634$ hectares and is strictly comprised in the test set of Open-Canopy, \emph{not} overlaping with the training set.

\begin{figure*}[t]
    \centering
\centering
\begin{tabular}{c@{\,}c@{\,}c@{\,}c@{\,}c@{\,}c}
\begin{subfigure}{.18\textwidth}
  \begin{tikzpicture}
    \node[anchor=south west,inner sep=0] (image) at (0,0) {\includegraphics[width=\textwidth]{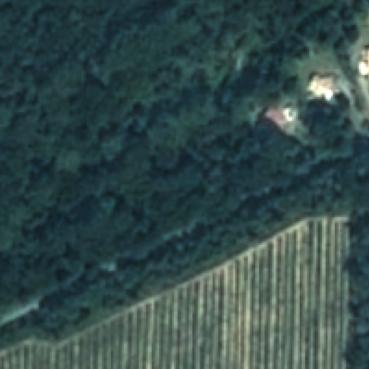}
    };
     \begin{scope}[x={(image.south east)},y={(image.north west)}]
     \node[fill=none, draw=none, text=black] (n1) at (0.145,0.89) {\includegraphics[width={0.1}\textwidth]{images/north.png}} ;
      \node[fill=none, draw=none, text=white] (n1) at (0.145,0.75) {{\scriptsize \bf N}} ;
      \draw[<->, fill=white, text=black, draw=white,  thick] (0.05,0.10) -- (0.24,0.10);
      \node[fill=none, text=white, draw=none] at (0.145,0.18)  {\scriptsize  {\bf 50m}};
     \end{scope}
\end{tikzpicture}
\caption{\scriptsize VHR year 2022 }
\label{fig:quali-dynamic:a}
\end{subfigure}
&
\begin{subfigure}{.18\textwidth}
\includegraphics[width=\textwidth]{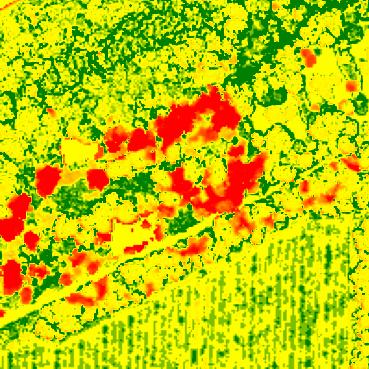}
\caption{\scriptsize  ALS change map }
\label{fig:quali-dynamic:b}
\end{subfigure}
&
\begin{subfigure}{.18\textwidth}
\includegraphics[width=\textwidth]{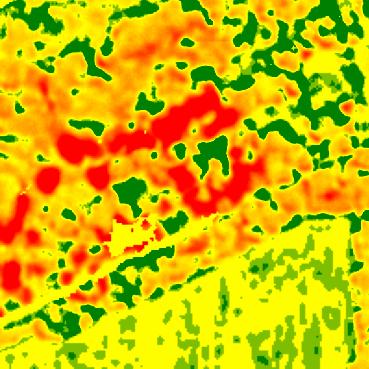}
\caption{\scriptsize {\bf Our change map}  }
\label{fig:quali-dynamic:g}
\end{subfigure}
&
\begin{subfigure}{.18\textwidth}
\includegraphics[width=\textwidth]{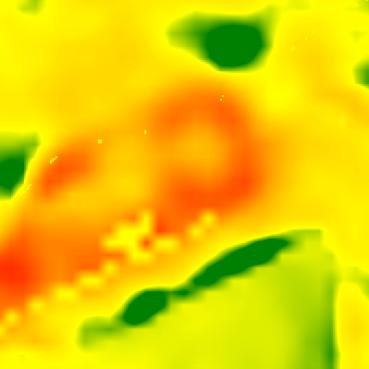}
\caption{\scriptsize Sentinel change map}
\label{fig:quali-dynamic:c}
\end{subfigure}
&
\begin{subfigure}{.18\textwidth}
\includegraphics[width=\textwidth]{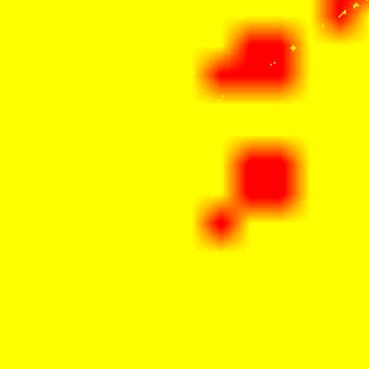}
\caption{\scriptsize GFC change map}
\label{fig:quali-dynamic:h}
\end{subfigure}
\\
\begin{subfigure}{.18\textwidth}
\includegraphics[width=\textwidth]{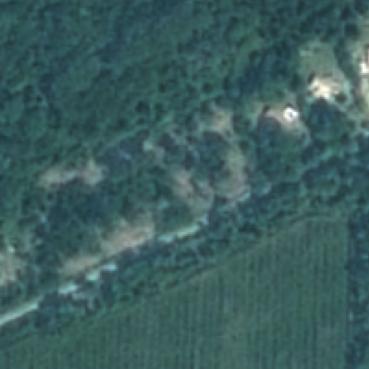}
\caption{\scriptsize VHR year 2023 }
\label{fig:quali-dynamic:f}
\end{subfigure}
&
\begin{subfigure}{.18\textwidth}
\includegraphics[width=\textwidth]{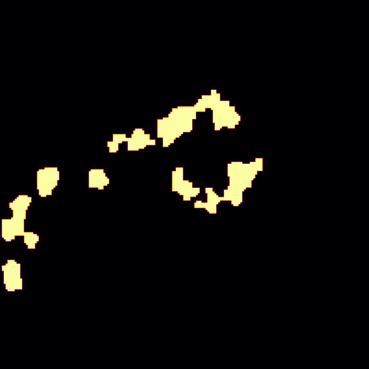}
\caption{\scriptsize ALS change mask }
\label{fig:quali-dynamic:d}
\end{subfigure}
&
\begin{subfigure}{.18\textwidth}
\includegraphics[width=\textwidth]{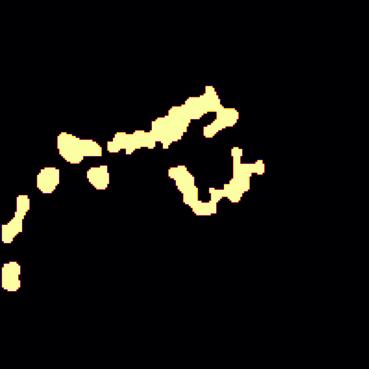}
\caption{\scriptsize \bf Our change mask}
\label{fig:quali-dynamic:i}
\end{subfigure}
&
\begin{subfigure}{.18\textwidth}
\includegraphics[width=\textwidth]{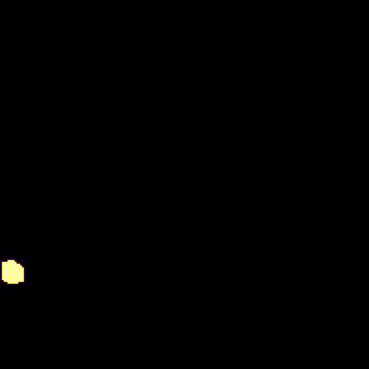}
\caption{\scriptsize {Sentinel change mask}}
\label{fig:quali-dynamic:e}
\end{subfigure}
&
\begin{subfigure}{.18\textwidth}
\includegraphics[width=\textwidth]{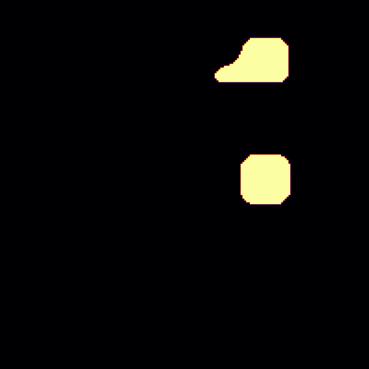}
\caption{\scriptsize GFC change mask}
\label{fig:quali-dynamic:j}
\end{subfigure}
&
\multirow[t]{2}{*}[1.5cm]{
\hspace{-0.5cm}
\def\colormapheight{3.85cm}
\pgfplotsset{
    /pgfplots/colormap={treegrowth}{
    rgb255(0cm)=(255,0,0), 
    rgb255(1cm)=(255,125,111), 
    rgb255(2cm)=(255,255,0), 
    rgb255(3cm)=(130,192,0), 
    rgb255(4cm)=(0,130,0), 
    }
}

\begin{tikzpicture}
    \begin{axis}[
        hide axis,
        scale only axis,
        height=5pt,
        width=50pt,
        xlabel={in m},
        colormap name=treegrowth,
        colorbar,
            colorbar style={
            width=.2cm,
            height=\colormapheight,
            ytick={0,0.5,1,1.5,2,3,4},
            yticklabels={-20m,-15m,-10m,-5m,0m,+1m,+2m},
            yticklabel style={font=\tiny},
            major tick length=1.5pt,
            line width=.05mm,
            grid style={draw=none} 
        },
        point meta min=0,
        point meta max=4,
    ]
    \addplot [draw=none] coordinates {(0,0)};
    \end{axis}
 
\end{tikzpicture}
}
\end{tabular}
    \vspace{-2mm}
    \caption{{\bf Canopy Height Change.} 
    We consider VHR images taken in the Chantilly Forest taken in 2022 \subref{fig:quali-dynamic:a} and 2023 \subref{fig:quali-dynamic:f}, and use ALS observations of the same years to derive a canopy height change map \subref{fig:quali-dynamic:b}. We represent the change map predicted by a PVTv2 model \subref{fig:quali-dynamic:g} and two competing approaches: Sentinel-derived maps from Schwartz \etal \citep{schwartz2023forms} \subref{fig:quali-dynamic:c} and Global Forest Change \citep{hansen2013high} \subref{fig:quali-dynamic:g}. Finally, we compare the binary change masks derived from ALS measurements \subref{fig:quali-dynamic:d} and the predicted change maps \subref{fig:quali-dynamic:e},\subref{fig:quali-dynamic:i},\subref{fig:quali-dynamic:j}.
    }
    \label{fig:qualitative-dynamic}
\end{figure*}

\para{Processing.} We generated a rasterized canopy change map by subtracting the ALS-based height map of 2022 from the map of 2023. Significant decreases in canopy height can result from various forest disturbance events such as fires, logging, diebacks, or maintenance activities. However, minor changes due to seasonal growth cycles, wind, or sensor errors can introduce noise.
To create robust binary \emph{change masks}, we focused on areas with substantial, localized, and consistent decreases in canopy height. The processing steps involve: (i) selecting pixels with a height loss exceeding $15$m, (ii) applying erosion and dilation operators using a 3-pixel kernel to regularize the binary masks, and (iii) removing connected components smaller than $200$m$^2$.
Each of the resulting 73 change areas was manually validated by a forest expert, ensuring the quality and accuracy of the benchmark. We assigned zero values to false positive in the ALS change map. Illustrations, detailed explanations of hyperparameter choices and verification processes are provided in \cref{fig:qualitative-dynamic} and the supplementary material.

\subsection{Results and Analysis}
\label{sec:delta:results}
We evaluate different approaches for detecting significant canopy height change between two VHR images, a task which holds significant applications in forestry management and deforestation monitoring.

\para{Setting.}
We provide each model images from 2022 and 2023 and generate two canopy height maps. We obtain a change map by taking the difference. We do not directly compare the ALS-derived and predicted change maps, as the estimation error of canopy height can be larger than normal tree growth. Instead, we apply the preprocessing described in \cref{sec:delta:data} to produce a binary mask of predicted canopy height change.

\para{Metrics.}
We evaluate the predicted canopy height change masks by computing the pixel-wise \textbf{Precision}, \textbf{Recall}, \textbf{F1 score}, and \textbf{IoU} with respect to the ALS-derived masks.

\para{Results.} We compare height change masks obtained with a PVTv2 model trained on Open-Canopy with those derived from height maps provided by \citep{schwartz2023forms} and Global Forest Change \citep{hansen2013high}. As detailed in \cref{tab:dynamic}, our model achieves significantly better performance than other methods. \cref{fig:qualitative-dynamic} illustrates that while our predicted change maps do not perfectly align with the ground truth maps, the consistency of our predictions suggests their potential utility in detecting significant year-to-year changes.

\begin{table}[t]
    \caption{{\bf  Forest Change Mask Evaluation.} We evaluate our best model (PVTv2) for the task of canopy height change detection.}
    \centering


\begin{tabular}{l cccc}
\toprule
& Precision  & Recall & F1 score & IoU\\
& (\%) & (\%) & (\%) & (\%) \\ \midrule
   Schwartz \citep{schwartz2023forms}
   & \bf 63.5 &  \hphantom{4}3.2 &  \hphantom{3}6.0 & \hphantom{2}3.1
   \\
   GFC \citep{hansen2013high}
   & \hphantom{3}0.9 & \hphantom{4}11.1 & \hphantom{3}1.7 & \hphantom{2}0.8
   \\
   \greyrule
   \bf PVTv2 (ours) & 53.8 & \bf 54.3 & \bf 54.1 & \bf 37.0
   \\
\bottomrule
\end{tabular}


    \label{tab:dynamic}
\end{table}



\section{Conclusion}
We introduced Open-Canopy, an open-access country-scale benchmark combining VHR satellite imagery with ALS-derived canopy height measurements. We evaluated multiple state-of-the-art computer vision models for canopy height estimation. Despite the dominance of convolutional networks in prior works, our findings suggest that transformer-based architectures exhibit superior performance. We also proposed Open-Canopy-$\Delta$, a benchmark for canopy height change detection in consecutive observations, a difficult task, even for the best-performing models. We hope that our open-access benchmarks will encourage the computer vision community to further explore canopy height estimation as a standard task for evaluating new architectures and inspire forestry experts to design bespoke architectures.

\FloatBarrier
\bibliographystyle{ieee_fullname}
\bibliography{mybib}

\begin{thebibliography}{77}
\providecommand{\natexlab}[1]{#1}
\providecommand{\url}[1]{\texttt{#1}}
\expandafter\ifx\csname urlstyle\endcsname\relax
  \providecommand{\doi}[1]{doi: #1}\else
  \providecommand{\doi}{doi: \begingroup \urlstyle{rm}\Url}\fi

\bibitem[gre()]{greco}
{Fiches descriptives des grandes régions écologiques ({GRECO}) et des sylvoécorégions ({SER})}.
\newblock \url{https://inventaire-forestier.ign.fr/spip.php?article773}.
\newblock Accessed: 2024-04-29.

\bibitem[red()]{reduce}
{PyTorch: ReduceLROnPlateau}.
\newblock \url{org/docs/stable/generated/torch.optim.lr_scheduler.ReduceLROnPlateau.html\#torch.optim.lr_scheduler.ReduceLROnPlateau}.
\newblock Accessed: 2024-02-29.

\bibitem[Asner et~al.(2004)Asner, Keller, Pereira, Zweede, and Silva]{asner2004canopy}
Gregory~P Asner, Michael Keller, Rodrigo Pereira, Jr, Johan~C Zweede, and Jose~NM Silva.
\newblock Canopy damage and recovery after selective logging in {A}mazonia: {F}ield and satellite studies.
\newblock \emph{Ecological Applications}, 2004.

\bibitem[Bastani et~al.(2023)Bastani, Wolters, Gupta, Ferdinando, and Kembhavi]{10377451}
Favyen Bastani, Piper Wolters, Ritwik Gupta, Joe Ferdinando, and Aniruddha Kembhavi.
\newblock Satlaspretrain: A large-scale dataset for remote sensing image understanding.
\newblock In \emph{ICCV}, pages 16726--16736, 2023.

\bibitem[Carlson and Ripley(1997)]{carlson1997relation}
Toby~N Carlson and David~A Ripley.
\newblock On the relation between {NDVI}, fractional vegetation cover, and leaf area index.
\newblock \emph{Remote sensing of Environment}, 1997.

\bibitem[Chen et~al.(2017)Chen, Papandreou, Schroff, and Adam]{chen2017rethinking}
Liang-Chieh Chen, George Papandreou, Florian Schroff, and Hartwig Adam.
\newblock Rethinking atrous convolution for semantic image segmentation.
\newblock \emph{arXiv preprint arXiv:1706.05587}, 2017.

\bibitem[Christie et~al.(2018)Christie, Fendley, Wilson, and Mukherjee]{christie2018functional}
Gordon Christie, Neil Fendley, James Wilson, and Ryan Mukherjee.
\newblock Functional map of the world.
\newblock In \emph{CVPR}, 2018.

\bibitem[Chu et~al.(2021)Chu, Tian, Wang, Zhang, Ren, Wei, Xia, and Shen]{chu2021twins}
Xiangxiang Chu, Zhi Tian, Yuqing Wang, Bo Zhang, Haibing Ren, Xiaolin Wei, Huaxia Xia, and Chunhua Shen.
\newblock {Twins: R}evisiting the design of spatial attention in vision transformers.
\newblock \emph{NeurIPS}, 2021.

\bibitem[Commission(2024)]{eudr}
European Commission.
\newblock Regulation on deforestation-free products.
\newblock \url{https://environment.ec.europa.eu/topics/forests/deforestation/regulation-deforestation-free-products_en}, 2024.
\newblock [Online; accessed 12-Sep-2024].

\bibitem[Das and Stephenson(2015)]{das2015improving}
Adrian~J Das and Nathan~L Stephenson.
\newblock Improving estimates of tree mortality probability using potential growth rate.
\newblock \emph{Canadian Journal of Forest Research}, 2015.

\bibitem[de~France(2024)]{chantilly}
Institut de France.
\newblock Collectif sauvons la foret de chantilly.
\newblock \url{https://chateaudechantilly.fr/la-foret/ensemble-sauvons-la-foret-de-chantilly/ }, 2024.
\newblock [Online; accessed 12-May-2024].

\bibitem[Decuyper et~al.(2022)Decuyper, Ch{\'a}vez, Lohbeck, Lastra, Tsendbazar, Hackl{\"a}nder, Herold, and V{\aa}gen]{decuyper2022continuous}
Mathieu Decuyper, Roberto~O Ch{\'a}vez, Madelon Lohbeck, Jos{\'e}~A Lastra, Nandika Tsendbazar, Julia Hackl{\"a}nder, Martin Herold, and Tor-G V{\aa}gen.
\newblock Continuous monitoring of forest change dynamics with satellite time series.
\newblock \emph{Remote Sensing of Environment}, 2022.

\bibitem[DINAMIS(2024)]{dinamis}
DINAMIS.
\newblock French national facility for institutional procurement of vhr satellite imagery.
\newblock \url{https://openspot-dinamis.data-terra.org}, 2024.
\newblock [Online; accessed 12-May-2024].

\bibitem[Dosovitskiy et~al.(2020)Dosovitskiy, Beyer, Kolesnikov, Weissenborn, Zhai, Unterthiner, Dehghani, Minderer, Heigold, Gelly, et~al.]{dosovitskiy2020image}
Alexey Dosovitskiy, Lucas Beyer, Alexander Kolesnikov, Dirk Weissenborn, Xiaohua Zhai, Thomas Unterthiner, Mostafa Dehghani, Matthias Minderer, Georg Heigold, Sylvain Gelly, et~al.
\newblock An image is worth 16x16 words: {T}ransformers for image recognition at scale.
\newblock In \emph{ICLR}, 2020.

\bibitem[Dubayah et~al.(2020)Dubayah, Blair, Goetz, Fatoyinbo, Hansen, Healey, Hofton, Hurtt, Kellner, Luthcke, et~al.]{dubayah2020global}
Ralph Dubayah, James~Bryan Blair, Scott Goetz, Lola Fatoyinbo, Matthew Hansen, Sean Healey, Michelle Hofton, George Hurtt, James Kellner, Scott Luthcke, et~al.
\newblock The global ecosystem dynamics investigation: {H}igh-resolution laser ranging of the {E}arth’s forests and topography.
\newblock \emph{Science of remote sensing}, 2020.

\bibitem[Erfanifard et~al.(2022)Erfanifard, Lotfi~Nasirabad, and Stere{\'n}czak]{erfanifard2022assessment}
Yousef Erfanifard, Mohsen Lotfi~Nasirabad, and Krzysztof Stere{\'n}czak.
\newblock Assessment of {I}ran’s mangrove forest dynamics (1990--2020) using {L}andsat time series.
\newblock \emph{Remote Sensing}, 2022.

\bibitem[Etalab(2024)]{etalab}
Etalab.
\newblock Open licence 2.0.
\newblock \url{https://www.etalab.gouv.fr/wp-content/uploads/2018/11/open-licence.pdf}, 2024.
\newblock [Online; accessed 12-May-2024].

\bibitem[Fassnacht et~al.(2024)Fassnacht, Mager, Waser, Kanjir, Sch{\"a}fer, Buhvald, Shafeian, Schiefer, Stan{\v{c}}i{\v{c}}, Immitzer, et~al.]{fassnacht2024forest}
Fabian~Ewald Fassnacht, Christoph Mager, Lars~T Waser, Ur{\v{s}}a Kanjir, Jannika Sch{\"a}fer, Ana~Poto{\v{c}}nik Buhvald, Elham Shafeian, Felix Schiefer, Liza Stan{\v{c}}i{\v{c}}, Markus Immitzer, et~al.
\newblock Forest practitioners’ requirements for remote sensing-based canopy height, wood-volume, tree species, and disturbance products.
\newblock \emph{Forestry: An International Journal of Forest Research}, 2024.

\bibitem[Foundation(2024)]{neon}
US~National~Science Foundation.
\newblock Neon (national ecological observatory network). ecosystem structure (dp3.30015.001)).
\newblock \url{https://data.neonscience.org/}, 2024.
\newblock Dataset accessed from https://data.neonscience.org/data-products/DP3.30015.001 on October 11, 2024.

\bibitem[Gaveau and Hill(2003)]{gaveau2003quantifying}
David~LA Gaveau and Ross~A Hill.
\newblock Quantifying canopy height underestimation by laser pulse penetration in small-footprint airborne laser scanning data.
\newblock \emph{Canadian Journal of Remote Sensing}, 2003.

\bibitem[Gebru et~al.(2021)Gebru, Morgenstern, Vecchione, Vaughan, Wallach, Iii, and Crawford]{gebru2021datasheets}
Timnit Gebru, Jamie Morgenstern, Briana Vecchione, Jennifer~Wortman Vaughan, Hanna Wallach, Hal~Daum{\'e} Iii, and Kate Crawford.
\newblock Datasheets for datasets.
\newblock \emph{Communications of the ACM}, 2021.

\bibitem[Getzin et~al.(2012)Getzin, Wiegand, and Sch{\"o}ning]{getzin2012assessing}
Stephan Getzin, Kerstin Wiegand, and Ingo Sch{\"o}ning.
\newblock Assessing biodiversity in forests using very high-resolution images and unmanned aerial vehicles.
\newblock \emph{Methods in ecology and evolution}, 2012.

\bibitem[Gillespie et~al.(1987)Gillespie, Kahle, and Walker]{gillespie1987color}
Alan~R Gillespie, Anne~B Kahle, and Richard~E Walker.
\newblock Color enhancement of highly correlated images. ii. channel ratio and “chromaticity” transformation techniques.
\newblock \emph{Remote Sensing of Environment}, 1987.

\bibitem[Hansen et~al.(2013)Hansen, Potapov, Moore, Hancher, Turubanova, Tyukavina, Thau, Stehman, Goetz, Loveland, and others.]{hansen2013high}
Matthew~C Hansen, Peter~V Potapov, Rebecca Moore, Matt Hancher, Svetlana~A Turubanova, Alexandra Tyukavina, David Thau, Stephen~V Stehman, Scott~J Goetz, Thomas~R Loveland, and others.
\newblock High-resolution global maps of 21st-century forest cover change.
\newblock 2013.

\bibitem[Hu et~al.(2022)Hu, Shen, Wallis, Allen-Zhu, Li, Wang, Wang, and Chen]{hu2021lora}
Edward~J Hu, Yelong Shen, Phillip Wallis, Zeyuan Allen-Zhu, Yuanzhi Li, Shean Wang, Lu Wang, and Weizhu Chen.
\newblock {LoRa}: {L}ow-rank adaptation of large language models.
\newblock \emph{ICLR}, 2022.

\bibitem[Huertas et~al.(2022)Huertas, Sabatier, Derroire, Ferry, Jackson, P{\'e}lissier, and Vincent]{huertas2022mapping}
Claudia Huertas, Daniel Sabatier, G{\'e}raldine Derroire, Bruno Ferry, Toby~D Jackson, Rapha{\"e}l P{\'e}lissier, and Gr{\'e}goire Vincent.
\newblock Mapping tree mortality rate in a tropical moist forest using multi-temporal {LiDAR}.
\newblock \emph{International Journal of Applied Earth Observation and Geoinformation}, 2022.

\bibitem[IGN()]{LIDARHD_doc}
IGN.
\newblock Lidar hd technical description.
\newblock \url{https://geoservices.ign.fr/sites/default/files/2023-10/DC_LiDAR_HD_1-0_PTS.pdf}.
\newblock Online; accessed 2024-02-21.

\bibitem[IGN(2024{\natexlab{a}})]{france_trees}
IGN.
\newblock More than 190 tree species inventoried in france.
\newblock \url{https://inventaire-forestier.ign.fr/spip.php?article175}, 2024{\natexlab{a}}.
\newblock [Online; accessed 12-May-2024].

\bibitem[IGN(2024{\natexlab{b}})]{lidarhd}
IGN.
\newblock {LiDAR HD : V}ers une nouvelle cartographie 3d du territoire.
\newblock \url{https://www.ign.fr/institut/lidar-hd-vers-une-nouvelle-cartographie-3d-du-territoire}, 2024{\natexlab{b}}.
\newblock [Online; accessed 12-May-2024].

\bibitem[IGN(2024{\natexlab{c}})]{link_to_forest_masks}
IGN.
\newblock Forest data base.
\newblock \url{https://geoservices.ign.fr/bdforet}, 2024{\natexlab{c}}.
\newblock [Online; accessed 12-May-2024].

\bibitem[Jackson and Adam(2020)]{jackson2020remote}
Colbert~M Jackson and Elhadi Adam.
\newblock Remote sensing of selective logging in tropical forests: {C}urrent state and future directions.
\newblock \emph{iForest-Biogeosciences and Forestry}, 2020.

\bibitem[Kalinicheva et~al.(2022)Kalinicheva, Landrieu, Mallet, and Chehata]{kalinicheva2022multi}
Ekaterina Kalinicheva, Loic Landrieu, Cl{\'e}ment Mallet, and Nesrine Chehata.
\newblock Multi-layer modeling of dense vegetation from aerial {LiDAR} scans.
\newblock In \emph{CVPR Workshop Earth Vision}, 2022.

\bibitem[Keenan(2015)]{keenan2015climate}
Rodney~J Keenan.
\newblock Climate change impacts and adaptation in forest management: {A} review.
\newblock \emph{Annals of forest science}, 2015.

\bibitem[Kingma and Ba(2015)]{kingma2014adam}
Diederik~P Kingma and Jimmy Ba.
\newblock {Adam: A} method for stochastic optimization.
\newblock \emph{ICLR}, 2015.

\bibitem[Kuemmerle et~al.(2009)Kuemmerle, Chaskovskyy, Knorn, Radeloff, Kruhlov, Keeton, and Hostert]{kuemmerle2009forest}
Tobias Kuemmerle, Oleh Chaskovskyy, Jan Knorn, Volker~C Radeloff, Ivan Kruhlov, William~S Keeton, and Patrick Hostert.
\newblock Forest cover change and illegal logging in the {U}krainian {C}arpathians in the transition period from 1988 to 2007.
\newblock \emph{Remote Sensing of Environment}, 2009.

\bibitem[Lang et~al.(2023)Lang, Jetz, Schindler, and Wegner]{lang2023high}
Nico Lang, Walter Jetz, Konrad Schindler, and Jan~Dirk Wegner.
\newblock A high-resolution canopy height model of the {E}arth.
\newblock \emph{Nature Ecology \& Evolution}, 2023.

\bibitem[Lecq et~al.(2017)Lecq, Loisel, Brischoux, Mullin, and Bonnet]{lecq2017importance}
St{\'e}phane Lecq, Anne Loisel, Francois Brischoux, Stephen~J Mullin, and Xavier Bonnet.
\newblock Importance of ground refuges for the biodiversity in agricultural hedgerows.
\newblock \emph{Ecological Indicators}, 2017.

\bibitem[Liu et~al.(2023)Liu, Brandt, Nord-Larsen, Chave, Reiner, Lang, Tong, Ciais, Igel, Pascual, et~al.]{liu2023overlooked}
Siyu Liu, Martin Brandt, Thomas Nord-Larsen, Jerome Chave, Florian Reiner, Nico Lang, Xiaoye Tong, Philippe Ciais, Christian Igel, Adrian Pascual, et~al.
\newblock The overlooked contribution of trees outside forests to tree cover and woody biomass across {E}urope.
\newblock \emph{Science Advances}, 2023.

\bibitem[Liu et~al.(2021)Liu, Lin, Cao, Hu, Wei, Zhang, Lin, and Guo]{liu2021swin}
Ze Liu, Yutong Lin, Yue Cao, Han Hu, Yixuan Wei, Zheng Zhang, Stephen Lin, and Baining Guo.
\newblock {SWIN} transformer: {H}ierarchical vision transformer using shifted windows.
\newblock In \emph{ICCV}, 2021.

\bibitem[MacDicken(2015)]{macdicken2015globalb}
Kenneth~G MacDicken.
\newblock Global forest resources assessment 2015: {W}hat, why and how?
\newblock \emph{Forest Ecology and Management}, 2015.

\bibitem[MacDicken et~al.(2015)MacDicken, Sola, Hall, Sabogal, Tadoum, and de~Wasseige]{macdicken2015global}
Kenneth~G MacDicken, Phosiso Sola, John~E Hall, Cesar Sabogal, Martin Tadoum, and Carlos de Wasseige.
\newblock Global progress toward sustainable forest management.
\newblock \emph{Forest Ecology and Management}, 2015.

\bibitem[McRoberts and Tomppo(2007)]{mcroberts2007remote}
Ronald~E McRoberts and Erkki~O Tomppo.
\newblock Remote sensing support for national forest inventories.
\newblock \emph{Remote sensing of environment}, 2007.

\bibitem[of~Agriculture(2024{\natexlab{a}})]{pnfb}
French~Ministry of Agriculture.
\newblock The national forest and wood programme ({PNFB}).
\newblock \url{https://inis.iaea.org/search/search.aspx?orig_q=RN:51010336}, 2024{\natexlab{a}}.
\newblock [Online; accessed 12-May-2024].

\bibitem[of~Agriculture(2024{\natexlab{b}})]{naip}
United States~Department of Agriculture.
\newblock National agriculture imagery program ({NAIP}).
\newblock \url{https://www.fsa.usda.gov/Assets/USDA-FSA-Public/usdafiles/APFO/support-documents/pdfs/naip_infosheet_2016.pdf}, 2024{\natexlab{b}}.
\newblock [Online; accessed 12-May-2024].

\bibitem[of~Topography~Swisstopo(2024{\natexlab{a}})]{swissortho}
Federal~Office of Topography~Swisstopo.
\newblock Swisstopo orthophotos.
\newblock \url{https://www.swisstopo.admin.ch/fr/orthophotos}, 2024{\natexlab{a}}.
\newblock [Online; accessed 12-May-2024].

\bibitem[of~Topography~Swisstopo(2024{\natexlab{b}})]{swisstopo}
Federal~Office of Topography~Swisstopo.
\newblock Swisstopo {LiDAR} data acquisition.
\newblock \url{https://www.swisstopo.admin.ch/en/lidar-data-swisstopo}, 2024{\natexlab{b}}.
\newblock [Online; accessed 12-May-2024].

\bibitem[Oquab et~al.(2023)Oquab, Darcet, Moutakanni, Vo, Szafraniec, Khalidov, Fernandez, HAZIZA, Massa, El-Nouby, et~al.]{oquab2023dinov2}
Maxime Oquab, Timoth{\'e}e Darcet, Th{\'e}o Moutakanni, Huy~V Vo, Marc Szafraniec, Vasil Khalidov, Pierre Fernandez, Daniel HAZIZA, Francisco Massa, Alaaeldin El-Nouby, et~al.
\newblock {DINOv2: L}earning robust visual features without supervision.
\newblock \emph{TMLR}, 2023.

\bibitem[Pauls et~al.(2024)Pauls, Zimmer, Kelly, Schwartz, Saatchi, Ciais, Pokutta, Brandt, and Gieseke]{pauls2024estimating}
Jan Pauls, Max Zimmer, Una~M Kelly, Martin Schwartz, Sassan Saatchi, Philippe Ciais, Sebastian Pokutta, Martin Brandt, and Fabian Gieseke.
\newblock Estimating canopy height at scale.
\newblock In \emph{ICML}, 2024.

\bibitem[Peel et~al.(2007)Peel, Finlayson, and McMahon]{peel2007updated}
Murray~C Peel, Brian~L Finlayson, and Thomas~A McMahon.
\newblock Updated world map of the {K}{\"o}ppen-{G}eiger climate classification.
\newblock \emph{Hydrology and earth system sciences}, 2007.

\bibitem[Peng and Yili(2022)]{peng2022research}
Guan Peng and ZHENG Yili.
\newblock Research on forest phenology prediction based on {LSTM} and {GRU} model.
\newblock \emph{Journal of Resources and Ecology}, 2022.

\bibitem[Potapov et~al.(2021)Potapov, Li, Hernandez-Serna, Tyukavina, Hansen, Kommareddy, Pickens, Turubanova, Tang, Silva, et~al.]{potapov2021mapping}
Peter Potapov, Xinyuan Li, Andres Hernandez-Serna, Alexandra Tyukavina, Matthew~C Hansen, Anil Kommareddy, Amy Pickens, Svetlana Turubanova, Hao Tang, Carlos~Edibaldo Silva, et~al.
\newblock Mapping global forest canopy height through integration of {GEDI} and {L}andsat data.
\newblock \emph{Remote Sensing of Environment}, 2021.

\bibitem[Pretzsch et~al.(2023)Pretzsch, Del~R{\'\i}o, Arcangeli, Bielak, Dudzinska, Forrester, Kl{\"a}dtke, Kohnle, Ledermann, Matthews, et~al.]{pretzsch2023forest}
Hans Pretzsch, Miren Del~R{\'\i}o, Catia Arcangeli, Kamil Bielak, Malgorzata Dudzinska, David~Ian Forrester, Joachim Kl{\"a}dtke, Ulrich Kohnle, Thomas Ledermann, Robert Matthews, et~al.
\newblock Forest growth in {E}urope shows diverging large regional trends.
\newblock \emph{Scientific Reports}, 2023.

\bibitem[Qin et~al.(2023)Qin, Ma, Zhu, Wu, and Su]{qin20233pg}
Jushuang Qin, Menglu Ma, Yutong Zhu, Baoguo Wu, and Xiaohui Su.
\newblock {3PG-MT-LSTM: A} hybrid model under biomass compatibility constraints for the prediction of long-term forest growth to support sustainable management.
\newblock \emph{Forests}, 2023.

\bibitem[Radford et~al.(2021)Radford, Kim, Hallacy, Ramesh, Goh, Agarwal, Sastry, Askell, Mishkin, Clark, et~al.]{radford2021learning}
Alec Radford, Jong~Wook Kim, Chris Hallacy, Aditya Ramesh, Gabriel Goh, Sandhini Agarwal, Girish Sastry, Amanda Askell, Pamela Mishkin, Jack Clark, et~al.
\newblock Learning transferable visual models from natural language supervision.
\newblock In \emph{ICML}, 2021.

\bibitem[Reed et~al.(2023)Reed, Gupta, Li, Brockman, Funk, Clipp, Keutzer, Candido, Uyttendaele, and Darrell]{reed2023scale}
Colorado~J Reed, Ritwik Gupta, Shufan Li, Sarah Brockman, Christopher Funk, Brian Clipp, Kurt Keutzer, Salvatore Candido, Matt Uyttendaele, and Trevor Darrell.
\newblock {Scale-MAE: A} scale-aware masked autoencoder for multiscale geospatial representation learning.
\newblock In \emph{ICCV}, 2023.

\bibitem[Reyer et~al.(2020)Reyer, Silveyra~Gonzalez, Dolos, Hartig, Hauf, Noack, Lasch-Born, R{\"o}tzer, Pretzsch, Meesenburg, et~al.]{reyer2020profound}
Christopher~PO Reyer, Ramiro Silveyra~Gonzalez, Klara Dolos, Florian Hartig, Ylva Hauf, Matthias Noack, Petra Lasch-Born, Thomas R{\"o}tzer, Hans Pretzsch, Henning Meesenburg, et~al.
\newblock The {PROFOUND} database for evaluating vegetation models and simulating climate impacts on {E}uropean forests.
\newblock \emph{Earth System Science Data}, 2020.

\bibitem[Ridnik et~al.(2021)Ridnik, Ben-Baruch, Noy, and Zelnik-Manor]{ridnik2021imagenet}
Tal Ridnik, Emanuel Ben-Baruch, Asaf Noy, and Lihi Zelnik-Manor.
\newblock {ImageNet-21K} pretraining for the masses.
\newblock In \emph{NeurIPS Datasets and Benchmarks Track}, 2021.

\bibitem[Ronneberger et~al.(2015)Ronneberger, Fischer, and Brox]{ronneberger2015u}
Olaf Ronneberger, Philipp Fischer, and Thomas Brox.
\newblock {UNet: C}onvolutional networks for biomedical image segmentation.
\newblock In \emph{MICCAI}. Springer, 2015.

\bibitem[Russakovsky et~al.(2015)Russakovsky, Deng, Su, Krause, Satheesh, Ma, Huang, Karpathy, Khosla, Bernstein, et~al.]{russakovsky2015imagenet}
Olga Russakovsky, Jia Deng, Hao Su, Jonathan Krause, Sanjeev Satheesh, Sean Ma, Zhiheng Huang, Andrej Karpathy, Aditya Khosla, Michael Bernstein, et~al.
\newblock Image{N}et large scale visual recognition challenge.
\newblock \emph{IJCV}, 2015.

\bibitem[Schleich et~al.(2023)Schleich, Durrieu, and Vega]{schleich2023improving}
Anouk Schleich, Sylvie Durrieu, and C{\'e}dric Vega.
\newblock Improving gedi footprint geolocation using a high resolution digital elevation model.
\newblock \emph{IEEE Journal of Selected Topics in Applied Earth Observations and Remote Sensing}, 2023.

\bibitem[Schwartz(2023)]{schwartz2023mapping}
Martin Schwartz.
\newblock \emph{Mapping forest height and biomass at high resolution in {F}rance with satellite remote sensing and deep learning}.
\newblock PhD thesis, Universit{\'e} Paris-Saclay, 2023.

\bibitem[Schwartz et~al.(2023)Schwartz, Ciais, De~Truchis, Chave, Ottl{\'e}, Vega, Wigneron, Nicolas, Jouaber, Liu, Brandt, and Fayad]{schwartz2023forms}
Martin Schwartz, Philippe Ciais, Aur{\'e}lien De~Truchis, J{\'e}r{\^o}me Chave, Catherine Ottl{\'e}, Cedric Vega, Jean-Pierre Wigneron, Manuel Nicolas, Sami Jouaber, Siyu Liu, Martin Brandt, and Ibrahim Fayad.
\newblock {FORMS: F}orest multiple source height, wood volume, and biomass maps in {F}rance at 10 to 30m resolution based on {S}entinel-1, {S}entinel-2, and {GEDI} data with a deep learning approach.
\newblock \emph{Earth System Science Data}, 2023.

\bibitem[Smith(1995)]{smith1995digital}
Gary~S Smith.
\newblock Digital orthophotography and {GIS}.
\newblock In \emph{Proceedings of the 1995 ESRI user conference}, 1995.

\bibitem[Stephenson et~al.(2014)Stephenson, Das, Condit, Russo, Baker, Beckman, Coomes, Lines, Morris, R{\"u}ger, et~al.]{stephenson2014rate}
Nathan~L Stephenson, AJ Das, R Condit, SE Russo, PJ Baker, Noelle~G Beckman, DA Coomes, ER Lines, WK Morris, Nadja R{\"u}ger, et~al.
\newblock Rate of tree carbon accumulation increases continuously with tree size.
\newblock \emph{Nature}, 2014.

\bibitem[Tang et~al.(2023)Tang, Stoker, Luthcke, Armston, Lee, Blair, and Hofton]{tang2023evaluating}
Hao Tang, Jason Stoker, Scott Luthcke, John Armston, Kyungtae Lee, Bryan Blair, and Michelle Hofton.
\newblock Evaluating and mitigating the impact of systematic geolocation error on canopy height measurement performance of {GEDI}.
\newblock \emph{Remote Sensing of Environment}, 2023.

\bibitem[Thompson and Magrath(2021)]{thompson2021preventing}
Sara~T Thompson and William~B Magrath.
\newblock Preventing illegal logging.
\newblock \emph{Forest Policy and Economics}, 2021.

\bibitem[Tian et~al.(2021)Tian, Cai, Jin, Hufkens, Scheifinger, Tagesson, Smets, Van~Hoolst, Bonte, Ivits, et~al.]{tian2021calibrating}
Feng Tian, Zhanzhang Cai, Hongxiao Jin, Koen Hufkens, Helfried Scheifinger, Torbern Tagesson, Bruno Smets, Roel Van~Hoolst, Kasper Bonte, Eva Ivits, et~al.
\newblock Calibrating vegetation phenology from {S}entinel-2 using eddy covariance, {PhenoCam}, and {PEP725} networks across {E}urope.
\newblock \emph{Remote Sensing of Environment}, 2021.

\bibitem[Tolan et~al.(2024)Tolan, Yang, Nosarzewski, Couairon, Vo, Brandt, Spore, Majumdar, Haziza, Vamaraju, et~al.]{tolan2024very}
Jamie Tolan, Hung-I Yang, Benjamin Nosarzewski, Guillaume Couairon, Huy~V Vo, John Brandt, Justine Spore, Sayantan Majumdar, Daniel Haziza, Janaki Vamaraju, et~al.
\newblock Very high resolution canopy height maps from {RGB} imagery using self-supervised vision transformer and convolutional decoder trained on aerial {LiDAR}.
\newblock \emph{Remote Sensing of Environment}, 2024.

\bibitem[Tomppo et~al.(2010)Tomppo, Gschwantner, Lawrence, McRoberts, Gabler, Schadauer, Vidal, Lanz, St{\aa}hl, Cienciala, et~al.]{tomppo2010national}
Erkki Tomppo, Thomas Gschwantner, Mark Lawrence, Ronald~E McRoberts, Karl Gabler, K Schadauer, Claude Vidal, A Lanz, G{\"o}ran St{\aa}hl, Emil Cienciala, et~al.
\newblock National forest inventories.
\newblock \emph{Pathways for Common Reporting. European Science Foundation}, 2010.

\bibitem[Turubanova et~al.(2023)Turubanova, Potapov, Hansen, Li, Tyukavina, Pickens, Hernandez-Serna, Arranz, Guerra-Hernandez, Senf, et~al.]{turubanova2023tree}
Svetlana Turubanova, Peter Potapov, Matthew~C Hansen, Xinyuan Li, Alexandra Tyukavina, Amy~H Pickens, Andres Hernandez-Serna, Adrian~Pascual Arranz, Juan Guerra-Hernandez, Cornelius Senf, et~al.
\newblock Tree canopy extent and height change in {E}urope, 2001--2021, quantified using {L}andsat data archive.
\newblock \emph{Remote Sensing of Environment}, 2023.

\bibitem[Wagner et~al.(2024)Wagner, Roberts, Ritz, Carter, Dalagnol, Favrichon, Hirye, Brandt, Ciais, and Saatchi]{wagner2306sub}
FH Wagner, S Roberts, AL Ritz, G Carter, R Dalagnol, S Favrichon, M Hirye, M Brandt, P Ciais, and S Saatchi.
\newblock Sub-meter tree height mapping of {C}alifornia using aerial images and {LiDAR}-informed {U-Net} model.
\newblock \emph{Remote Sensing of Environment}, 2024.

\bibitem[Wang et~al.(2022)Wang, Xie, Li, Fan, Song, Liang, Lu, Luo, and Shao]{wang2021pvtv2}
Wenhai Wang, Enze Xie, Xiang Li, Deng-Ping Fan, Kaitao Song, Ding Liang, Tong Lu, Ping Luo, and Ling Shao.
\newblock {PVTv2: I}mproved baselines with pyramid vision transformer.
\newblock \emph{Computational Visual Media}, 2022.

\bibitem[Yang and Zhu(2013)]{yang2013ortho}
Guo~Dong Yang and Xiang Zhu.
\newblock Ortho-rectification of {SPOT} 6 satellite images based on {RPC} models.
\newblock \emph{Applied Mechanics and Materials}, 2013.

\bibitem[Ye et~al.(2019)Ye, Gao, Marcos-Martinez, Mallants, and Bryan]{ye2019projecting}
Long Ye, Lei Gao, Raymundo Marcos-Martinez, Dirk Mallants, and Brett~A Bryan.
\newblock Projecting {A}ustralia's forest cover dynamics and exploring influential factors using deep learning.
\newblock \emph{Environmental Modelling \& Software}, 2019.

\bibitem[Yu et~al.(2004)Yu, Hyypp{\"a}, Kaartinen, and Maltamo]{yu2004automatic}
Xiaowei Yu, Juha Hyypp{\"a}, Harri Kaartinen, and Matti Maltamo.
\newblock Automatic detection of harvested trees and determination of forest growth using airborne laser scanning.
\newblock \emph{Remote sensing of Environment}, 2004.

\bibitem[Yu et~al.(2006)Yu, Hyypp{\"a}, Kukko, Maltamo, and Kaartinen]{yu2006change}
Xiaowei Yu, Juha Hyypp{\"a}, Antero Kukko, Matti Maltamo, and Harri Kaartinen.
\newblock Change detection techniques for canopy height growth measurements using airborne laser scanner data.
\newblock \emph{Photogrammetric Engineering \& Remote Sensing}, 2006.

\bibitem[Zhang et~al.(2019)Zhang, Wu, and Yang]{zhang2019forests}
Yanchao Zhang, Hanxuan Wu, and Wen Yang.
\newblock Forests growth monitoring based on tree canopy {3D} reconstruction using {UAV} aerial photogrammetry.
\newblock \emph{Forests}, 2019.

\end{thebibliography}


\maketitlesupplementary

\renewcommand\thefigure{\Alph{figure}}
\renewcommand\thesection{\Alph{section}}
\renewcommand\thetable{\Alph{table}}
\renewcommand\theequation{\Alph{equation}}
\setcounter{equation}{0}
\setcounter{section}{0}
\setcounter{figure}{0}
\setcounter{table}{0}




We present additional results, analyses, and experiments to support our study. First, we detail our validation of the ground truth using terrain measurements and manual verifications in \cref{sec:sup:field}.
Next, we provide further results in \cref{sec:sup:results}, including new analyses, qualitative illustrations, and experimental settings. We then conduct a detailed ablation study in \cref{sec:sup:ablation} to examine the influence of key hyperparameters and design choices. Additionally, we offer a comprehensive description of the dataset and its construction in \cref{sec:sup:dataset}.
Finally, we provide the Datasheet for Dataset \citep{gebru2021datasheets} for our benchmark.

\section{Validation with Field Measurements}
\begin{figure*}[t]
    \centering
    \begin{tabular}{c}
    \begin{subfigure}{\linewidth}

\begin{tabular}{cc}
\centering
\begin{tikzpicture}
    \begin{axis}[
        width=0.5\textwidth,
        title={\small Estimated Tree Height (in m)},
        xlabel={\small Field Measurement},
        ylabel={\small ALS-derived height},
        legend pos=south east,
        xmin=10,
        xmax=42,
        ymin=10,
        ymax=42
    ]
    \addplot[
        only marks,
        mark=*,
        mark options={black}
    ] 
    table [
        x index=0, 
        y index=1, 
        col sep=comma
    ] {figures/scatter_plot_data2.csv};
    
    \addplot[
        red,
        thick,
    ] 
    table [
        x index=0, 
        y index=1, 
        col sep=comma
    ] {figures/line_data2.csv};

    \addplot[
        red,
        thick,
        dotted
    ] 
    table [
        x index=0, 
        y index=1, 
        col sep=comma
    ] {figures/scatter_plot_data_xy.csv};

    \end{axis}
\end{tikzpicture}
&
\begin{tikzpicture}
    \begin{axis}[
        width=0.5\textwidth,
        title={\small Estimated Tree Height (in m)},
        xlabel={\small Field Measurements},
        ylabel={\small Best Model Prediction},
        legend pos=south east,
        xmin=10,
        xmax=42,
        ymin=10,
        ymax=42
    ]
    \addplot[
        only marks,
        mark=*,
        mark options={black}
    ] 
    table [
        x index=0, 
        y index=1, 
        col sep=comma
    ] {figures/scatter_plot_data1.csv};
    
    \addplot[
        red,
        thick,
    ] 
    table [
        x index=0, 
        y index=1, 
        col sep=comma
    ] {figures/line_data1.csv};

    \addplot[
        red,
        thick,
        dotted
    ] 
    table [
        x index=0, 
        y index=1, 
        col sep=comma
    ] {figures/scatter_plot_data_xy.csv};

    \end{axis}
\end{tikzpicture} \\
$R^2=0.85$&
$R^2=0.78$
\end{tabular}
        \caption{{\bf Plot-Level Scatter-Plot.} We represent the maximum heights as measured through ALS or predicted by a PVTv2 model against the manual field measurements. }
          \label{fig:field-verification:scatter}
      \end{subfigure}
       \\~\\
      \begin{subfigure}{\linewidth}\centering
      \caption{{\bf Plot-Level Quantitative Evaluation.} We compare the precision of ALS and a PVTv2 model when taking the field measurements as ground truth. }
         \label{tab:field-verification:table-plot}
      \begin{tabular}{l cccc}
\toprule
 	&MAE (m) 	&nMAE (\%) 	&RMSE (m) 	&Bias (m)\\
\midrule

ALS vs Field measurements 	&2.3 	&13.5 	&3.1 	&1.2\\
PVTv2 vs Field measurements 	&2.9 	&14.4 	&3.7 	&-1.1\\

\bottomrule
\end{tabular}
      \end{subfigure}
     \\~\\
    \begin{subfigure}{\linewidth}
    \caption{{\bf Tree-Level Quantitative Evaluation.} We compare the precision of ALS and a PVTv2 model when taking the field measurements as ground truth. }
         \label{tab:field-verification:table-tree}
      \begin{tabular}{l cccc}
\toprule
 	&MAE (m) 	&nMAE (\%) 	&RMSE (m) 	&Bias (m)\\
\midrule

ALS vs Field measurements 	&1.45&	6.7&	2.0&	0.22\\
PVTv2 vs Field measurements 	&4.0&	15.4	&5.1	&-3.2\\

\bottomrule
\end{tabular}\centering
      \end{subfigure}
    \end{tabular}
       \caption{{\bf Field-Verification of Height Maps.} We validate our ALS-based ground truth by comparing it to field measurements at two scales: plot-level and tree-level.}
    \label{fig:field-verification}
\end{figure*}
\label{sec:sup:field}
Ensuring the accuracy of our ground truth data is crucial for the validity of any computer vision benchmark. While the ALS data from LIDAR-HD have been calibrated and validated internally by the French Mapping Agency (IGN) using plots annotated by the National Forest Office (ONF), we performed additional manual verifications to further confirm their reliability, performed at plot-level and tree-level.

\paragraph{Plot-Level Assessment.} 
We sourced measurements from 135 plots, each with a 15 m radius, across the test set of Open-Canopy. These plots were measured in the field by forestry experts from the National Forest Inventory (INF) within two years of the ALS acquisition. For each plot, we compared the height of the tallest tree measured in situ with the maximum canopy height in the plot as estimated by the ALS data and predicted by our best-performing computer vision model (PVTv2). As shown in \cref{fig:field-verification:scatter} and detailed in  \cref{tab:field-verification:table-plot}, the ALS-derived heights exhibit smaller errors compared to our model's estimates and align closely with the field measurements. This validation confirms the suitability of the ALS data as ground truth for our open-access benchmark.

\paragraph{Tree-Level Assessment.} 
We extended our validation to the individual tree level using data provided by the ONF, consisting of 44 geolocated trees in the Grand Est region. For each tree, we compared the measured height with the highest estimated or predicted height within a 1.5 m radius around the tree's center. The metrics presented in \cref{tab:field-verification:table-tree} corroborate the plot-level findings, further validating the ALS-derived heights. This validation process emphasizes the reliability of our ground truth data, which is essential for advancing computer vision methods in canopy height estimation.

\begin{figure*}
    \centering
    \begin{tabular}{cccl}
     \begin{subfigure}{0.29\textwidth}
        \centering
        \includegraphics[width=\linewidth]{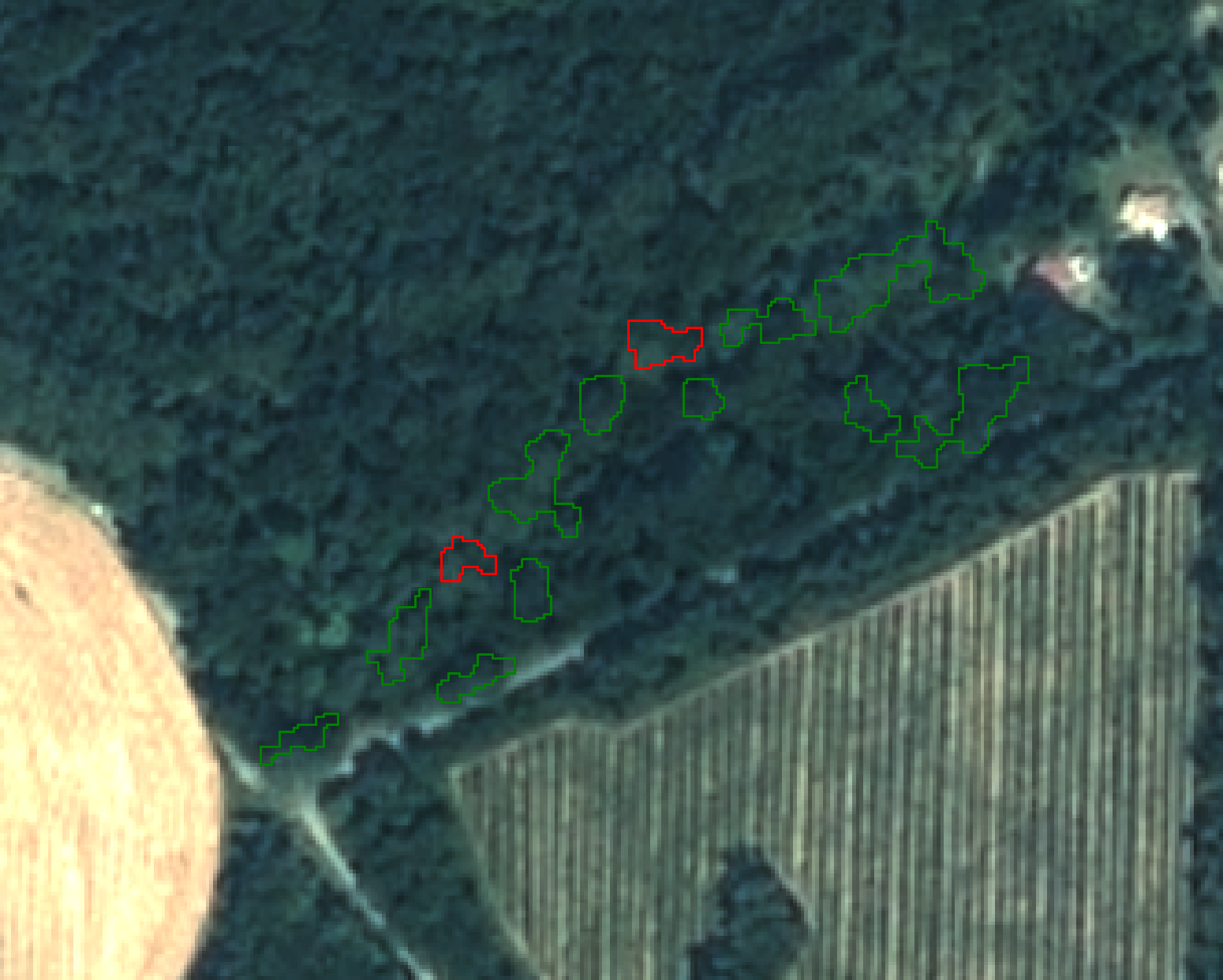}
        \caption{2022 VHR}
        \label{fig:vhr}
    \end{subfigure}
    &
    \begin{subfigure}{0.29\textwidth}
        \centering
        \includegraphics[width=\linewidth]{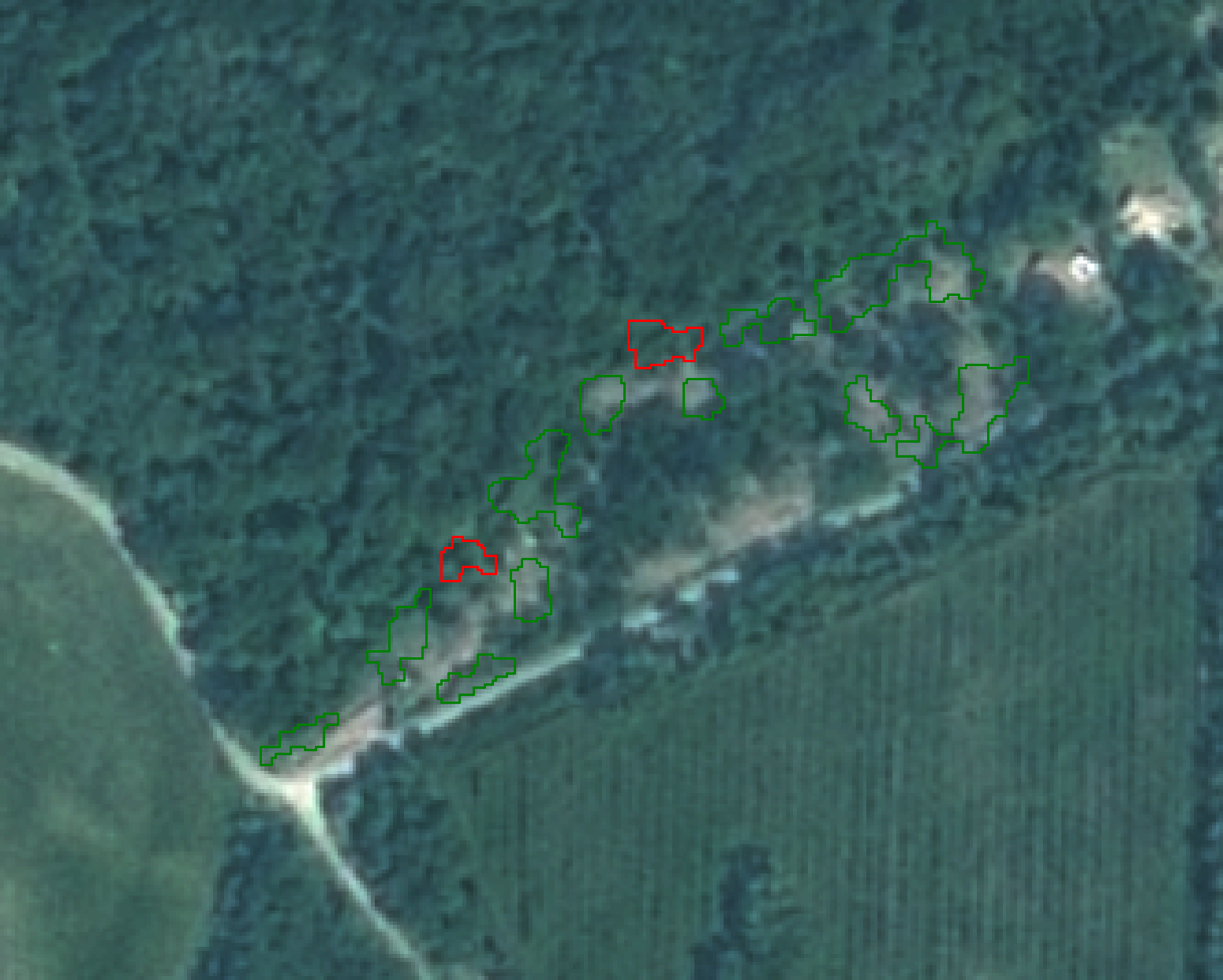}
        \caption{2023 VHR}
        \label{fig:vhr}
    \end{subfigure}
    &
    \begin{subfigure}{0.29\textwidth}
        \centering
        \includegraphics[width=\linewidth]{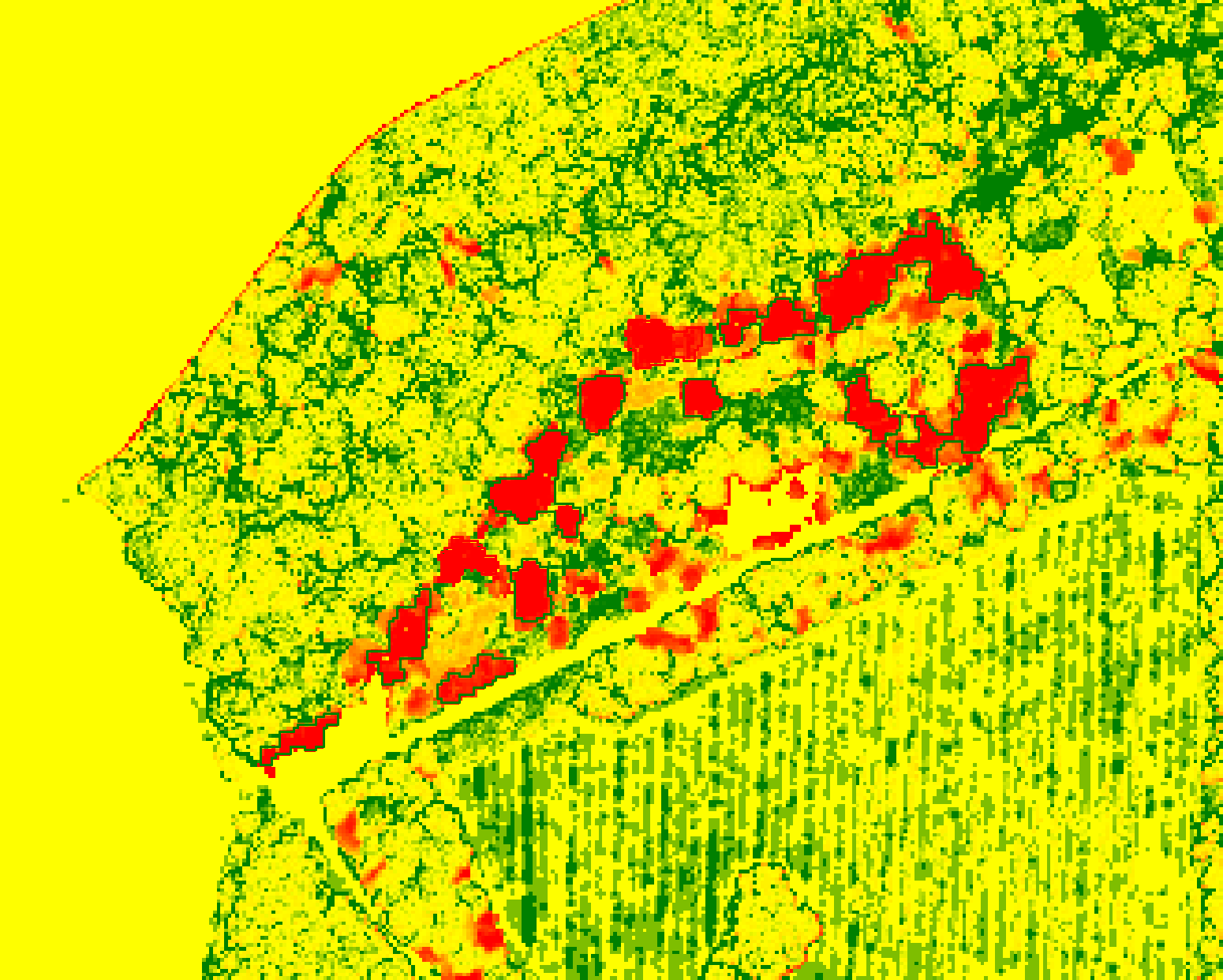}
        \caption{ALS-based change map}
        \label{fig:vhr}
    \end{subfigure}
    &
    \multirow{1}{*}[4.35cm]{
    \hspace{-0.5cm}
    \def\colormapheight{4.0cm}
\pgfplotsset{
    /pgfplots/colormap={treegrowth}{
    rgb255(0cm)=(255,0,0), 
    rgb255(1cm)=(255,125,111), 
    rgb255(2cm)=(255,255,0), 
    rgb255(3cm)=(130,192,0), 
    rgb255(4cm)=(0,130,0), 
    }
}

\begin{tikzpicture}
    \begin{axis}[
        hide axis,
        scale only axis,
        height=5pt,
        width=50pt,
        xlabel={in m},
        colormap name=treegrowth,
        colorbar,
            colorbar style={
            width=.2cm,
            height=\colormapheight,
            ytick={0,0.5,1,1.5,2,3,4},
            yticklabels={-20m,-15m,-10m,-5m,0m,+1m,+2m},
            yticklabel style={font=\tiny},
            major tick length=1.5pt,
            line width=.05mm,
            grid style={draw=none} 
        },
        point meta min=0,
        point meta max=4,
    ]
    \addplot [draw=none] coordinates {(0,0)};
    \end{axis}
 
\end{tikzpicture}
    }
    \\
    \multicolumn{4}{c}{
    \begin{tabular}{rl@{\qquad\qquad}rl}
        \raisebox{-.4\height}{\tikz \draw [ultra thick, green] (0,0) circle(5mm);}
        & validated
        &
        \raisebox{-.4\height}{\tikz \draw [ultra thick, red] (0,0) circle(5mm);}
        & rejected
    \end{tabular}
    }
    \end{tabular}
    \caption{{\bf Visual Validation of Change Components:} Example of a pair of successive VHR images and the corresponding change maps (derived from differences in ALS-based canopy height). We highlight the contours of the change masks validated by forestry experts through visual inspection.
    }
\label{fig:sup:change-curation}
\end{figure*}

\paragraph{Change Dataset Curation.}
To ensure the quality and accuracy of the Open-Canopy-$\Delta$ benchmark, we conducted a thorough manual validation of the dieback areas constituting its ground truth. As detailed in Section 4.1, each of the 73 change areas was carefully examined and validated by a forest expert. This meticulous process guarantees the reliability of the dataset for challenging computer vision tasks involving canopy height change detection. An example of visual annotation from this validation process is shown in \cref{fig:sup:change-curation}. Some false positives were identified, likely due to selective logging activities occurring between the ALS and SPOT acquisitions within the same year.


\section{Additional Results}
\label{sec:sup:results}
We present several additional analyses of the performance of our models. First, we provide additional qualitative illustrations in \cref{sec:sup:illustration}. Then, we offer a detailed analysis of how tree height influences the quality of the results
\cref{sec:sup:height}. Finally, we re-evaluate our models and other products at different resolutions (\cref{sec:sup:resolution}),  providing a fair comparison in settings more advantageous to coarser predictions.

\subsection{Qualitative Illustrations}
\label{sec:sup:illustration}
We provide here additional illustrations for qualitative assessment.

\begin{figure*}[!hp] 
    \centering
\centering
\begin{tabular}{p{3mm}c@{\,}c@{\,}c@{\,}cl}
 \raisebox{1cm}{1}&
\includegraphics[width=.2\textwidth]{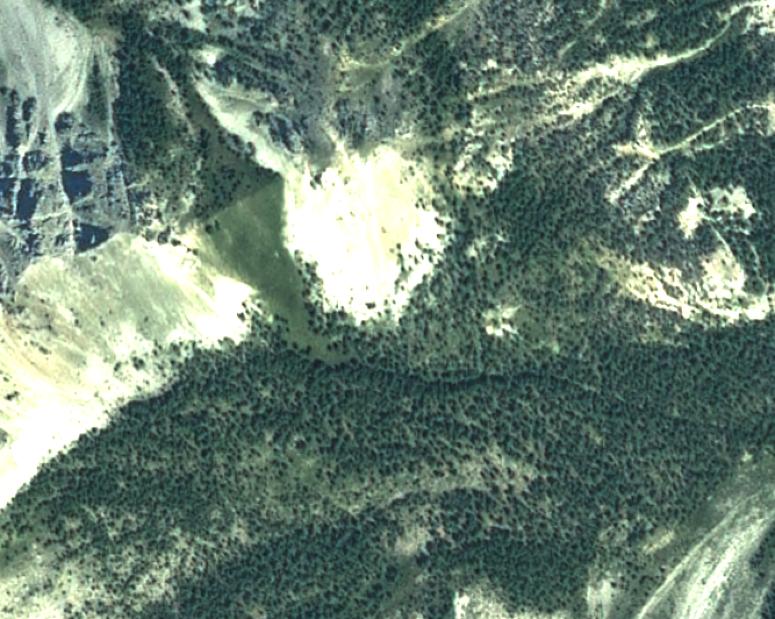}
     &  
\includegraphics[width=.2\textwidth]{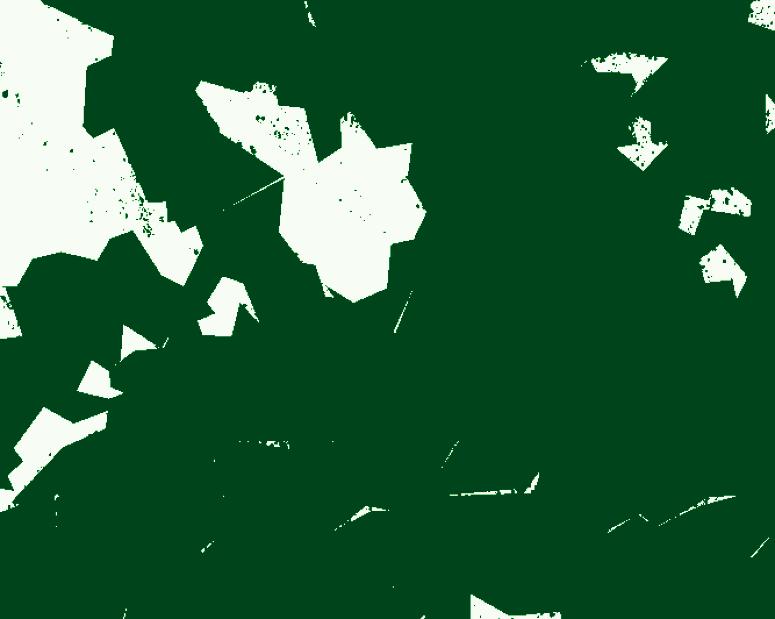}
     &
\includegraphics[width=.2\textwidth]{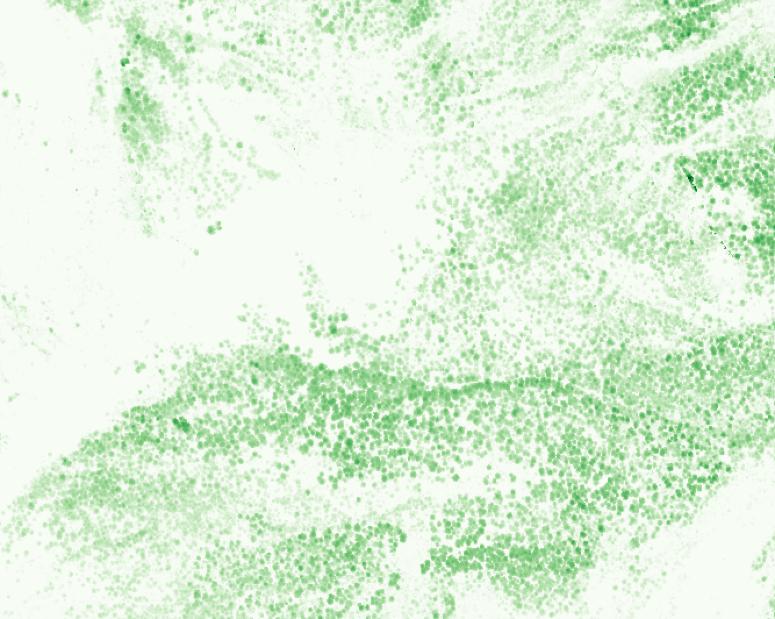}
     &
 \begin{tikzpicture}
    \node[anchor=south west,inner sep=0] (image) at (0,0) {\includegraphics[width=.2\textwidth]{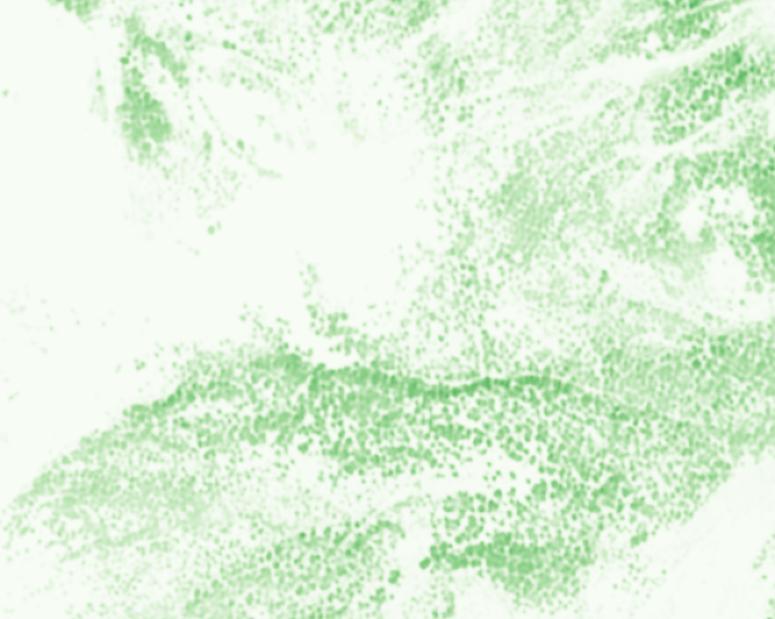}
    };
     \begin{scope}[x={(image.south east)},y={(image.north west)}]
     \node[fill=none, draw=none, text=black] (n1) at (0.9,0.89) {\includegraphics[width={0.025}\linewidth]{images/north.png}} ;
      \node[fill=none, draw=none, text=black] (n1) at (0.9,0.75) {\contour{white}{\scriptsize N}} ;
       \draw[-, fill=white, text=black, draw=white, ultra thick] (0.72,0.10) -- (0.93,0.10);
      \draw[<->, fill=white, text=black, draw=black,  thick] (0.70,0.10) -- (0.95,0.10);
      \draw[-, draw=black, thick] (0.825,0.10) -- (0.825,0.05);
      \node[fill=none, text=black, draw=none] at (0.825,0.18)  {\tiny  \contour{white}{\bf 250m}};
     \end{scope}
\end{tikzpicture}
&
\multirow{2}{*}[2.4cm]{
\hspace{-0.9cm}
\def\colormapheight{2.45cm}
\pgfplotsset{
    /pgfplots/colormap={treeheights}{
        rgb255(0cm)=(255,255,255), 
        rgb255(1cm)=(199,234,194), 
        rgb255(2cm)=(113,198,117), 
        rgb255(3cm)=(36,137,69), 
        rgb255(4cm)=(15,54,27)      
    }
}

\begin{tikzpicture}
    \begin{axis}[
        hide axis,
        scale only axis,
        height=5pt,
        width=50pt,
        xlabel={in m},
        colormap name=treeheights,
        colorbar,
           colorbar,
            colorbar style={
            width=.2cm,
            height=\colormapheight,
            ytick={0,1,2,3,4},
            yticklabels={0m,10m,20m,30m,40m},
            yticklabel style={font=\tiny},
            major tick length=1.5pt, 
            line width=.05mm,
            grid style={draw=none} 
        },
        point meta min=0,
        point meta max=4
    ]
    \addplot [draw=none] coordinates {(0,0)};
    \end{axis}
 
\end{tikzpicture}
}
\\
 \raisebox{1cm}{2}&
\includegraphics[width=.2\textwidth]{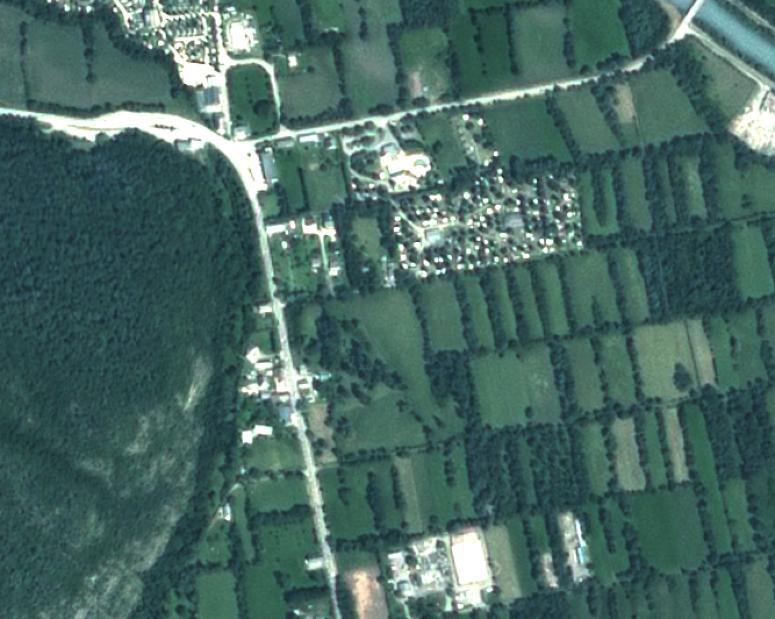}
     &  
\includegraphics[width=.2\textwidth]{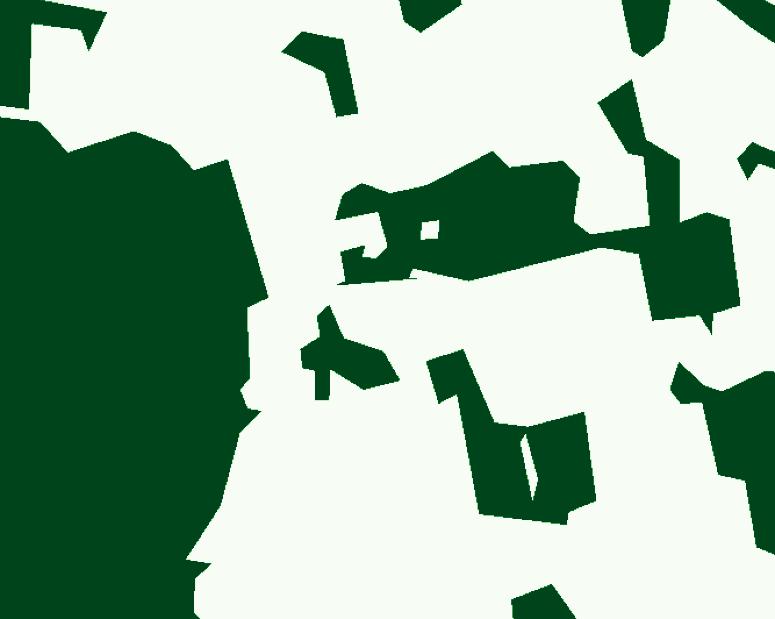}
     &
\includegraphics[width=.2\textwidth]{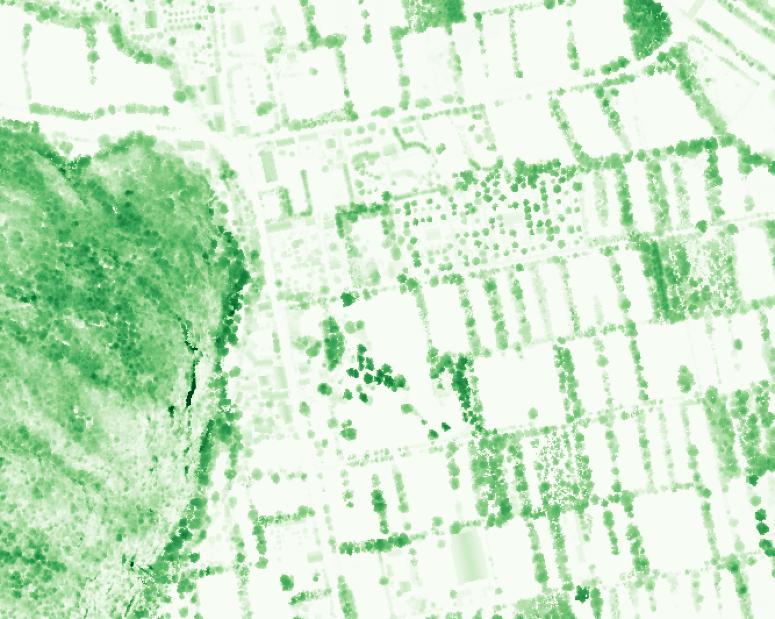}
     &
\includegraphics[width=.2\textwidth]{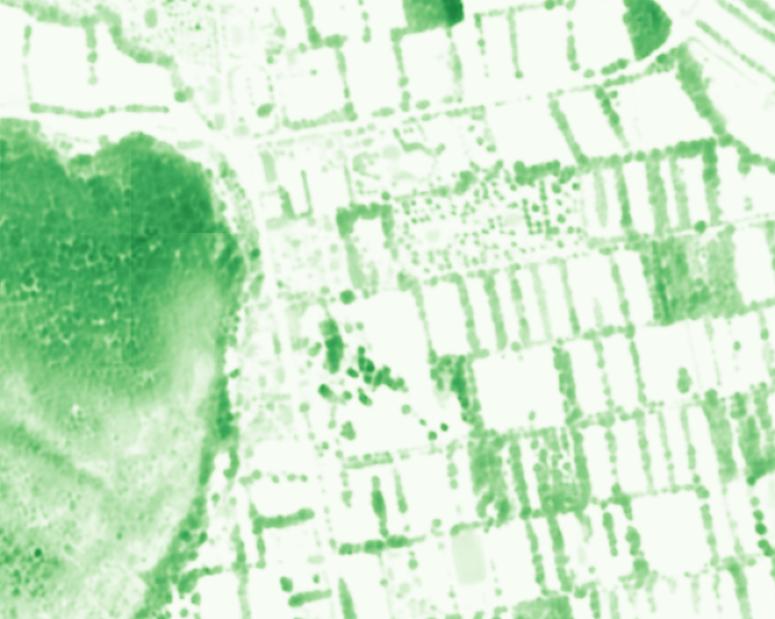}
\\
 \raisebox{1cm}{3}&
\includegraphics[width=.2\textwidth]{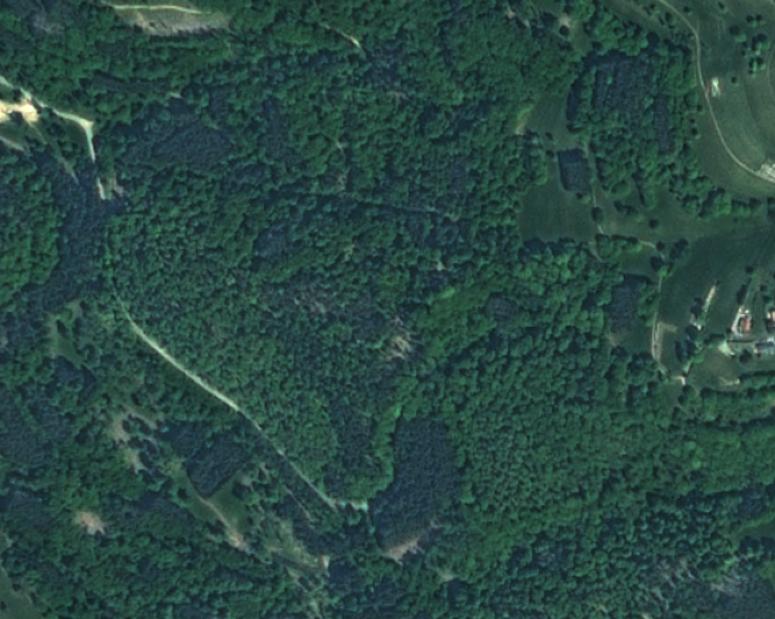}
     &  
\includegraphics[width=.2\textwidth]{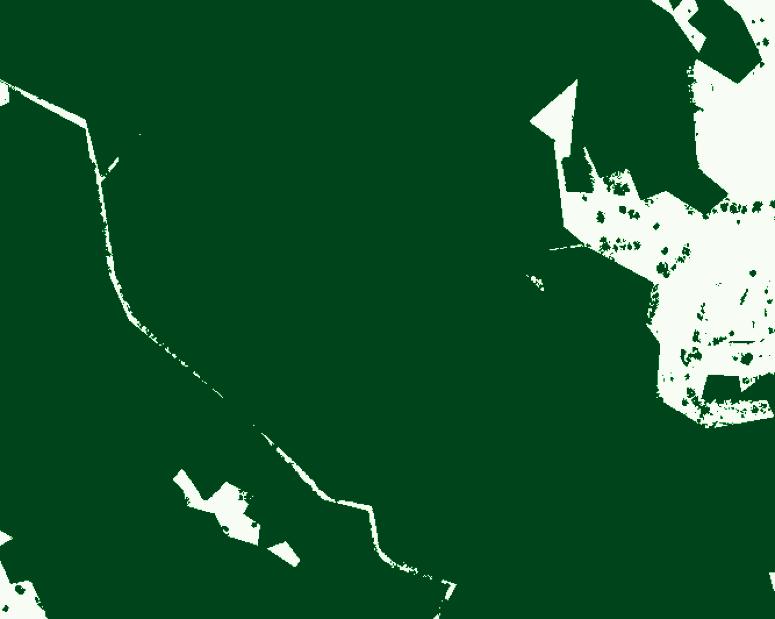}
     &
\includegraphics[width=.2\textwidth]{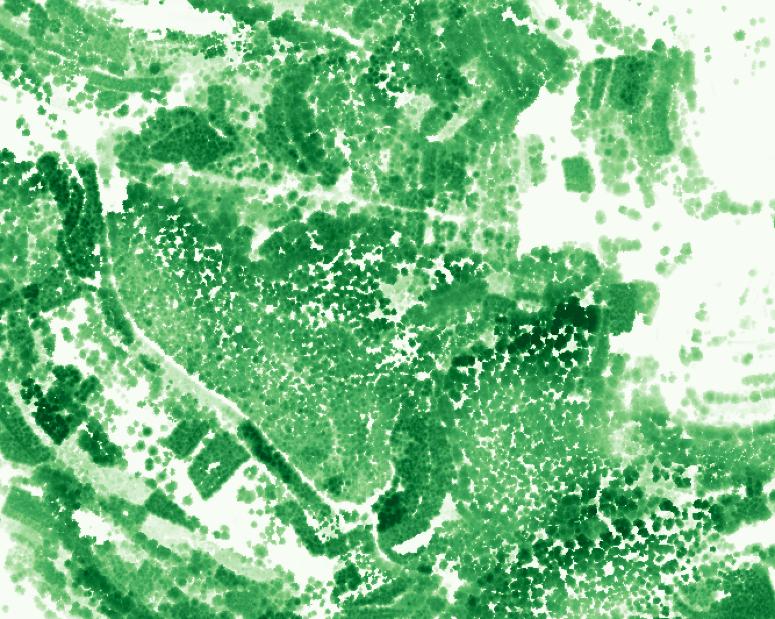}
     &
\includegraphics[width=.2\textwidth]{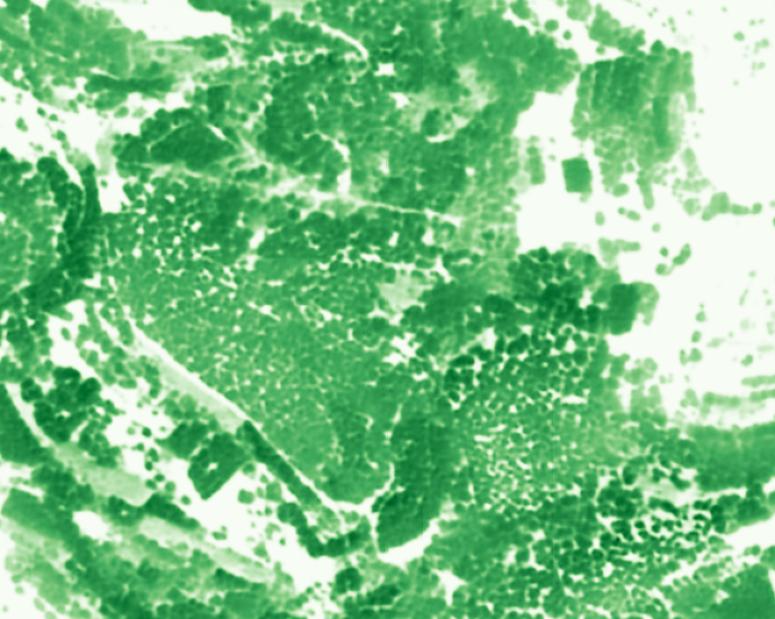}
\\
 \raisebox{1cm}{4}&
\includegraphics[width=.2\textwidth]{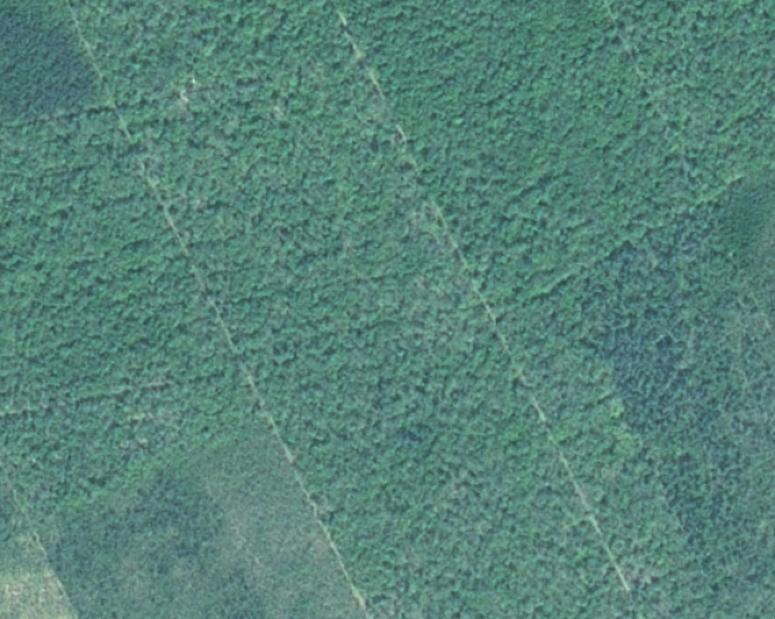}
     &  
\includegraphics[width=.2\textwidth]{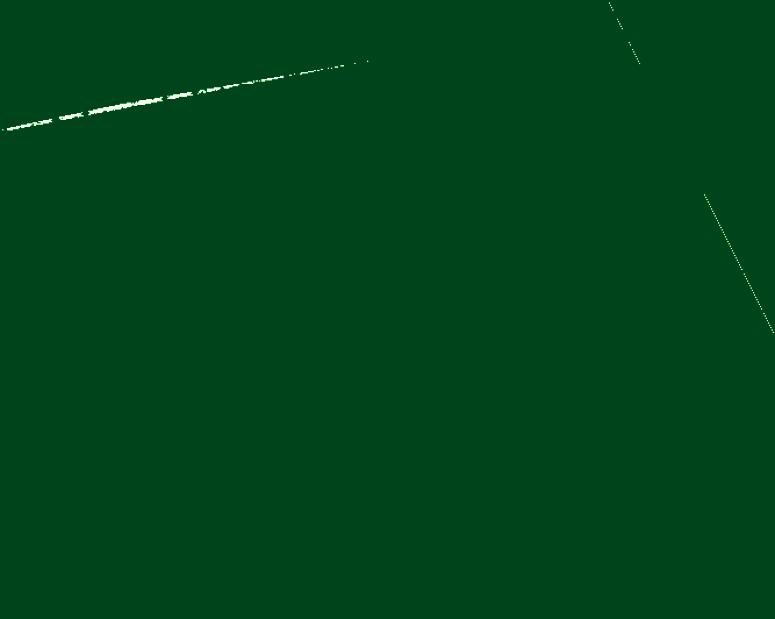}
     &
\includegraphics[width=.2\textwidth]{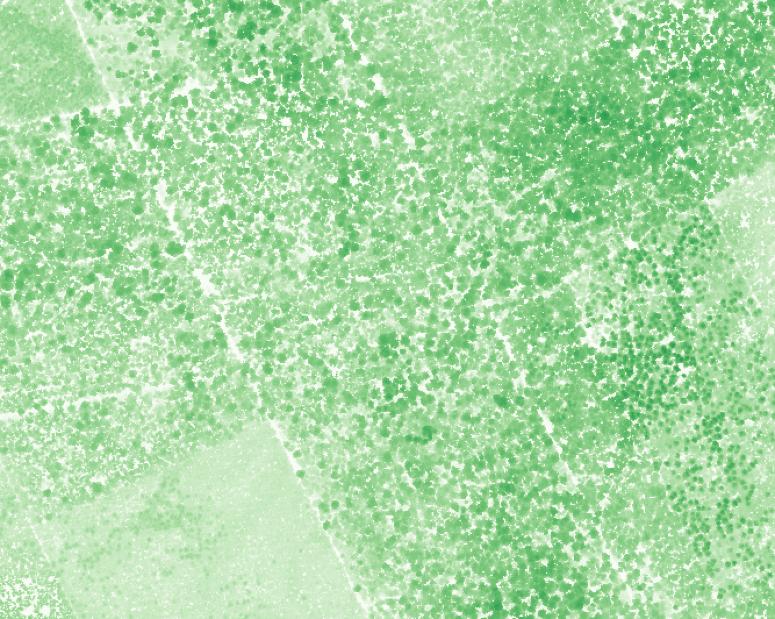}
     &
\includegraphics[width=.2\textwidth]{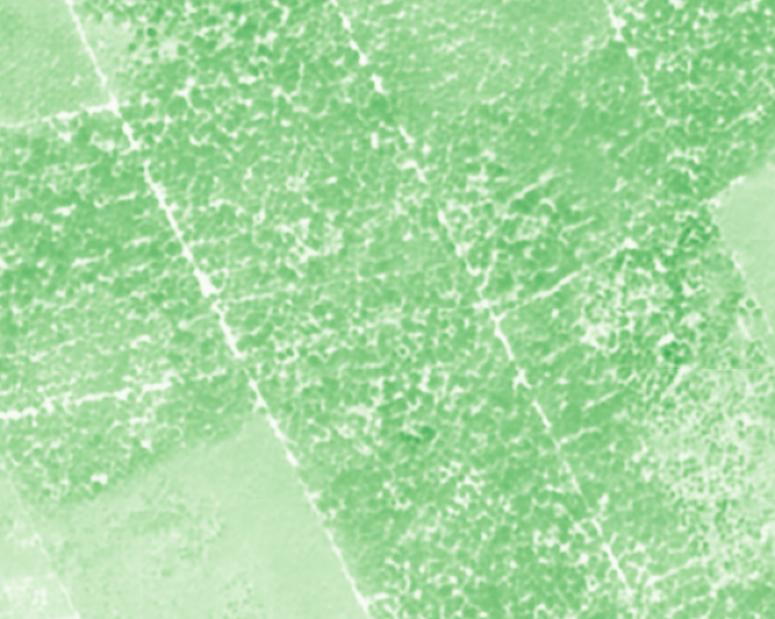}
&
\\
 \raisebox{1cm}{5}&
\includegraphics[width=.2\textwidth]{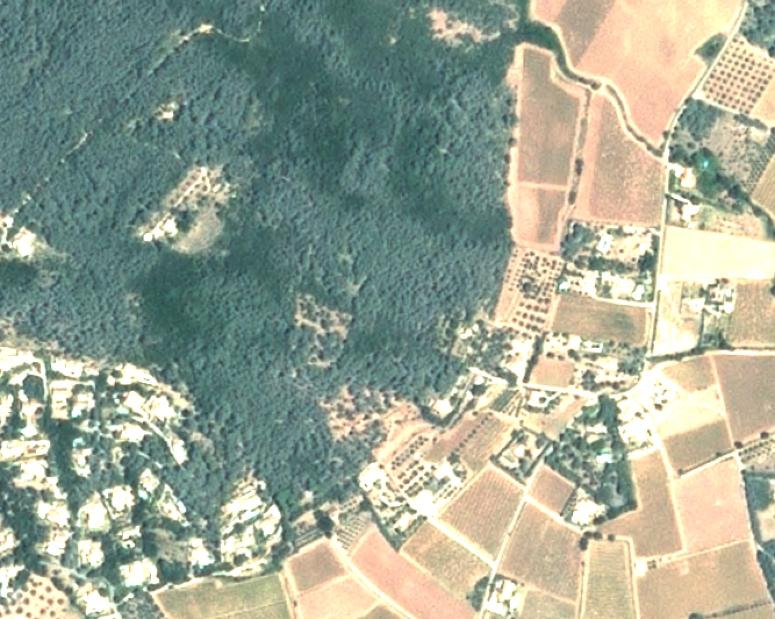}
     &  
\includegraphics[width=.2\textwidth]{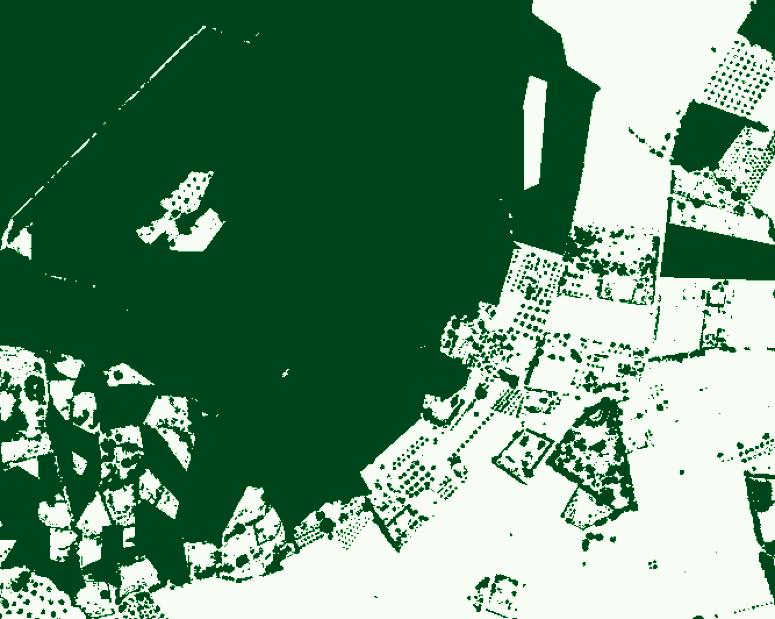}
     &
\includegraphics[width=.2\textwidth]{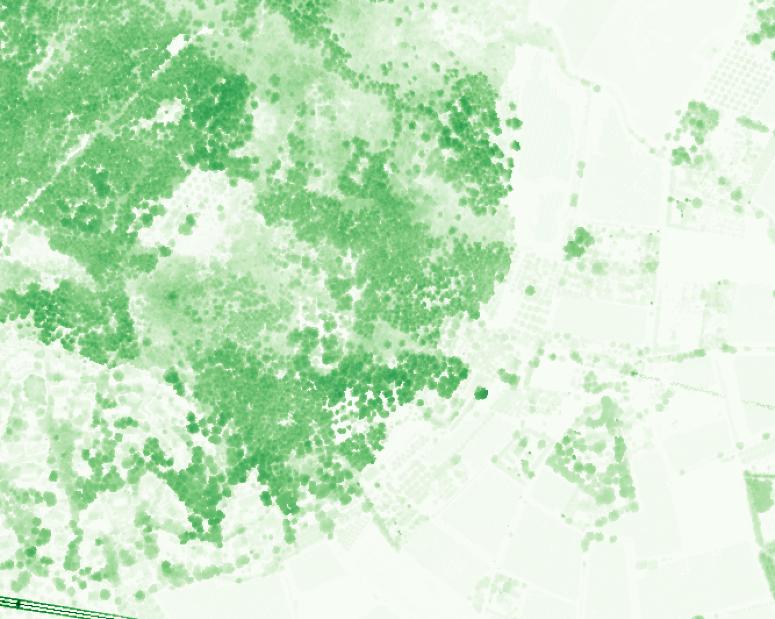}
     &
\includegraphics[width=.2\textwidth]{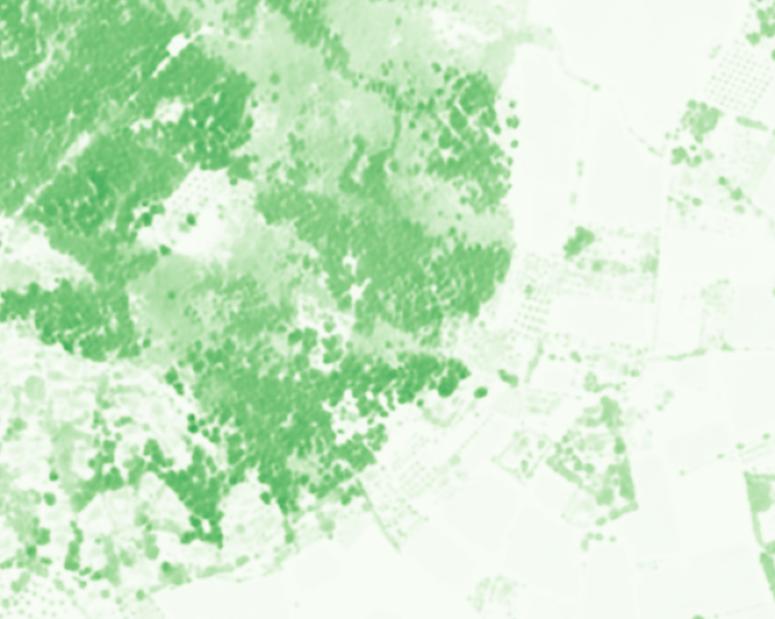}

\\
 \raisebox{1cm}{6}&
\includegraphics[width=.2\textwidth]{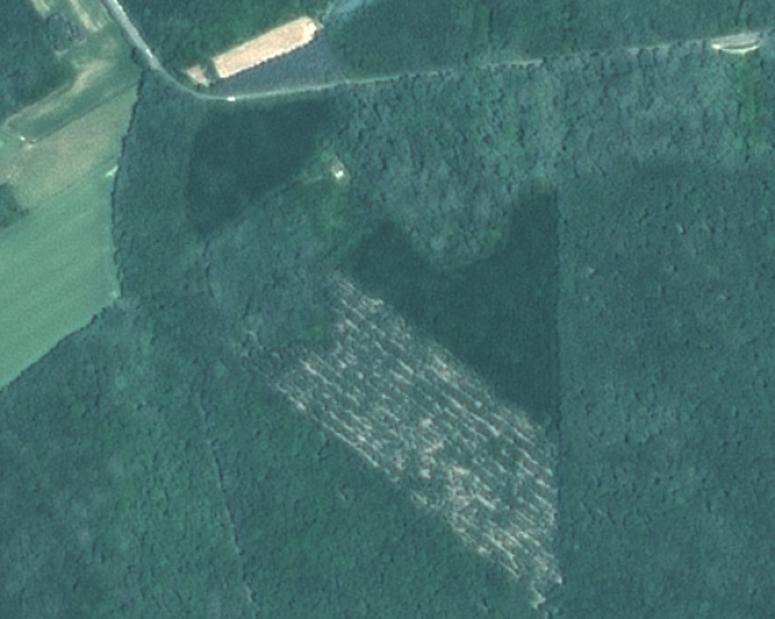}
     &  
\includegraphics[width=.2\textwidth]{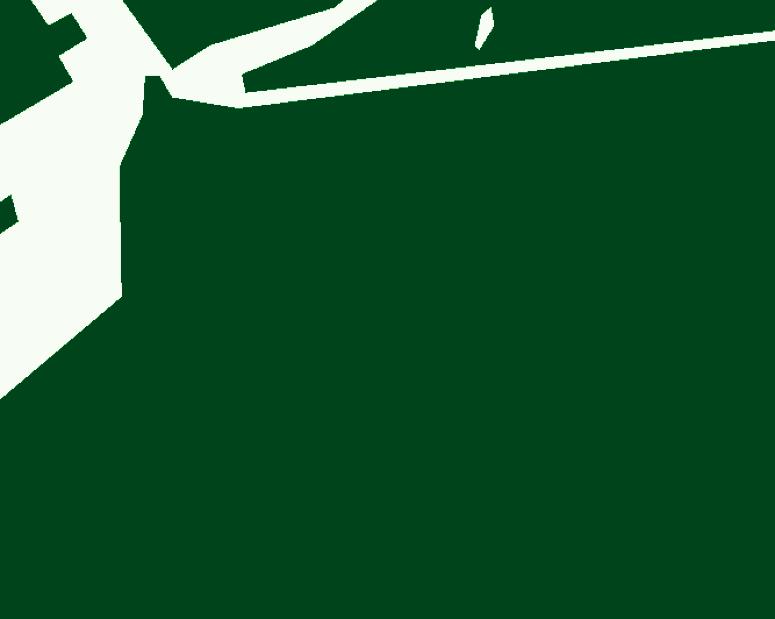}
     &
\includegraphics[width=.2\textwidth]{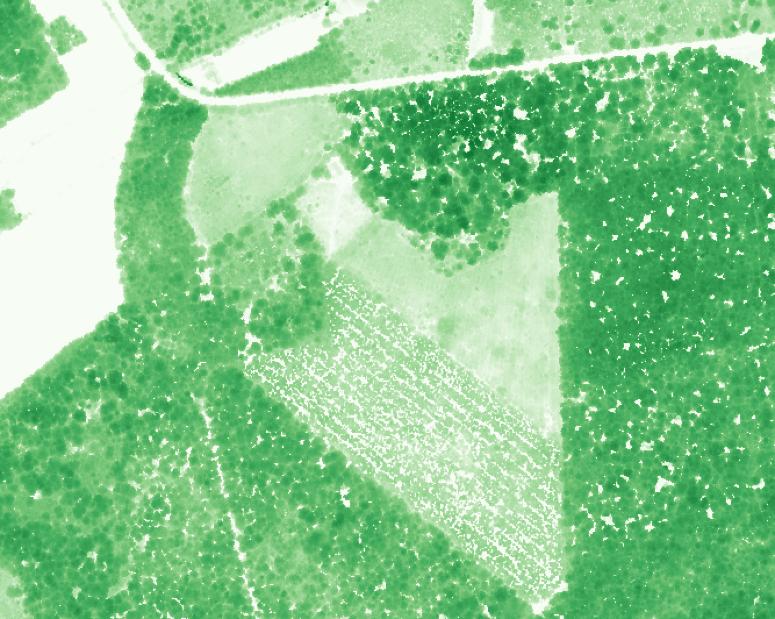}
     &
\includegraphics[width=.2\textwidth]{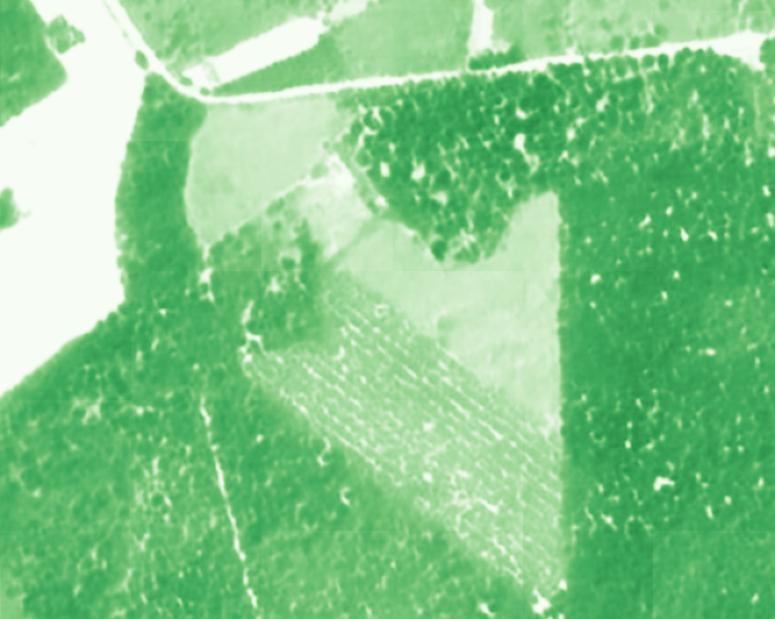}

\\
 \raisebox{1cm}{7}&
\includegraphics[width=.2\textwidth]{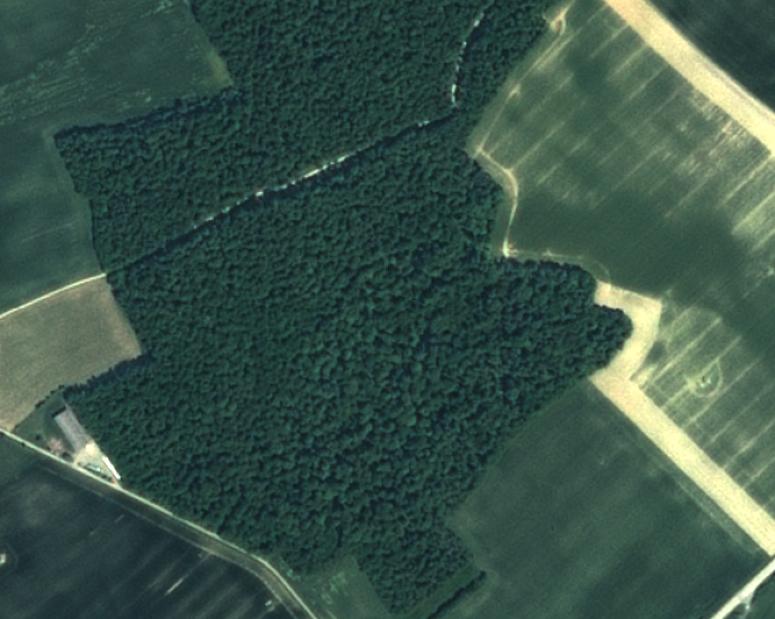}
     &  
\includegraphics[width=.2\textwidth]{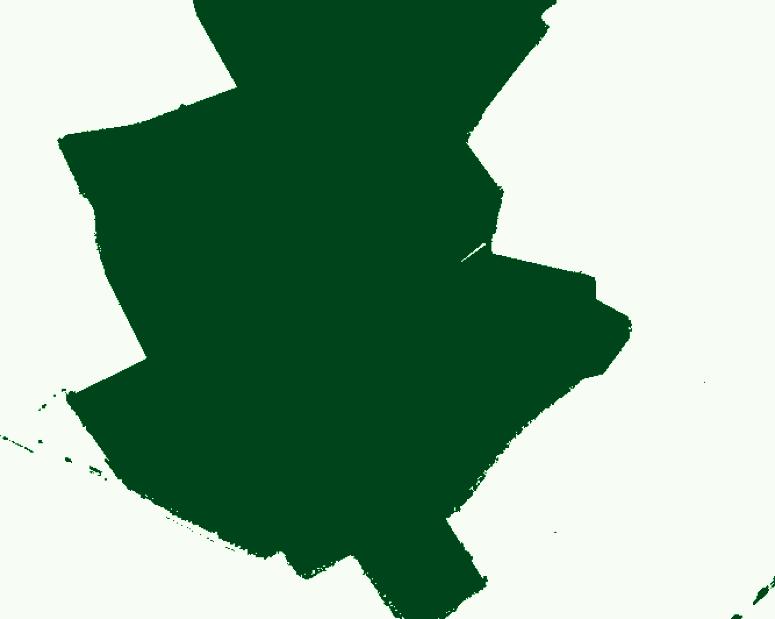}
     &
\includegraphics[width=.2\textwidth]{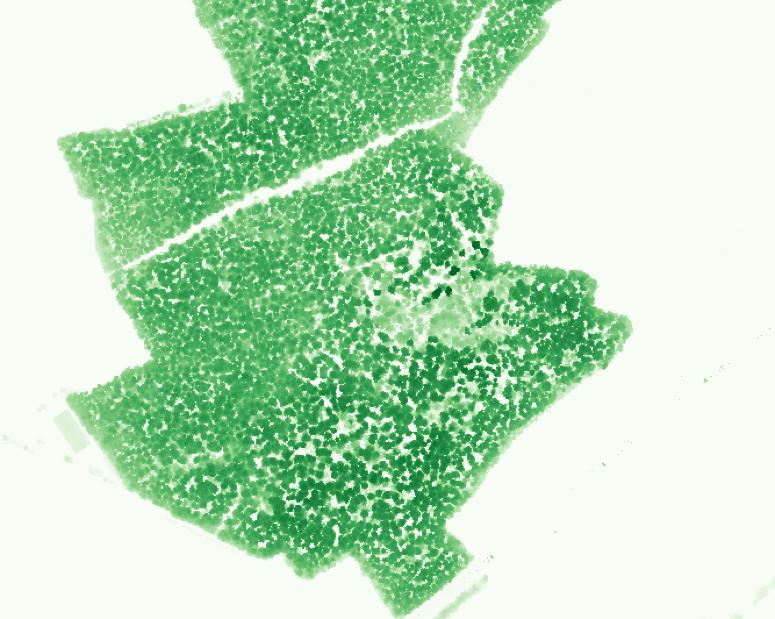}
     &
\includegraphics[width=.2\textwidth]{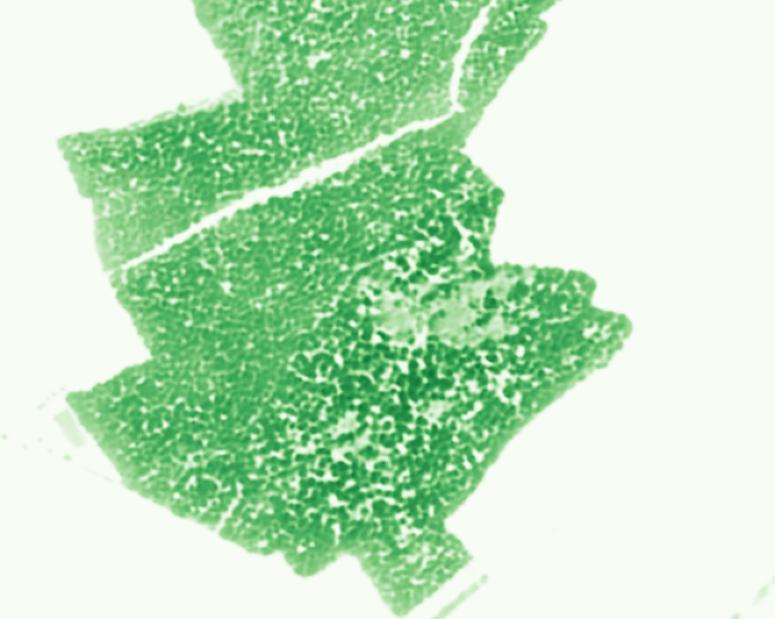}
\\&
\begin{subfigure}{0.225\linewidth}
\caption{VHR image}
\label{fig:supmat-height:a}
\end{subfigure}
&
\begin{subfigure}{0.225\linewidth}
\caption{Vegetation mask}
\label{fig:supmat-height:b}
\end{subfigure}
&
\begin{subfigure}{0.225\linewidth}
\caption{ALS height map}
\label{fig:supmat-height:c}
\end{subfigure}
&
\begin{subfigure}{0.225\linewidth}
\caption{PVTv2 height map }
\label{fig:supmat-height:d}
\end{subfigure}
\end{tabular}

    \caption{{\bf Canopy Height Estimation Illustrations.} We select seven areas of interest and represent the available VHR image (\subref{fig:supmat-height:a}), the vegetation mask used for evaluation (\subref{fig:supmat-height:b}), the ground truth ALS-derived height map (\subref{fig:supmat-height:c}), and the height map estimated with PVTv2 model from the VHR image (\subref{fig:supmat-height:d}). Scale and orientation are shared across all subfigures.}
    \label{fig:supmat-height}
\end{figure*}

\paragraph{Canopy Height} 
\cref{fig:supmat-height} showcases a comparison between the ALS-derived canopy height map and the height map predicted by our model using SPOT images. Our model demonstrates the ability to accurately estimate vegetation height across a variety of challenging scenarios:
\begin{compactitem}

\item {\bf Mountainous Areas} (first row): Capturing complex terrain and varied vegetation.

 \item {\bf  Agricultural Lands} (second row): Detecting small hedges and understory vegetation.
 
 \item {\bf  Dense Forests} (rows 3 and 4): Handling thick canopy cover and shadowed regions.
 
 \item {\bf  Urban Environments} (row 5): Distinguishing trees amidst buildings and infrastructure.
 
 \item {\bf  Mixed Scenes} (rows 6 and 7): Managing heterogeneous landscapes with multiple land cover types.
 
\end{compactitem}

The high spatial resolution of our predictions not only captures fine-grained details but also enables the identification of man-made features such as forest paths, which are crucial for forest management applications.

\begin{figure*}[t] 
    \centering
        \centering
    \begin{tabular}{c@{\;}c@{\;}c@{\;}c@{\;}l}
     \begin{subfigure}{0.21\textwidth}
        \centering
        \includegraphics[width=\linewidth]{images/spot.jpg}
        \caption{VHR image}
        \label{fig:vhr}
    \end{subfigure}
    &
         \begin{subfigure}{0.21\textwidth}
        \centering
        \includegraphics[width=\linewidth]{images/vit_abs.jpg}
        \caption{Vit-B Error, MAE=8.0}
        \label{fig:unet:mae}
    \end{subfigure}
    &
    \begin{subfigure}{0.21\textwidth}
        \centering
        \includegraphics[width=\linewidth]{images/pvtv2_abs.jpg}
        \caption{{PVTv2 Error, MAE=3.9}}
        \label{fig:pvt:mae}
    \end{subfigure}
    &
    \begin{subfigure}{0.21\textwidth}
        \centering
        \includegraphics[width=\linewidth]{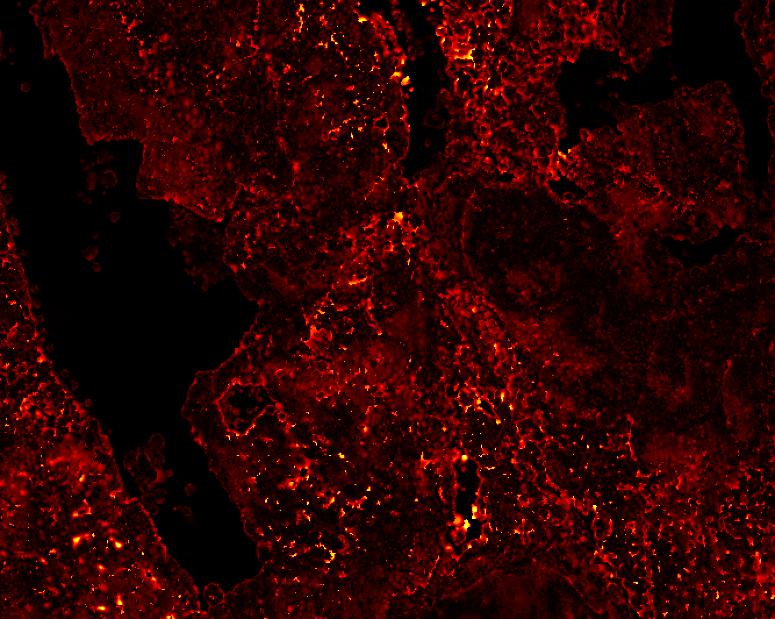}
        \caption{SWIN Error, MAE=4.1}
        \label{fig:usat:mae}
    \end{subfigure}
    &
    \multirow{1}{*}[3.1cm]{
    \hspace{-0.5cm}
    \def\colormapheight{2.51cm}
\pgfplotsset{
    /pgfplots/colormap={heat}{
        rgb255(0cm)=(0,0,0),       
        rgb255(1cm)=(128,0,0),     
        rgb255(2cm)=(255,0,0),     
        rgb255(3cm)=(255,255,0),   
        rgb255(4cm)=(255,255,255)  
    }
}

\begin{tikzpicture}
    \begin{axis}[
        hide axis,
        scale only axis,
        height=5pt,
        width=50pt,
        xlabel={in m},
        colormap name=heat,
        colorbar,
           colorbar,
            colorbar style={
            width=.2cm,
            height=\colormapheight,
            ytick={0,1,2,3,4},
            yticklabels={0m,10m,20m,30m,40m},
            yticklabel style={font=\tiny},
            major tick length=1.5pt, 
            line width=.05mm,
            grid style={draw=none} 
        },
        point meta min=0,
        point meta max=4
    ]
    \addplot [draw=none] coordinates {(0,0)};
    \end{axis}
 
\end{tikzpicture}
    }
    \\
    \begin{subfigure}{0.21\textwidth}
        \centering
        \includegraphics[width=\linewidth]{images/als_height.jpg}
        \caption{ALS-derived height}
        \label{fig:als}
    \end{subfigure}
        &
    \begin{subfigure}{0.21\textwidth}
        \centering
        \includegraphics[width=\linewidth]{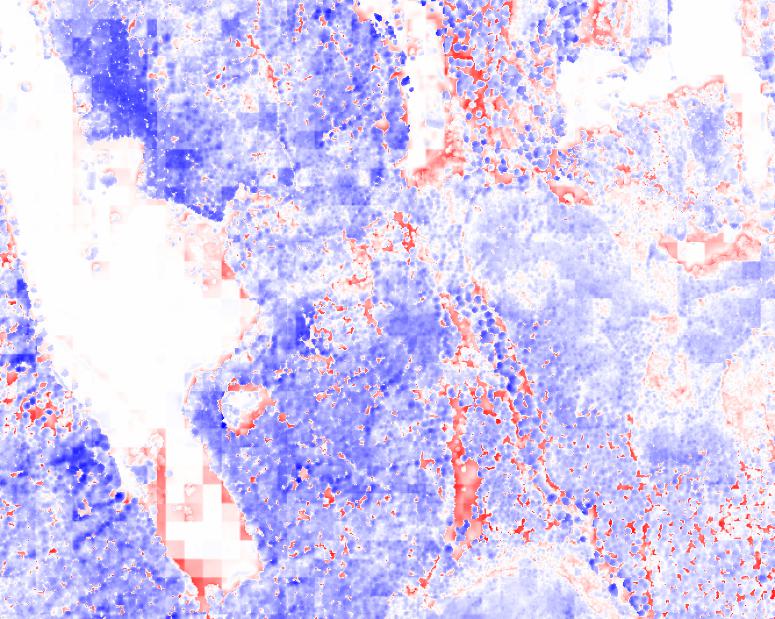}
        \caption{Vit-B Error}
        \label{fig:unet:mae}
    \end{subfigure}
    &
    \begin{subfigure}{0.21\textwidth}
        \centering
        \includegraphics[width=\linewidth]{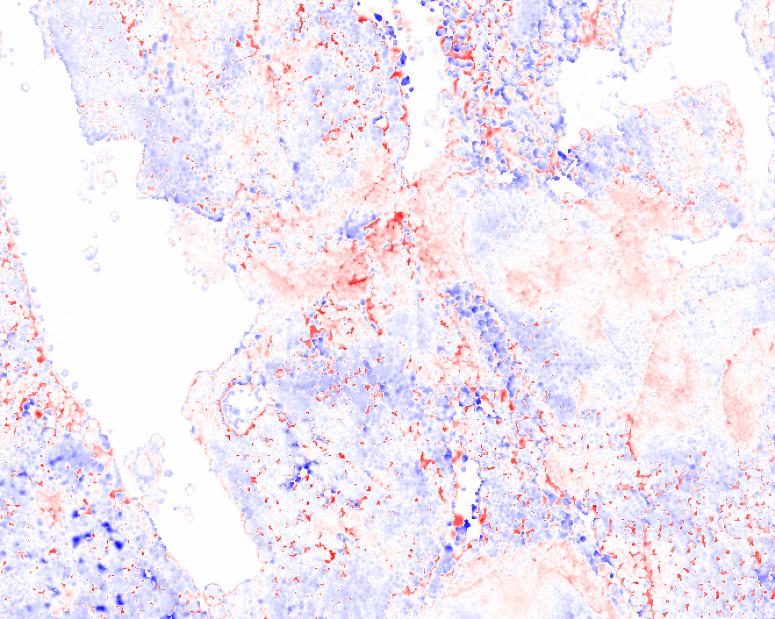}
        \caption{PVTv2 Error}
        \label{fig:pvt:mae}
    \end{subfigure}
    &
    \begin{subfigure}{0.21\textwidth}
        \centering
          \begin{tikzpicture}
    \node[anchor=south west,inner sep=0] (image) at (0,0) {\includegraphics[width=1\textwidth]{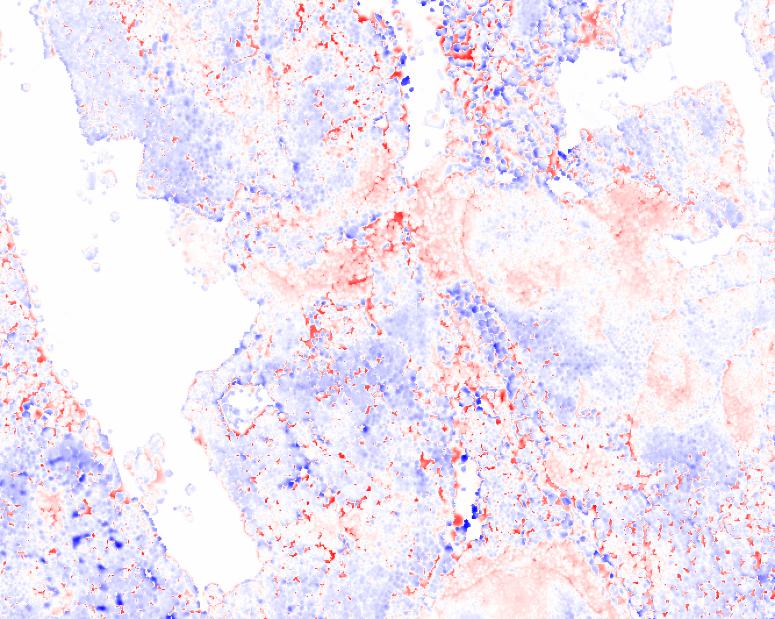}};
     \begin{scope}[x={(image.south east)},y={(image.north west)}]
     \node[fill=none, draw=none, text=black] (n1) at (0.9,0.89) {\includegraphics[width={0.1}\textwidth]{images/north.png}} ;
      \node[fill=none, draw=none, text=black] (n1) at (0.9,0.75) {\contour{white}{\scriptsize N}} ;
       \draw[-, fill=white, text=black, draw=white, ultra thick] (0.72,0.10) -- (0.93,0.10);
      \draw[<->, fill=white, text=black, draw=black,  thick] (0.70,0.10) -- (0.95,0.10);
      \draw[-, draw=black, thick] (0.825,0.10) -- (0.825,0.05);
      \node[fill=none, text=black, draw=none] at (0.825,0.18)  {\tiny  \contour{white}{\bf 250m}};
     \end{scope}
\end{tikzpicture}

        \caption{SWIN Error}
        \label{fig:usat:mae}
    \end{subfigure}
    &
    \multirow{1}{*}[3.1cm]{
    \hspace{-0.5cm}
    \def\colormapheight{2.51cm}\pgfplotsset{
    /pgfplots/colormap={seismic}{
        rgb255(0cm)=(0,0,255),     
        rgb255(2cm)=(255,255,255), 
        rgb255(4cm)=(255,0,0)      
    }
}

\begin{tikzpicture}
    \begin{axis}[
        hide axis,
        scale only axis,
        height=5pt,
        width=50pt,
        xlabel={in m},
        colormap name=seismic,
        colorbar,
           colorbar,
            colorbar style={
            width=.2cm,
            height=\colormapheight,
            ytick={0,1,2,3,4},
            yticklabels={-30m,-15m,0m,15m,30m},
            yticklabel style={font=\tiny},
            major tick length=1.5pt, 
            line width=.05mm,
            grid style={draw=none} 
        },
        point meta min=0,
        point meta max=4
    ]
    \addplot [draw=none] coordinates {(0,0)};

    \end{axis}
 
\end{tikzpicture}
    }
    \end{tabular}
    \vspace{-2mm}
    \caption{{\bf Difference Maps:} Per-pixel absolute (top row) and relative (bottom row) errors for three models: ViT-B, PVTv2, and SWIN. While the differences between PVTv2 and SWIN are subtle (approximately 20cm on average), the advantage of these models over ViT-B is visible.}
    \label{fig:qualitative}
\end{figure*}

We further compare the performance of three models in\cref{fig:qualitative}: a standard Vision Transformer (ViT) and two hierarchical models, PVTv2 and SWIN. The hierarchical models exhibit significantly lower errors, which corroborates our quantiative results.

\paragraph{Canopy Height Change}
We provide additional illustrations of height change detection in \cref{fig:supmat-change}. While our model tends to over predict small growth or loss of canopy height, the areas of strong disturbances---as denoted by our smoothed and filtered binary change maps---are overall well detected and delineated. Our illustration covers areas of dense forests (first row) and mixed scenes (row 2 and 3). Our method can detect disturbances such as clear and selective cuts.

Note that the Sentinel-derived height maps for 2022 and 2023 were provided by the authors of \citep{schwartz2023forms}, as only the map for 2020 is available online.

\subsection{Influence of Tree Height}

\label{sec:sup:height}
We analyzed the performance of our canopy height estimation model across different ranges of true tree heights to understand how tree height influences prediction accuracy. The results are summarized in \cref{tab:metrics-per-bin}.

\begin{compactitem}
\item Note that the nMAE (normalized Mean Absolute Error) is computed for all ranges as the average of the pixel-wise normalized absolute error:
\begin{align}
    \text{nMAE}=\frac{\left| (z_\text{true}-z_\text{pred}) \right|}{1 + z_\text{true}}~,
\end{align}
 where $z_\text{true}$ and $z_\text{pred}$ are respectively the ALS-derived and predicted height for a given pixel. The additional $1$ term in the denominator makes this measure more robust for pixels corresponding to low vegetation.
 \item When computing the nMAE for the overall range of 0–60 m, we exclude the 0–2 m bin. This exclusion is necessary because values in this range can produce disproportionately large errors due to the normalization, which can dominate the metric and skew the results. Additionally, including this bin may unfairly disadvantage models with lower spatial resolutions that aim to predict the highest value within larger pixels, potentially overlapping with bare soil at higher resolutions. 
\end{compactitem}

As shown in \cref{tab:metrics-per-bin} by the bias of our model for different ranges, our model tends to over-predict the height of small trees and under-predict the height of tall trees. While the average error is higher for larger trees, our model has the lowest nMAE for the 20-30m range, with a value of $12.1\%$.

\begin{table*}  
    \caption{{\bf Canopy Height Prediction Per Height Bins.} We report the metrics for different bins of true tree height for the PVTv2\citep{wang2021pvtv2} model.}
    \label{tab:metrics-per-bin}
    \centering
\begin{tabular}{l m{1cm}m{1cm}m{1cm}m{1cm}m{1cm}m{1cm}m{1cm}c}
\toprule
 {Range in m} &  {0-2} & {2-5} &{5-10} & {10-15} & {15-20} & {20-30} & {30-60} & \textbf{0-60} \\
\midrule
MAE in m
& \hphantom{1}1.67	& 2.29&	2.65&	\hphantom{1}2.70&	\hphantom{-1}2.61	&\hphantom{-1}3.00&	\hphantom{-1}5.52&	\hphantom{-}\textbf{2.52}
\\
nMAE in \%
& 138.8 & 53.6&	32.1&	\hphantom{1}20.3&	\hphantom{-1}14.3&	\hphantom{-1}12.1&	\hphantom{-1}16.0&	\hphantom{-}\textbf{22.9}
\\
RMSE in m
& \hphantom{1}4.31	&3.67&	3.69&	\hphantom{1}3.60&	\hphantom{-1}3.53&	\hphantom{-1}4.19&	\hphantom{-1}7.56&	\hphantom{-}\textbf{4.02}
\\
Bias in m
& \hphantom{1}1.49	&0.87	&0.65&	\hphantom{1}0.21&	\hphantom{1}-0.42&	\hphantom{1}-1.90&	\hphantom{1}-5.31&	\hphantom{-}\textbf{0.00}
\\
Tree cov. IoU (\%)
&\hphantom{13.}-&	72.6&	96.5&	\hphantom{1}99.3&	\hphantom{-1}99.7&	\hphantom{-1}99.8&	\hphantom{-1}99.6&	\hphantom{-}\textbf{90.5}
\\
\bottomrule
\end{tabular}
\end{table*}

\subsection{Evaluation at a resolution of 10m}
\label{sec:sup:resolution}
\begin{table*}[t]   
    \caption{{\bf Canopy Height Prediction at 10m resolution.} We resample all ground truth and predicted maps on a $10$ m grid.}
    \label{tab:metrics-10m}
    \centering
\begin{tabular}{l lc m{1cm}m{1cm}m{1cm}m{1cm}c}
\toprule
 \multirow{2}{*}{Map} & \multirow{2}{*}{Backbone} & {Initial res.} &  {MAE} & {nMAE} &{RMSE} & {Bias} &{Tree cov.}\\
&& in m &  {in m}&   {in \%} &{in m}&{in m} &{IoU in \%} 
  \\\midrule
  Potapov \citep{potapov2021mapping} & UNet & 30 &6.17 &	44.6&	\hphantom{1}8.33&	-3.31&	80.2\\
  \rowcolor{gray!10} Schwartz \citep{schwartz2023forms, schwartz2023mapping}& UNet & 10 
  & 4.00&	26.9&	\hphantom{1}5.28&	-1.38&	90.1
  \\
  Lang \citep{lang2023high}    & CNN & 10 & 8.64&	92.9&	29.25&	\hphantom{-}6.27&	90.1 \\
  \rowcolor{gray!10} Pauls \citep{pauls2024estimating}    & UNet & 10 & 4.59&	32.9&	\hphantom{1}5.96&	\hphantom{-}0.34&	90.1
  	
  \\\greyrule
   Liu \citep{liu2023overlooked}     & UNet & 3.0 & 4.58	&37.4&	10.97&	-1.26&	88.2
  \\
  \rowcolor{gray!10} Tolan \citep{tolan2024very} & ViT-L   & 1.0 & 6.10&	42.1&	\hphantom{1}7.95&	-5.37&	81.6
  \\\greyrule
   Open-Canopy &  UNet &  1.5 
  & 2.72 &	19.0&	\hphantom{1}3.95&	-2.06&	\textbf{93.4}
  \\
  \rowcolor{gray!10} {Open-Canopy} &  {PVTv2} &  {1.5} & \textbf{2.42} & \textbf{17.6} &	\hphantom{1}\textbf{3.57}&	\textbf{-1.69}&	93.3
  \\
\bottomrule
\end{tabular}
\end{table*}

To provide a fair comparison with models predicting canopy height at a 10 m resolution, we resampled both our ground truth and predicted height maps to a 10 m grid and re-evaluated all available models. We performed this by aggregating the higher-resolution data as follows:

For each 10 m pixel, we took the maximum value from the overlapping 1.5 m pixels. This approach is equivalent to rasterizing the full ALS 3D point cloud directly onto a 10 m grid.
Taking the maximum value aligns with models trained to predict metrics like GEDI RH100 or RH95 (relative height at the 100th or 95th percentile), which represent the tallest canopy elements within a pixel.

We report  the results in \cref{tab:metrics-10m}, and observe a similar ordering than in Table 3 of the main paper.  All methods see improved metrics as the problem is simpler, except for Tolan \etal. In particular, the tree coverage problem becomes significantly easier at this resolution, with all $10$ m-resolution methods nearing $90$\% IoU.
Note that the height map of \citep{liu2023overlooked} at a resolution of 3m was provided directly by the authors and is not available online.

\section{Ablation Study }
\label{sec:sup:ablation}
We propose an analysis of the influence of several of our hyperparameters and design choices.

\subsection{Parameters of the Change Detection}
\label{sec:sup:change}

We evaluate how different configurations of the ground truth binary change map affect canopy height change detection. Specifically, we examine:
(i) Minimum Height Difference: The threshold for considering a pixel as having a significant change in canopy height;
(ii) Minimum Contiguous Change Area: The smallest area of connected changed pixels considered significant.

\begin{figure*}[t]
    \centering
\centering
\begin{tabular}{c@{\,}c@{\,}c@{\,}c@{\,}c@{\,}c}
\begin{subfigure}{.16\textwidth}
  \begin{tikzpicture}
    \node[anchor=south west,inner sep=0] (image) at (0,0) {\includegraphics[width=\textwidth]{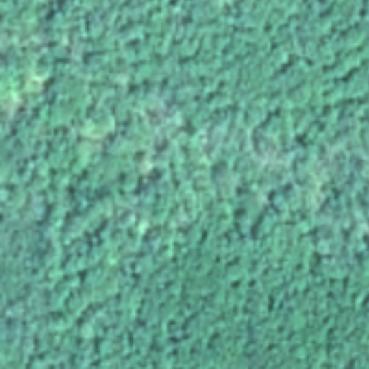}
    };
     \begin{scope}[x={(image.south east)},y={(image.north west)}]
     \node[fill=none, draw=none, text=black] (n1) at (0.145,0.89) {\includegraphics[width={0.1}\textwidth]{images/north.png}} ;
      \node[fill=none, draw=none, text=black] (n1) at (0.145,0.75) {\contour{white}{\scriptsize N}} ;
      \draw[-, fill=white, text=black, draw=white, ultra thick] (0.07,0.10) -- (0.36,0.10);
      \draw[<->, fill=white, text=black, draw=black,  thick] (0.05,0.10) -- (0.38,0.10);
      \draw[-, draw=black, thick] (0.215,0.10) -- (0.215,0.05);
      \node[fill=none, text=black, draw=none] at (0.215,0.18)  {\tiny  \contour{white}{\bf 100m}};
     \end{scope}
\end{tikzpicture}
\caption{\scriptsize VHR year 2022 }
\label{fig:supmat-change:spot1}
\end{subfigure}
&
\begin{subfigure}{.16\textwidth}
\includegraphics[width=\textwidth]{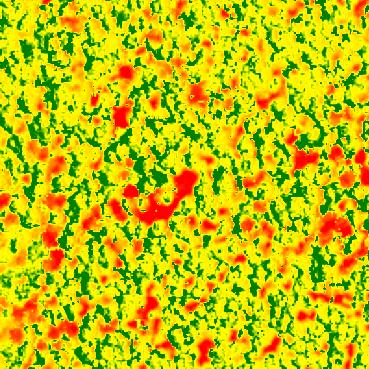}
\caption{\scriptsize  ALS change map }
\label{fig:supmat-change:alsdiff}
\end{subfigure}
&
\begin{subfigure}{.16\textwidth}
\includegraphics[width=\textwidth]{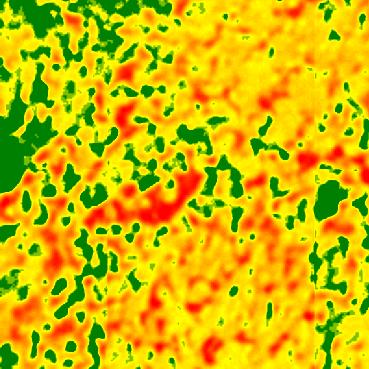}
\caption{\scriptsize {Our change map}  }
\label{fig:supmat-change:preddiff}
\end{subfigure}

&
\begin{subfigure}{.16\textwidth}
\includegraphics[width=\textwidth]{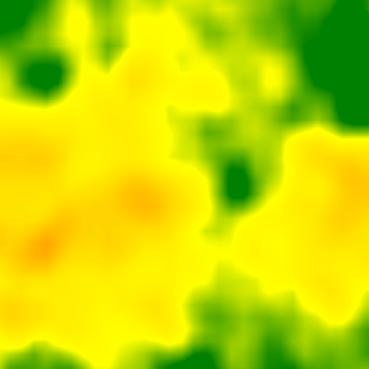}
\caption{\scriptsize  Sentinel change map }
\label{fig:supmat-change:formsdiff}
\end{subfigure}

\\
\begin{subfigure}{.16\textwidth}
\includegraphics[width=\textwidth]{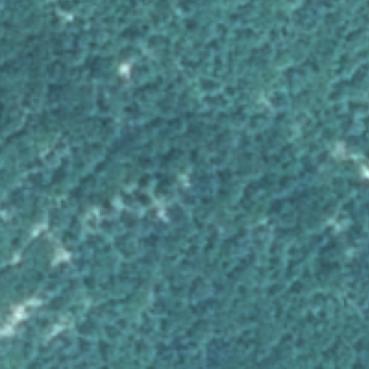}
\caption{\scriptsize VHR year 2023 }
\label{fig:supmat-change:spot2}
\end{subfigure}
&
\begin{subfigure}{.16\textwidth}
\includegraphics[width=\textwidth]{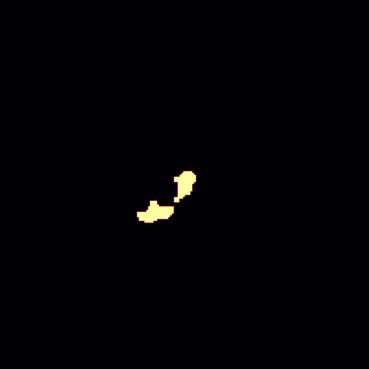}
\caption{\scriptsize ALS change mask }
\label{fig:supmat-change:alsmask}
\end{subfigure}

&
\begin{subfigure}{.16\textwidth}
\includegraphics[width=\textwidth]{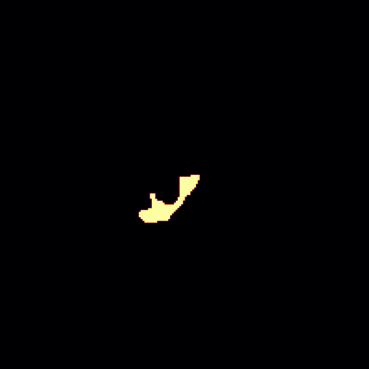}
\caption{\scriptsize Our change mask}
\label{fig:supmat-change:predmask}
\end{subfigure}

&
\begin{subfigure}{.16\textwidth}
\includegraphics[width=\textwidth]{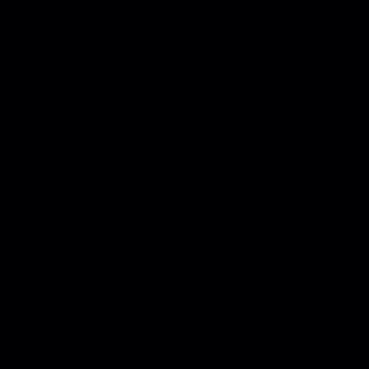}
\caption{\scriptsize {Sentinel change mask}  }
\label{fig:supmat-change:formsmask}
\end{subfigure}

&
\multirow[t]{2}{*}[+1.1cm]{
\hspace{-0.5cm}
\def\colormapheight{4.5cm}
\pgfplotsset{
    /pgfplots/colormap={treegrowth}{
    rgb255(0cm)=(255,0,0), 
    rgb255(1cm)=(255,125,111), 
    rgb255(2cm)=(255,255,0), 
    rgb255(3cm)=(130,192,0), 
    rgb255(4cm)=(0,130,0), 
    }
}

\begin{tikzpicture}
    \begin{axis}[
        hide axis,
        scale only axis,
        height=5pt,
        width=50pt,
        xlabel={in m},
        colormap name=treegrowth,
        colorbar,
            colorbar style={
            width=.2cm,
            height=\colormapheight,
            ytick={0,0.5,1,1.5,2,3,4},
            yticklabels={-20m,-15m,-10m,-5m,0m,+1m,+2m},
            yticklabel style={font=\tiny},
            major tick length=1.5pt,
            line width=.05mm,
            grid style={draw=none} 
        },
        point meta min=0,
        point meta max=4,
    ]
    \addplot [draw=none] coordinates {(0,0)};
    \end{axis}
 
\end{tikzpicture}
}
\setcounter{subfigure}{0}
\\\midrule
\begin{subfigure}{.16\textwidth}
\includegraphics[width=\textwidth]{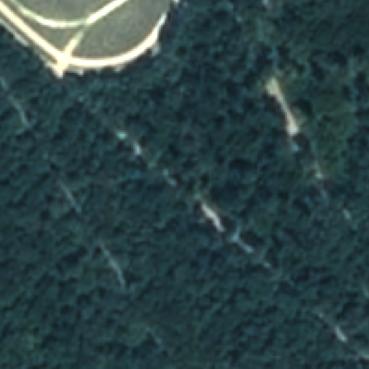}
\caption{\scriptsize VHR year 2022 }
\label{fig:supmat-change:spot1}
\end{subfigure}
&
\begin{subfigure}{.16\textwidth}
\includegraphics[width=\textwidth]{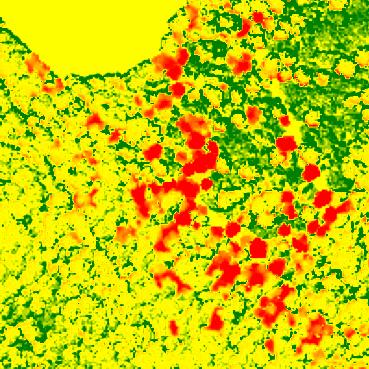}
\caption{\scriptsize  ALS change map }
\label{fig:supmat-change:alsdiff}
\end{subfigure}
&
\begin{subfigure}{.16\textwidth}
\includegraphics[width=\textwidth]{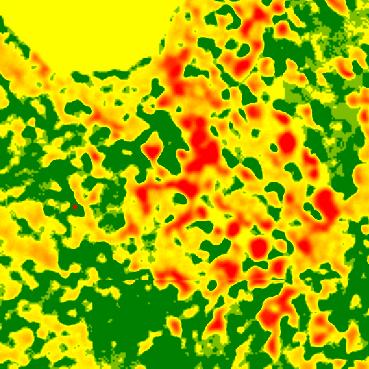}
\caption{\scriptsize {Our change map}  }
\label{fig:supmat-change:preddiff}
\end{subfigure}

&
\begin{subfigure}{.16\textwidth}
\includegraphics[width=\textwidth]{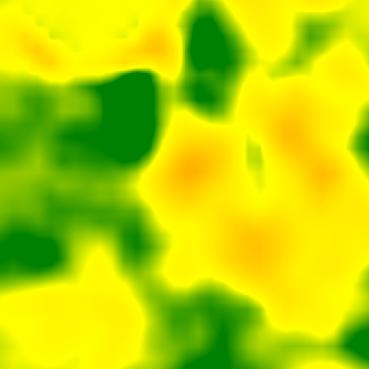}
\caption{\scriptsize {Sentinel change map}  }
\label{fig:supmat-change:formsdiff}
\end{subfigure}

\\
\begin{subfigure}{.16\textwidth}
\includegraphics[width=\textwidth]{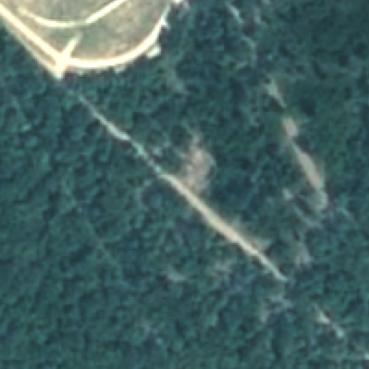}
\caption{\scriptsize VHR year 2023 }
\label{fig:supmat-change:spot2}
\end{subfigure}
&
\begin{subfigure}{.16\textwidth}
\includegraphics[width=\textwidth]{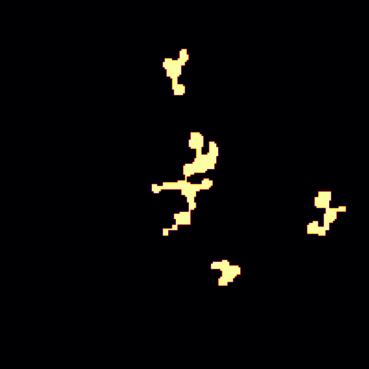}
\caption{\scriptsize ALS change mask }
\label{fig:supmat-change:alsmask}
\end{subfigure}

&
\begin{subfigure}{.16\textwidth}
\includegraphics[width=\textwidth]{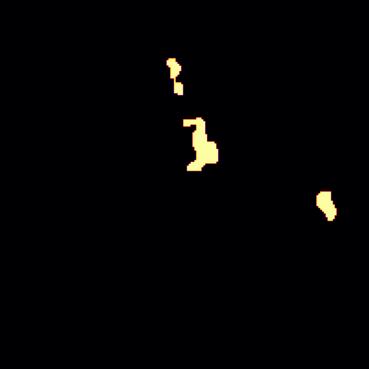}
\caption{\scriptsize Our change mask}
\label{fig:supmat-change:predmask}
\end{subfigure}
&
\begin{subfigure}{.16\textwidth}
\includegraphics[width=\textwidth]{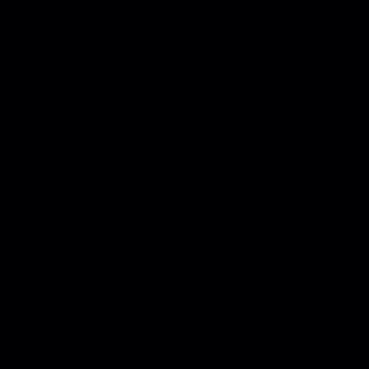}
\caption{\scriptsize {Sentinel change mask}  }
\label{fig:supmat-change:formsmask}
\end{subfigure}

&
%
\setcounter{subfigure}{0}
\\\midrule
\begin{subfigure}{.16\textwidth}
\includegraphics[width=\textwidth]{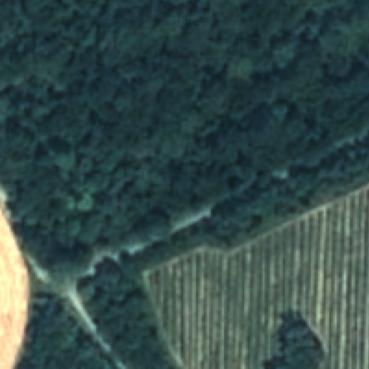}
\caption{\scriptsize VHR year 2022 }
\label{fig:supmat-change:spot1}
\end{subfigure}
&
\begin{subfigure}{.16\textwidth}
\includegraphics[width=\textwidth]{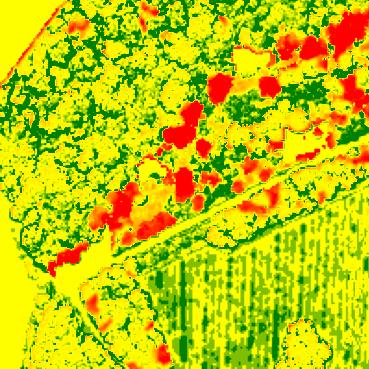}
\caption{\scriptsize  ALS change map }
\label{fig:supmat-change:alsdiff}
\end{subfigure}
&
\begin{subfigure}{.16\textwidth}
\includegraphics[width=\textwidth]{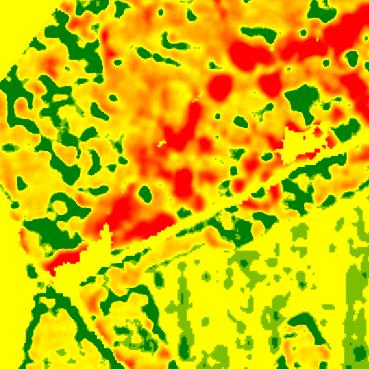}
\caption{\scriptsize {Our change map}  }
\label{fig:supmat-change:preddiff}
\end{subfigure}
&
\begin{subfigure}{.16\textwidth}
\includegraphics[width=\textwidth]{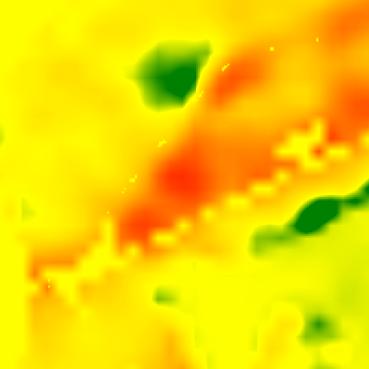}
\caption{\scriptsize {Sentinel change map}  }
\label{fig:supmat-change:formsdiff}
\end{subfigure}

\\
\begin{subfigure}{.16\textwidth}
\includegraphics[width=\textwidth]{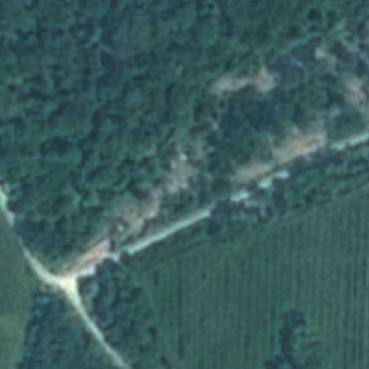}
\caption{\scriptsize VHR year 2023 }
\label{fig:supmat-change:spot2}
\end{subfigure}
&
\begin{subfigure}{.16\textwidth}
\includegraphics[width=\textwidth]{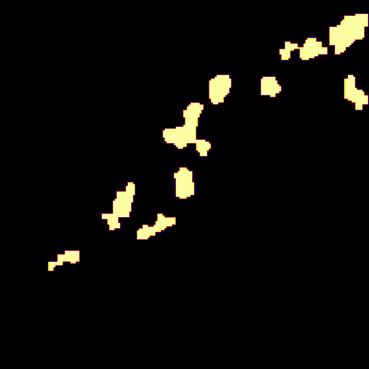}
\caption{\scriptsize ALS change mask }
\label{fig:supmat-change:alsmask}
\end{subfigure}

&
\begin{subfigure}{.16\textwidth}
\includegraphics[width=\textwidth]{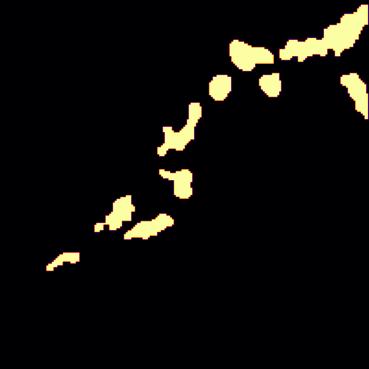}
\caption{\scriptsize Our change mask}
\label{fig:supmat-change:predmask}
\end{subfigure}
&
\begin{subfigure}{.16\textwidth}
\includegraphics[width=\textwidth]{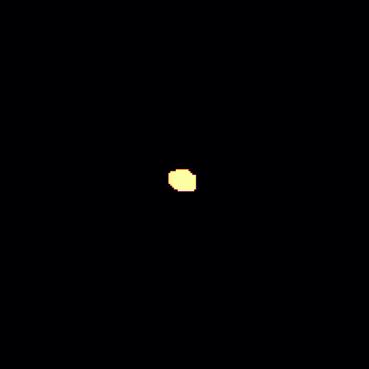}
\caption{\scriptsize {Sentinel change map}  }
\label{fig:supmat-change:formsmask}
\end{subfigure}

\end{tabular}

    \caption{{\bf Canopy Height Change.} 
    We consider VHR images taken in 2022 and 2023 in Chantilly Forest: \subref{fig:supmat-change:spot1} and \subref{fig:supmat-change:spot2}, and use ALS observations of the same years to derive a canopy height change map \subref{fig:supmat-change:alsdiff}. We compare this map to the ones predicted by a PVTv2 model \subref{fig:supmat-change:preddiff} and by a model from Schwartz \etal trained on Sentinel data \citep{schwartz2023forms}. We also compare the binary change masks derived from ALS measurements \subref{fig:supmat-change:alsmask} and from predicted change maps: \subref{fig:supmat-change:predmask} and \subref{fig:supmat-change:formsmask}. Scale and orientation are shared across all subfigures.
    }
    \label{fig:supmat-change}
\end{figure*}

\cref{tab:change-hparams} presents the IoU metrics for various combinations of these parameters. Naturally, focusing on larger change areas simplifies the detection problem due to reduced complexity. The influence of the minimum tree height change threshold is less straightforward; higher thresholds require precise detection of significant height reductions, which can be more challenging. Our chosen parameters—15 m minimum height difference and 200 m$^2$ minimum change area—represent changes that are visually detectable between images (see \cref{fig:supmat-change}), providing a realistic yet challenging task for computer vision models.
01
\begin{table*}[h]  
    \caption{{\bf Canopy Height Change Detection} We compute the IoU metric (in \%) for various minimum height difference (row, in m) and minimum contiguous area of change (column, in m$^2$). The values chosen in the benchmark are \underline{underlined}.}
    \label{tab:change-hparams}
    \centering
\begin{tabular}{r d{4.1} d{4.1} d{4.1} d{4.1} d{4.1} d{4.1}}
\toprule
\diagbox{min\\diff}{min\\surf}& \multicolumn{1}{r}{10 \text{m}$^2$} & \multicolumn{1}{r}{25 m$^2$}  & \multicolumn{1}{r}{100 m$^2$} & \multicolumn{1}{r}{\underline{200 m$^2$}} & \multicolumn{1}{r}{300 m$^2$} & \multicolumn{1}{r}{400 m$^2$}
\\ \midrule
-5 m  & \applycolor{7.0} & \applycolor{7.1} & \applycolor{7.2} & \applycolor{6.2} & \applycolor{5.2} & \applycolor{4.2} \\ 
-10 m & \applycolor{17.1} & \applycolor{17.9} & \applycolor{22.6} & \applycolor{23.6} & \applycolor{25.1} & \applycolor{28.7} \\ 
\underline{-15 m} & \applycolor{22.1} & \applycolor{23.4} & \applycolor{28.8} & \applycolor{37.0} & \applycolor{40.6} & \applycolor{40.8} \\ 
-20 m & \applycolor{18.9} & \applycolor{20.2} & \applycolor{31.4} & \applycolor{36.6} & \applycolor{31.8} & \applycolor{31.5} \\

\bottomrule
\end{tabular}

\end{table*}

\subsection{Impact of Initialization Strategy}

\begin{table*}   
    \caption{{\bf Ablation Study.} We evaluate the impact of omitting the NIR channel from input images and assess various initialization strategies for fine-tuning networks initially trained only on RGB data to accommodate an additional NIR channel.}
    \label{tab:ablation}
    \centering
    


\begin{tabular}{lll cccc}
\toprule
  &          & &MAE (m) 	&nMAE (\%) 	&RMSE (m) 	&Bias (m) 
  \\\midrule
  Channels & backbone & {pretraining}
  \\\midrule
  RGB    & UNet& ImageNet1K & 2.77& 24.8& 4.34& -0.17\\
  RGB+IR  & UNet& ImageNet1K  & 2.67& 23.8& 4.18& -0.30
   \\\greyrule
   RGB    & PVTv2& ImageNet1K & 3.73& 32.6& 5.53& -0.50 \\
   RGB+IR  & PVTv2& ImageNet1K & \textbf{2.52}& \textbf{22.9}& \textbf{4.02}& \textbf{0.00} 
         \\\midrule
 {Initialization} & backbone & {pretraining} 
         \\\midrule
\rowcolor{gray!10} {Fully random} & & & 11.17    &85.77   & 14.38 & -10.94 	\\
{Rand. 1st layer}                   &&& 2.87& 24.3& 4.24& -0.04 \\
\rowcolor{gray!10} {LoRA (rank 32)}  &&& 3.64& 32.8& 5.40& -0.27  \\
 {Proposed}  & 
 \makebox[0pt][l]{\smash{\raisebox{\dimexpr3\baselineskip+16.5pt\relax}{PVTv2}}}  &
 \makebox[0pt][l]{\smash{\raisebox{\dimexpr3\baselineskip+16.5pt\relax}{ImageNet1K}}}
                                    &   \textbf{2.52}& \textbf{22.9}& \textbf{4.02}& \textbf{0.00} 
         \\\bottomrule     
\end{tabular}
\end{table*}

We provide in \cref{tab:ablation} the results of ablation experiments. We evaluate the impact of omitting the near infrared (NIR) band from input images. We can see in \cref{tab:ablation} that removing the NIR channel from input images decreases the performance for both UNet and PVTv2 backbones. Moreover, we assess various initialization strategies for fine-tuning networks initially trained only on RGB data to accommodate an additional NIR channel. Those include training from scratch, randomizing the first layer, and using LoRa. In \cref{fig:lora-finetunning} we show the results for different LoRa ranks and show only the best rank ($32$) in \cref{tab:ablation}. We see a clear benefit in using our proposed initialization scheme.

\if 10
\YOH{\subsection{Tolan et al. fine-tunning}
Following the authors recommendation we fine-tune Tolan et al.\citep{tolan2024very} ViT Large using the following parameters: Image size of 256 pixels, AdamW optimizer with a learning rate of $3\times 10^{-5}$ and a weight decay of 0.01.}
\fi

\section{Dataset description}
\label{sec:sup:dataset}
We describe here in details the dataset used in Open-Canopy and provide information about its constitution.

\subsection{Access}

\begin{compactitem}

\item The dataset and model weights are hosted at [URL]
with download and usage instructions at [URL].

\item The data is governed by the Open License 2.0 of Etalab (\url{https://www.etalab.gouv.fr/wp-content/uploads/2018/11/open-licence.pdf}).

\item Codes for data preprocessing, training models and evaluation are available at [URL].

\end{compactitem}

\subsection{Composition}
\label{sec:dataset-composition}

We describe here the organization of the dataset. See \secref{sec:dataset-preparation} for details on how the dataset was prepared.

The dataset is organized in the following way:
\begin{compactitem}
\item The folder \texttt{canopy\_height} contains data for canopy height estimation.
\item The folder \texttt{canopy\_height\_change} contains data for canopy height change estimation.
\end{compactitem}

The composition of the \texttt{canopy\_height} folder is the following:
\begin{compactitem}
\item The file \texttt{geometries.geojson} stores a list of 95,429 $1 km^2$ square geolocated geometries, giving access to the splits of the dataset. It can be loaded using the python package geopandas \footnote{\url{https://geopandas.org/en/stable/}}.
Each geometry designates either a train, validation, test or buffer area. This information is stored in the column \texttt{split}.
There are 8,046 \texttt{buffer} tiles, 66,339 train tiles, 7,369 validation tiles and 13,675 test tiles. Additionally, each geometry is associated to a year (corresponding to the year of the corresponding LiDAR acquisition), stored in the column \texttt{lidar\_year}.
\item The file \texttt{forest\_mask.parquet} stores geolocated geometries of forests' outlines. It can be loaded using the python package geopandas. The parquet format is used to accelerate loading time.
\item Each folder 2021, 2022 and 2023 contains three files:
\begin{compactitem}
\item spot.vrt is a geolocalized virtual file that gives access to SPOT 6-7 images stored in the subfolder \texttt{spot}. It can be accessed through Qgis software \footnote{\url{https://www.qgis.org/en/site/}3} or python rasterio library \footnote{\url{https://rasterio.readthedocs.io/en/stable/}} for instance. It has the same extent as the geometries of the associated year.
\item Similarly \texttt{lidar.vrt} gives access to ALS-derived (LiDAR) canopy height maps stored in the subfolder \texttt{lidar}.
\item Similarly \texttt{lidar\_classification.vrt} gives access to classification rasters stored in the subfolder \texttt{lidar\_classification}.
\end{compactitem}
\end{compactitem}

The composition of the \texttt{canopy\_height\_change} folder is the following:
\begin{compactitem}
\item The file \texttt{spot\_1.tif} is a geolocalized image extracted from SPOT 6-7 images in the year 2022 in the area of Chantilly,  France. 
\item The file \texttt{spot\_2.tif} is a geolocalized image extracted from SPOT 6-7 images in the year 2023 in the area of Chantilly (France).
\item The file \texttt{lidar\_1.tif} is a geolocalized ALS-derived height map in the year 2022 in the area of Chantilly (France), derived from LiDAR HD \citep{lidarhd}.
\item The file \texttt{lidar\_2\_m.tif} is a geolocalized ALS-derived height map in the year 2023 in the area of Chantilly (France), provided by \citep{chantilly}, at a resolution of 1m, with height in meters, and covering only forests.
\item The file \texttt{predictions\_1\_m.tif} is a geolocalized height map predicted by a PVTv2 model in 2022 in the area of Chantilly (France), in meter unit.
\item The file \texttt{predictions\_2\_m.tif} is a geolocalized height map predicted by a PVTv2 model in 2023 in the area of Chantilly (France), in meter unit.
\item The file \texttt{lidar\_classification.tif} is an ALS-derived classification raster in  2022 in the area of Chantilly (France).
\item Additionally, files that follow the following pattern \texttt{*\_masked.tif} designate images masked on the extent of the available ALS data for 2023.
\item The file \texttt{change\_mask\_delta\_15\_surface\_200\_}  \texttt{annotated.geojson} can be loaded with geopandas and gives access to geometries detected as "change" for a minimum height difference of 15m and a minimum surface of 200m. We also provide manual annotations of detections in the column "Rating", where "true" indicates a true positive and "false" a false positive.

\end{compactitem}

\subsection{Characteristics}

\begin{compactitem}

\item We provide SPOT 6-7 images, ALS-derived height maps and classification rasters covering 95,429 km$^2$ (including a "buffer" area of 8046 km$^2$, a train area of 66,339 km$^2$, a validation area of 7,369 km$^2$ and a test area of 13,675 km$^2$). 
Each image has a resolution of 1.5m, with one annotation per pixel, for a total of 42,455,312,381 annotations.
\item Additionally, we provide SPOT 6-7 imagery, ALS-derived height maps and a classification raster on the Chantilly forest area for 2022 and 2023 (166 km$^2$).

\item The Open-Canopy dataset is derived from a larger dataset of SPOT 6-7 acquisitions across the full metropolitan French territory between 2013 and 2023 \footnote{\url{https://openspot-dinamis.data-terra.org}\hspace*{\fill}}, and a larger dataset of ALS acquisitions from the IGN campaign that started in 2021 and aims at covering the full metropolitan French territory (LiDAR HD) \footnote{\url{https://geoservices.ign.fr/lidarhd}}.
The Open-Canopy dataset focuses on domains that are representative of the diversity of French forests and where LiDAR HD is available at the time of writing, with the goal of limiting the dataset's size to approximately 300 GB, in order to facilitate its usage by the machine learning community.

\item Each SPOT image is at a resolution of $1.5$\:m per pixel, and features $4$ spectral channels: red, blue, green, and near-infrared. 
\item Each height map image is at a resolution of $1.5$\:m per pixel, and features 1 channel (height in decimeters except if notified in the filename in the following format: "<name>\_<unit>.tif").
\item Each classification image is at a resolution of $1.5$\:m per pixel, and features 1 channel (classification \citep{LIDARHD_doc} for a description of classes). Forests' outlines are stored as geometries in a parquet file. A Python utility is provided to create a vegetation mask from the classification raster and the forests' outlines.
\end{compactitem}

\begin{figure*} 
    \centering
    \centering
\begin{tabular}{cc}
    \begin{subfigure}{0.45\textwidth}
        \begin{tikzpicture}
            \begin{axis}[
                width=1\textwidth,
                xlabel={\small LoRA rank},
                ylabel={\small $\leftarrow$ MAE},
                legend pos=north west,
                xmin=00,
                xmax=128,
                ymin=2,
                ymax=6
            ]
            \addplot[
                mark=*,
                color=red
            ] 
            table [
                x index=0, 
                y index=1, 
                col sep=comma
            ] {figures/lora_data.csv};
            \addlegendentry{PVTv2 with LoRA}

            \addplot[
                blue,
                thick,
                solid
            ] coordinates {
                (0,2.52) (128,2.52)
            };
            \addlegendentry{PVTv2 fine-tuned}
            
            \end{axis}
        \end{tikzpicture}

    \end{subfigure}
&
    \begin{subfigure}{0.45\textwidth}
        \begin{tikzpicture}
            \begin{axis}[
                width=1\textwidth,
                xlabel={\small LoRA rank},
                ylabel={\small $\leftarrow$ nMAE (\%)},
                legend pos=south east,
                xmin=00,
                xmax=128,
                ymin=20,
                ymax=50,
                yticklabel={\hphantom{0}\pgfmathprintnumber{\tick}} 
            ]
            \addplot[
                mark=*,
                color=red
            ] 
            table [
                x index=0, 
                y index=3, 
                col sep=comma
            ] {figures/lora_data.csv};

            \addplot[
                blue,
                thick,
                solid
            ] coordinates {
                (0,22.9) (128,22.9)
            };
            
            \end{axis}
        \end{tikzpicture}

    \end{subfigure}
\\
    \begin{subfigure}{0.45\textwidth}
        \begin{tikzpicture}
            \begin{axis}[
                width=1\textwidth,
                xlabel={\small LoRA rank},
                ylabel={\small $\leftarrow$ RMSE},
                legend pos=south east,
                xmin=00,
                xmax=128,
                ymin=3.8,
                ymax=8
            ]
            \addplot[
                mark=*,
                color=red
            ] 
            table [
                x index=0, 
                y index=2, 
                col sep=comma
            ] {figures/lora_data.csv};
            
            \addplot[
                blue,
                thick,
                solid
            ] coordinates {
                (0,4.02) (128,4.02)
            };
            
            \end{axis}
        \end{tikzpicture}
    \end{subfigure}
&
    \begin{subfigure}{0.45\textwidth}
        \begin{tikzpicture}
            \begin{axis}[
                width=1\textwidth,
                xlabel={\small LoRA rank},
                ylabel={\small IoU (\%) $\rightarrow$},
                legend pos=south east,
                xmin=00,
                xmax=128,
                ymin=80,
                ymax=95
            ]
            \addplot[
                mark=*,
                color=red
            ] 
            table [
                x index=0, 
                y index=4, 
                col sep=comma
            ] {figures/lora_data.csv};

            \addplot[
                blue,
                thick,
                solid
            ] coordinates {
                (0,90.5) (128,90.5)
            };
            
            \end{axis}
        \end{tikzpicture}

    \end{subfigure}

\end{tabular}
    \caption{{{\bf LoRa Fine-Tuning of PVTv2}. We fine-tun PVTv2 using LoRa for different rank. To allow the network to adapt to the NIR modality, we still train the first layer fully. The best results, obtained with rank $= 32$, are noticeably inferior to a fully fine-tunned PVTv2 } }
    \label{fig:lora-finetunning}
\end{figure*}

\section{Dataset preparation}
\label{sec:dataset-preparation}

\subsection{Splits}
\label{sec:supp:dataset:splits}
Our sampling strategy is semi-automated and proceeds as follows:
\begin{compactitem}
\item SPOT images were associated to LiDAR height maps of the same year and geolocation (each LiDAR height map corresponds to a 1km$^2$ geolocalized square tile, referred to as ``geometry'' in the following).
\item Geometries on overlapping areas between spot full images were removed.
\item Geometries that had more than 100 zeros on the first spot band (\eg, on edges of a full spot image) were discarded to avoid tiles with missing data.
\item Test geometries of 1km$^2$ were sampled (with a fixed seed) to form contiguous squares of 7km$^2$ and to cover 20,000 km$^2$. 
\item Test geometries that overlapped each other were dropped.
\item Test geometries that covered different years in terms of LiDAR acquisitions were dropped.
\item This process resulted in a total test area of 13,675 km$^2$.
\item A buffer of 1$km$ was applied around each test area of 7km$^2$.
\item Validation and train geometries were randomly sampled (with a fixed seed) among the remaining geometries, with a proportion of 10\% for validation and 90\% for training.
\item This process resulted in a training area of 66,339 km$^2$ and a validation area of 7,369 km$^2$.

\subsection{SPOT 6-7 satellite imagery}
\begin{compactitem}
\item The aerial images are sampled from the DINAMIS \footnote{\url{https://openspot-dinamis.data-terra.org}} collection. This collection consists of an annual mosaic of selected tiles taken by SPOT 6-7 satellites between March and October of each year between 2013 and 2023, covering the entire French metropolitan territory. All images are orthorectified by IGN and mapped onto a unified cartographic coordinate reference system (Lambert 93). Each tile consists of an image with four spectral bands: red, green, blue, and near-infrared at a resolution of $6$m, and an image with one panchromatic band at a resolution of $1.5m$ that can be downloaded separately.
\item A total of 52 pairs of spectral and panchromatic images were downloaded from the DINAMIS website, for each year from 2021 to 2023, to cover a very diverse range of forest types in areas where LiDAR HD was available at the time of the creation of the dataset.
\item We applied pansharpening with the weighted Brovey algorithm \citep{gillespie1987color} to upsample all four spectral bands to a resolution of $1.5$m, resulting in one image with four bands for each tile. 
\item We cropped each image to the area covered by the ALS acquisitions of the same year.
\item Pixels values were clipped to a maximum value of 2000 to avoid outliers (upper bound both quantitatively and qualitatively assessed through histograms and visualization).
\item Resulting images were normalized to a 0-255 range and saved as uint8 in a block-tiled compressed tiff format ($256\times256$).
\item The pansharpening and normalization procedures were voluntarily kept relatively simple in order to facilitate reproducibility. They may not be optimal for visualization, \eg, lacking harmonization, but we expect deep learning models to be robust to such variations in input data.
\end{compactitem}

\subsection{ALS data}

\begin{compactitem}
\item The ALS classified point clouds were downloaded from the \href{https://geoservices.ign.fr/lidarhd}{LiDAR HD} website (IGN). A reference to each download link is saved in the file \texttt{geometries.geojson}.
\item For each geometry, canopy height images were derived from ALS data by taking the maximum difference between the height of each point and the one of its nearest point classified as ground within its pixel, interpolating values in areas without data. 
\item LiDAR point clouds were classified by IGN into the main types of land cover (water, ground, high vegetation over 1.5m, buildings...). We use this classification to produce classification rasters at a resolution of 1.5m, where each pixel takes the value of the most frequent class of the corresponding LiDAR points.
\item We then create vegetation masks by taking the union of the ALS-derived mask indicating vegetation over 1.5m in height, with the official forest plots outlines (file \texttt{forest\_mask.parquet}), both provided by IGN. The resulting vegetation masks cover trees and shrubs within forest plots as well as outside, such as hedges and urban trees.
\item The official forests' outlines were extracted from ``BD foret'' \footnote{\url{https://geoservices.ign.fr/bdforet\#telechargementv2}} and "simplified" using geopandas python library to a precision of 10m, with the goal to limit their size.
\end{compactitem}

\end{compactitem}

\section{Datasheet for Open-Canopy dataset}
\label{sec:datasheet}
\begin{compactitem}

\subsection{Motivation}

\item \textbf{For what purpose was the dataset created?} Was there a specific task in mind? Was there a particular gap that needed to be filled? Please provide a description.

The Open-Canopy dataset was created to train and evaluate models that  (i) predict very-high resolution canopy height maps from satellite imagery using LiDAR scans for ground truth, and (ii) detect canopy height changes between images from different years.
The main gap we are addressing is the lack of curated open-source datasets with both very high resolution imagery and ALS-based (LiDAR) canopy height maps.

\item \textbf{Who created the dataset (e.g., which team, research group) and on behalf of which entity (e.g., company, institution, organization)?}
This dataset was curated by a team of researchers from \REDACTED{ENS Paris (Ecole Normale Supérieure), LSCE (Laboratoire des Sciences du Climat et de l'Environnement), ENPC (Ecole des Ponts ParisTech), and IGN (the French National Institute of Geographical and Forest Information),} using data made available by \href{https://dinamis.data-terra.org}{DINAMIS} and {IGN}. DINAMIS \citep{dinamis} is a French platform that provides access to earth observation products for public benefit programs. {The IGN is a French public state administrative establishment aiming to produce and maintain geographical information for France.}

\item \textbf{Who funded the creation of the dataset?} If there is an associated grant, please provide the name of the grantor and the grant name and number.

The funding of the Open-Canopy dataset is 100\% public.
Open-Canopy benefited from funding by \REDACTED{the French National Research Agency (grant \href{https://anr.fr/Projet-ANR-22-FAI1-0002}{ANR-22-FAI1-0002}).}

\item \textbf{Any other comments?}

\answerNA{}

\subsection{Composition}

\item \textbf{What do the instances that comprise the dataset represent (e.g., documents, photos, people, countries)?}

The dataset is split into square areas of width $1.0005$ km, rasterized to a $1.5$ m resolution ($667 \times 667$ pixels). Each instance corresponds to an area of $1$ km$^2$ on the French metropolitan territory.


\item \textbf{How many instances are there in total (of each type, if appropriate)?}

We provide 95,429 instances of $1$km$^2$: 66,339 train tiles, 7,369 validation tiles,  13,675 test tiles, and  8,046 ``buffer'' tiles. This corresponds to a total of 42,455,312,381 individual annotated pixels.
%

\item \textbf{Does the dataset contain all possible instances or is it a sample (not necessarily random) of instances from a larger set?} 

The Open-Canopy dataset covers $17\%$ of the French metropolitan territory. It is derived from a larger dataset of SPOT 6-7 acquisitions across the full metropolitan French territory between 2013 and 2023 (\url{https://openspot-dinamis.data-terra.org}), and a larger dataset of ALS acquisitions from the campaign that started in 2021 and aims at covering the full metropolitan French territory (\href{https://geoservices.ign.fr/lidarhd}{LiDAR HD})\citep{lidarhd}.
The Open-Canopy dataset focuses on domains that are representative of the diversity of French forests and where LiDAR HD is available at the time of submission. We also aimed to limit the dataset's size to 300 GB to facilitate its use.

\item \textbf{What data does each instance consist of?}

Each instance consists of a GeoJSON geometry ($1km^2$), for which a $667\times667$ SPOT image, a height map, and a vegetation mask can be extracted from associated .vrt files, in order to associate to each pixel the following values:
(i) RGB and near Infrared channels derived from pan-sharpened and ortho-rectified satellite images from SPOT 6-7 acquired between 2021 and 2023; 
(ii) canopy height derived from LiDAR HD's 3D point clouds \citep{lidarhd}  acquired in the same year;
(iii) label (\eg, vegetation, ground, water, building) derived from LiDAR HD's 3D point clouds \citep{lidarhd}.

Additionally, we provide forest outlines obtained from IGN's portal \citep{link_to_forest_masks} stored as a parquet file.


\item \textbf{Is there a label or target associated with each instance?} 

\answerYes{} We provide a complete pixel-precise height map and classification raster of the same extent as the satellite images.

\item \textbf{Is any information missing from individual instances?} 

\answerNo{} We provide dense information (radiometry, canopy height, class label) for all pixels with the exception of areas that have been selected by the French government as ``sensitive''  for security reasons (\eg, nuclear plants, military area). We do not provide the 3D point clouds from LiDAR HD, but they are accessible on their platform.

\item \textbf{Are relationships between individual instances made explicit (e.g., users' movie ratings, social network links)?} 

\answerNA{}

\item \textbf{Are there recommended data splits (e.g., training, development/validation, testing)?} 

Yes, we provide data splits for reproducing the results of the benchmark. The test split has been explicitly selected to address the complex domain shifts of geospatial data and separated from the train and validation splits by a 1 km$^2$ buffer to avoid data contamination.

\item \textbf{Are there any errors, sources of noise, or redundancies in the dataset?} 

The annotations from ALS (LiDAR) data include inherent inaccuracies due to the nature of the acquisition process. Multipath effects from multiple echoes can introduce errors, and outlier points may impact the quality of the canopy height maps. Additionally, variations in tree height due to different acquisition times across seasons can affect consistency between ALS and VHR acquisitions, as trees might be at various stages of their growth cycle.
Input images sourced from satellite data pre-processed by IGN and DINAMIS may still exhibit artifacts due to cloud cover or contain small registration errors that can impact the analysis.

Classification rasters derived from ALS data are also subject to inaccuracies. These can stem from inherent limitations in the ALS technology, including noise in the data which may lead to errors in vegetation classification.

\item \textbf{Is the dataset self-contained, or does it link to or otherwise rely on external resources (e.g., websites, tweets, other datasets)?} 
This dataset is self-contained and will be stored on the \href{https://huggingface.co}{Huggingface} platform. The dataset is under the Open License 2.0 of Etalab.

\item \textbf{Does the dataset contain data that might be considered confidential (e.g., data that is protected by legal privilege or by doctor–patient confidentiality, data that includes the content of individuals’ non-public communications)?} 

\answerNo{} The classification raster does not contain any information that would not be available in other open-access sources (DINAMIS, BD-Foret, LiDAR-HD). We have specifically avoided high-risk areas such as military installations or nuclear plants.

\item \textbf{Does the dataset contain data that, if viewed directly, might be offensive, insulting, threatening, or might otherwise cause anxiety?} \textit{If so, please describe why.}

\answerNo{}

\item \textbf{Does the dataset identify any subpopulations (e.g., by age, gender)?}

\answerNo{}

\item \textbf{Is it possible to identify individuals (i.e., one or more natural persons), either directly or indirectly (i.e., in combination with other data) from the dataset?}

\answerNo{} The resolution of 1.5m per pixel and the aerial perspective makes identifying individuals impossible.

\item \textbf{Does the dataset contain data that might be considered sensitive in any way (e.g., data that reveals racial or ethnic origins, sexual orientations, religious beliefs, political opinions or union memberships, or locations; financial or health data; biometric or genetic data; forms of government identification, such as social security numbers; criminal history)?} 

\answerNo{}

\item \textbf{Any other comments?}

\answerNo{}

\subsection{Collection Process}

\item \textbf{How was the data associated with each instance acquired?} 

The satellite images are sampled from the DINAMIS \href{https://openspot-dinamis.data-terra.org}{open SPOT} collection. This collection consists of an annual mosaic of selected images taken by SPOT 6-7 satellites between March and October of each year between 2013 and 2023, covering the entire French metropolitan territory. All images are preprocessed by IGN and mapped onto a unified cartographic coordinate reference system (Lambert 93). 

\item The ALS classified point clouds were downloaded from the \href{https://geoservices.ign.fr/lidarhd}{LiDAR HD} website (IGN).

\item \textbf{What mechanisms or procedures were used to collect the data (e.g., hardware apparatus or sensor, manual human curation, software program, software API)?} 

The IGN selected several acquisition companies through a call for tender with strict specifications.

\item \textbf{If the dataset is a sample from a larger set, what was the sampling strategy (e.g., deterministic, probabilistic with specific sampling probabilities)?}

The sampling strategy was semi-automated. First a manual selection of spot images was manually chosen and downloaded from DINAMIS website, so as to cover a diverse range of forests types in areas where LiDAR HD was also available. Then training, validation, and test splits were randomly sampled, with constraints such as test tiles having a size of 7 $km^2$ and being separated from other tiles by a buffer of 1 $km^2$, and covering an area of about 14,000 $km^2$. See \secref{sec:supp:dataset:splits} for more details.

\item \textbf{Who was involved in the data collection process (e.g., students, crowdworkers, contractors) and how were they compensated (e.g., how much were crowdworkers paid)?}

The data collection process for the dataset was managed by the European Space Agency (ESA), which provided the Very High Resolution (VHR) Imagery, and the French Mapping Agency (IGN), which provided the LiDAR HD data. The curation of this dataset was overseen by two individuals who were associated with academic institutions as a postdoctoral researcher (ENS)  and an intern (LSCE) during the dataset's creation.

\item \textbf{Over what timeframe was the data collected? Does this timeframe match the creation timeframe of the data associated with the instances (e.g., recent crawl of old news articles)?} 

The collection of satellite imagery and ALS data spans from 2021 to 2023, which coincides with the period of availability of LiDAR HD data at the time of the creation of the dataset.

\item \textbf{Were any ethical review processes conducted (e.g., by an institutional review board)?} 

\answerNo{}

\item \textbf{Does the dataset relate to people?} 

\answerNo{}

\item \textbf{Did you collect the data from the individuals in question directly, or obtain it via third parties or other sources (e.g., websites)?}

\answerNA{}

\item \textbf{Were the individuals in question notified about the data collection?} 

\answerNA{}

\item \textbf{Did the individuals in question consent to the collection and use of their data?} 

\answerNA{}

\item \textbf{If consent was obtained, were the consenting individuals provided with a mechanism to revoke their consent in the future or for certain uses?} 

\answerNA{}

\item \textbf{Has an analysis of the potential impact of the dataset and its use on data subjects (e.g., a data protection impact analysis) been conducted?} 

\answerNo{} Given the nature of the dataset---which involves high-resolution canopy height data that does not include personal identifiers or directly impact individual privacy---it is unlikely that the dataset poses significant risks to data subjects. The focus is primarily on environmental features rather than personal data.

\item \textbf{Any other comments?}

\answerNo{}

\subsection{Preprocessing, Cleaning, and/or Labeling}

\item \textbf{Was any preprocessing/cleaning/labeling of the data done (e.g., discretization or bucketing, tokenization, part-of-speech tagging, SIFT feature extraction, removal of instances, processing of missing values)?} 

Canopy Height Maps were derived from ALS data by taking the maximum difference between the height of each point and the one of its nearest point classified as ground within its pixel, interpolating values in areas without data.
\item For vegetation masks, we take the union of the ALS-derived mask indicating vegetation over 1.5m in height, with the official forest plots outlines, both provided by IGN. The resulting vegetation mask covers trees and shrubs within forest plots as well as outside, such as hedges and urban trees. The official forests' outlines were ``simplified'' using geopandas python library to a precision of 10m, in order to limit their size.
\item SPOT 6-7 images were pansharpenened with the weighted Brovey algorithm to upsample all four spectral bands to a resolution of 1.5m. Then all pixels values were clipped to a maximum value of 2000 to avoid outliers and normalized to a 0-255 range to be saved as uint8, in a block-tiled compressed tiff format.

\item \textbf{Was the ``raw'' data saved in addition to the preprocessed/cleaned/labeled data (e.g., to support unanticipated future uses)?} \textit{If so, please provide a link or other access point to the “raw” data.}

\answerYes{}  The raw data can be downloaded from \href{https://openspot-dinamis.data-terra.org}{DINAMIS} and \href{https://geoservices.ign.fr/lidarhd}{LiDAR HD} websites.

\item \textbf{Is the software used to preprocess/clean/label the instances available?} 

\answerYes{} All the codes to preprocess the data are available on the Github of the project \REDACTED{\url{https://github.com/fajwel/Open-Canopy}.}

\item \textbf{Any other comments?}

\answerNo{}

\subsection{Uses}

\item \textbf{Has the dataset been used for any tasks already?} 

\answerNo{}

\item \textbf{What (other) tasks could the dataset be used for?}

We encourage future researchers to use the Open-Canopy dataset for several tasks. Particularly, the dataset could be used to predict land cover in addition to canopy height, using the classification rasters as complimentary labels.
It could also be used for pre-training of models for other tasks such as tree cover segmentation and tree species classification.

\item \textbf{Is there anything about the composition of the dataset or the way it was collected and preprocessed/cleaned/labeled that might impact future uses?} 

This dataset is geographically limited to metropolitan France. Although France's territory is diverse, featuring oceanic, continental, Mediterranean, and mountainous bioclimatic regions, it does not contain tropical or desert areas.
\item The Open-Canopy dataset's reliance on purely optical data may limit the applicability of the models trained on it to regions with pervasive cloud cover.

\item \textbf{Are there tasks for which the dataset should not be used?} 

\answerNo{}

\item \textbf{Any other comments?}

\answerNo{}

\subsection{Distribution}

\item \textbf{Will the dataset be distributed to third parties outside of the entity (e.g., company, institution, organization) on behalf of which the dataset was created?} 

\answerYes{} the dataset will be open-source.

\item \textbf{How will the dataset be distributed (e.g., tarball on website, API, GitHub)?} 

The data will be hosted on Huggingface platform (\REDACTED{\url{https://huggingface.co/datasets/fajwel/Open-Canopy}}), with download and usage instructions on  the Open-Canopy project page hosted on GitHub (\REDACTED{\url{https://github.com/fajwel/Open-Canopy}}).

\item \textbf{When will the dataset be distributed?}

All data is already released under an open-source license, see below.

\item \textbf{Will the dataset be distributed under a copyright or other intellectual property (IP) license, and/or under applicable terms of use (ToU)?} \textit{If so, please describe this license and/or ToU, and provide a link or other access point to, or otherwise reproduce, any relevant licensing terms or ToU, as well as any fees associated with these restrictions.}

\answerYes{} The data is governed by the Open Licence 2.0 of Etalab (\url{https://www.etalab.gouv.fr/wp-content/uploads/2018/11/open-licence.pdf}).

\item \textbf{Have any third parties imposed IP-based or other restrictions on the data associated with the instances?} 

\answerNo{}

\item \textbf{Do any export controls or other regulatory restrictions apply to the dataset or to individual instances?} 

\answerNo{}

\item \textbf{Any other comments?}

\answerNo{}

\subsection{Maintenance}

\item \textbf{Who will be supporting/hosting/maintaining the dataset?}

Hugginface will support hosting of the dataset and metadata. \REDACTED{LSCE} will support maintenance of the dataset in case of revisions.

\item \textbf{How can the owner/curator/manager of the dataset be contacted (e.g., email address)?}

\REDACTED{\url{fajwel.fogel@ens.fr} and \url{loic.landrieu@enpc.fr}}

\item \textbf{Is there an erratum?} 

\answerNo{} There is no erratum for our initial release. Errata will be documented as future releases on the dataset web page.

\item \textbf{Will the dataset be updated (e.g., to correct labeling errors, add new instances, delete instances)?} 

Additional satellite imagery and ALS-derived height maps may be added to future versions of the Open-Canopy dataset. 

\item \textbf{If the dataset relates to people, are there applicable limits on the retention of the data associated with the instances (e.g., were individuals in question told that their data would be retained for a fixed period of time and then deleted)?} 

\answerNA{}.

\item \textbf{Will older versions of the dataset continue to be supported/hosted/maintained?} 

\answerYes{} We are dedicated to providing ongoing support for the Open-Canopy dataset.

\item \textbf{If others want to extend/augment/build on/contribute to the dataset, is there a mechanism for them to do so?} 

Proposed extensions or corrections to the Open-Canopy dataset may be submitted to the providers for consideration. The providers will assess the feasibility of incorporating the suggested modifications, considering factors such as data licensing, maintenance requirements, and relevance.

\item \textbf{Any other comments?}

\answerNo{}

\end{compactitem}

\end{document}